\documentclass[conference]{IEEEtran}
\usepackage{times}

\usepackage[numbers]{natbib}
\usepackage{multicol}
\usepackage[bookmarks=true]{hyperref}

\usepackage[usenames,dvipsnames]{xcolor}
\usepackage{graphicx}
\usepackage{lipsum}
\usepackage{url}            
\usepackage{booktabs}       
\usepackage{amsfonts}       
\usepackage{nicefrac}       
\usepackage{microtype}      
\usepackage{amsmath,amssymb,bbm}
\usepackage{textcomp}       
\usepackage{gensymb}        
\usepackage{multirow}
\usepackage{algorithm}
\usepackage{algorithmic}
\usepackage{subcaption}
\usepackage{wrapfig}        
\usepackage{siunitx}        
\usepackage{mdframed}
\usepackage{footmisc}  
\usepackage[capitalize,noabbrev]{cleveref}  

\newcommand{\Skip}[1]{}

\newcommand{\myfigref}[1]{Figure~\ref{#1}}

\definecolor{gray}{rgb}{0.5, 0.5, 0.5}

\newcommand*\rot{\multicolumn{1}{R{30}{1em}}}

\usepackage{rotating}
\usepackage{adjustbox}
\definecolor{forestgreen}{rgb}{0.13, 0.55, 0.13}
\definecolor{indiagreen}{rgb}{0.07, 0.53, 0.03}
\usepackage{colortbl}
\usepackage{placeins}
\usepackage{array}
\usepackage{fontawesome}
\newcolumntype{R}[2]{%
    >{\adjustbox{angle=#1,lap=\width-(#2)}\bgroup}%
    l%
    <{\egroup}%
}

\hypersetup{
    colorlinks=true,
    linkcolor=black,
    urlcolor=brown,
    citecolor=black,
}

\usepackage{listings}
\usepackage[normalem]{ulem}
\definecolor{codegreen}{rgb}{0,0.6,0}
\definecolor{codegray}{rgb}{0.5,0.5,0.5}
\definecolor{codepurple}{rgb}{0.58,0,0.82}
\definecolor{backcolour}{rgb}{0.95,0.95,0.92}
\lstdefinestyle{mystyle}{
    backgroundcolor=\color{backcolour},   
    commentstyle=\color{codegreen},
    keywordstyle=\color{magenta},
    numberstyle=\tiny\color{codegray},
    stringstyle=\color{codepurple},
    basicstyle=\ttfamily\footnotesize,
    breakatwhitespace=false,         
    breaklines=true,                 
    captionpos=b,                    
    keepspaces=true,                 
    numbers=left,                    
    numbersep=5pt,                  
    showspaces=false,                
    showstringspaces=false,
    showtabs=false,                  
    tabsize=2
}
\begin{document}


\title{FurnitureBench: Reproducible Real-World Benchmark\,for\,Long-Horizon\,Complex\,Manipulation}


\author{
\authorblockN{Minho Heo$^*$}
\authorblockA{KAIST}
\and
\authorblockN{Youngwoon Lee$^*$}
\authorblockA{UC Berkeley}
\and
\authorblockN{Doohyun Lee}
\authorblockA{KAIST}
\and
\authorblockN{Joseph J. Lim}
\authorblockA{KAIST}
}

\makeatletter
\let\@oldmaketitle\@maketitle
\renewcommand{\@maketitle}{\@oldmaketitle
    \centering
    \vspace{-1em}
    \texttt{\href{https://clvrai.com/furniture-bench}{https://clvrai.com/furniture-bench}} \\
    \bigskip
    \medskip
    \includegraphics[width=1.0\linewidth,height=0.37\linewidth]{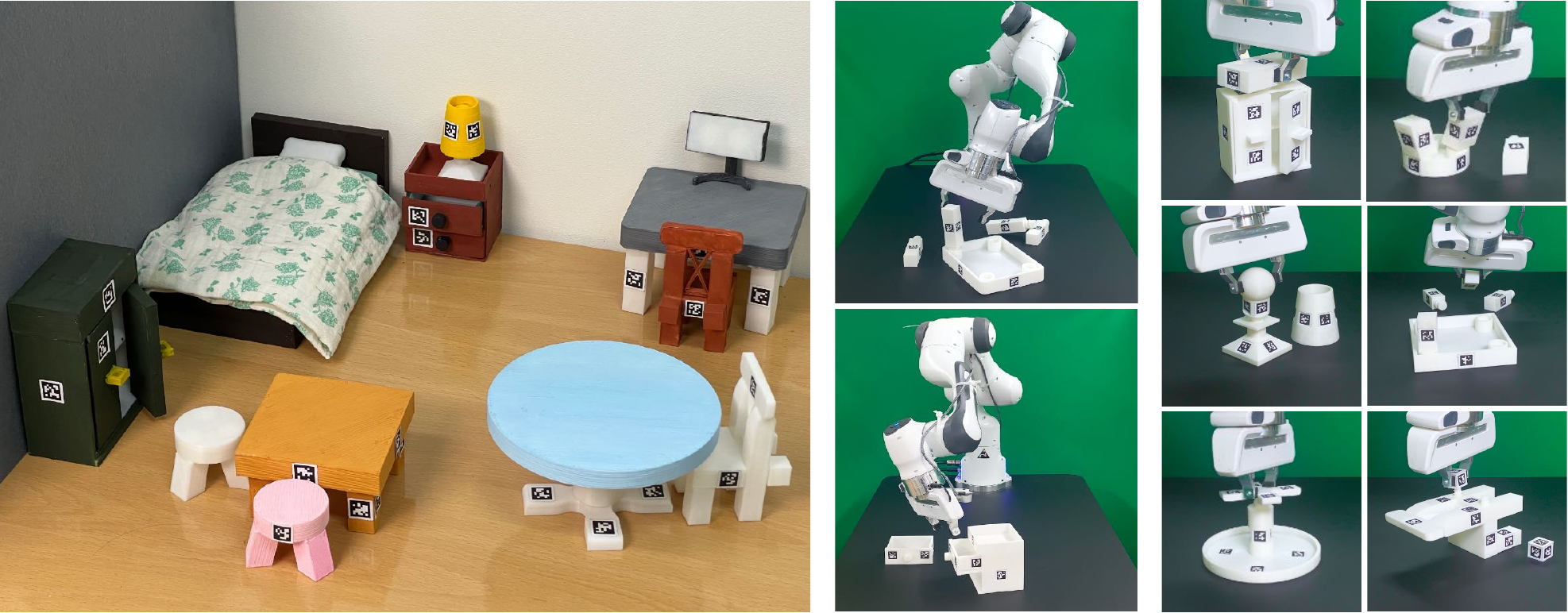}
    \captionof{figure}{
        \textbf{FurnitureBench: reproducible real-world furniture assembly benchmark.} 
        Benchmarking furniture assembly poses to address many robotic manipulation challenges: long-horizon planning, dexterous control, and visual perception. FurnitureBench is designed to be \textit{easy-to-reproduce} and \textit{easy-to-use} with the $3$D printable furniture models, robot control software stack, environment setup guide, and large demonstration data. (Left) A decorated room in the real world with furniture models our robot assembled. (Right) A suite of $8$ furniture models in our benchmark. 
    }
    \label{fig:teaser}
    \smallskip}
\makeatother

\maketitle
\addtocounter{figure}{-1}

\begin{abstract}
    Reinforcement learning (RL), imitation learning (IL), and task and motion planning (TAMP) have demonstrated impressive performance across various robotic manipulation tasks. However, these approaches have been limited to learning simple behaviors in current real-world manipulation benchmarks, such as pushing or pick-and-place. To enable more complex, long-horizon behaviors of an autonomous robot, we propose to focus on real-world furniture assembly, a complex, long-horizon robot manipulation task that requires addressing many current robotic manipulation challenges to solve. We present FurnitureBench, a reproducible real-world furniture assembly benchmark aimed at providing a low barrier for entry and being easily reproducible, so that researchers across the world can reliably test their algorithms and compare them against prior work. For ease of use, we provide 200+ hours of pre-collected data (5000+ demonstrations), 3D printable furniture models, a robotic environment setup guide, and systematic task initialization. Furthermore, we provide FurnitureSim, a fast and realistic simulator of FurnitureBench. We benchmark the performance of offline RL and IL algorithms on our assembly tasks and demonstrate the need to improve such algorithms to be able to solve our tasks in the real world, providing ample opportunities for future research.
\end{abstract}

\renewcommand{\thefootnote}{\fnsymbol{footnote}}
\footnotetext[1]{Equal Contributions.}
\renewcommand*{\thefootnote}{\arabic{footnote}}

\IEEEpeerreviewmaketitle

\section{Introduction}
\label{sec:introduction}

Everyday physical tasks we want to automate with robots, such as doing laundry, tidying messy rooms, cooking meals, and assembling furniture, require a robot to understand environments, plan, and execute complex long-horizon behaviors. These activities have long time spans, consist of diverse semantically composable behaviors, and require dexterous, precise manipulation skills. How can we enable autonomous robots to perform such complex, long-horizon tasks? 

Recently, deep reinforcement learning (RL), imitation learning (IL), and task and motion planning (TAMP) have presented impressive results in robotic manipulation~\citep{levine2016end, suarez2016framework, akkaya2019solving, jang2021bc, ha2021flingbot, rt12022arxiv}. Yet, many of these works are hardly reproducible due to the absence of standardized robots, environments, and software used for experiments. There have been efforts on improving reproducibility in robotic manipulation benchmarks~\citep{lee2021beyond, yang2019replab, ahn2020robel, bottarel2020graspa, 8950086, 8957044, 8989777, bauer2022real}; however, they are still limited to simple short-horizon tasks, such as pick-and-place. 

\begin{figure}[t]
    \centering
    \begin{subfigure}[b]{0.55\linewidth}
        \centering
        \includegraphics[width=\linewidth]{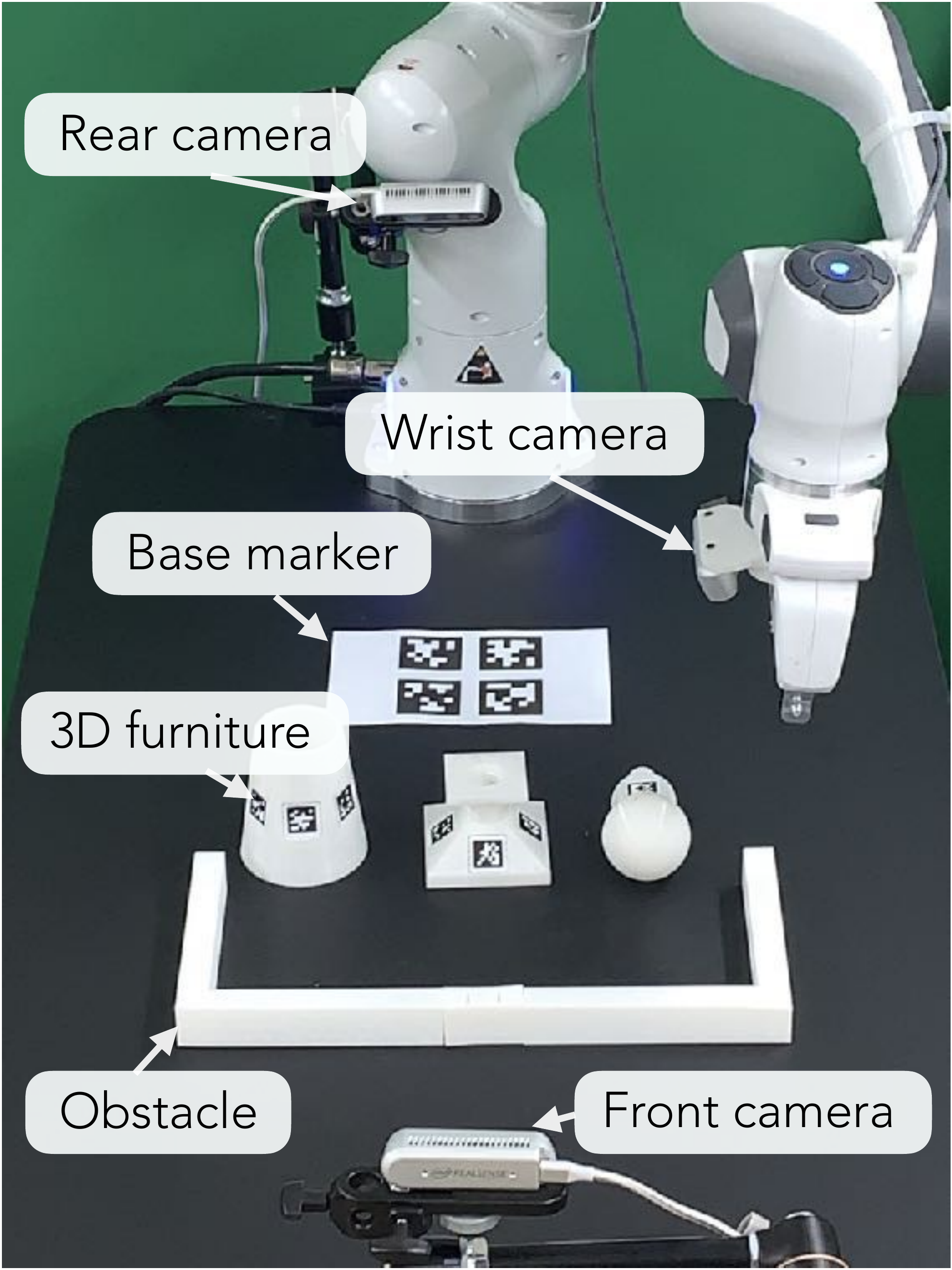}
        \caption{Robot system}
        \label{fig:environment:setup}
    \end{subfigure}
    \begin{subfigure}[b]{0.415\linewidth}
        \centering
        \includegraphics[width=\linewidth]{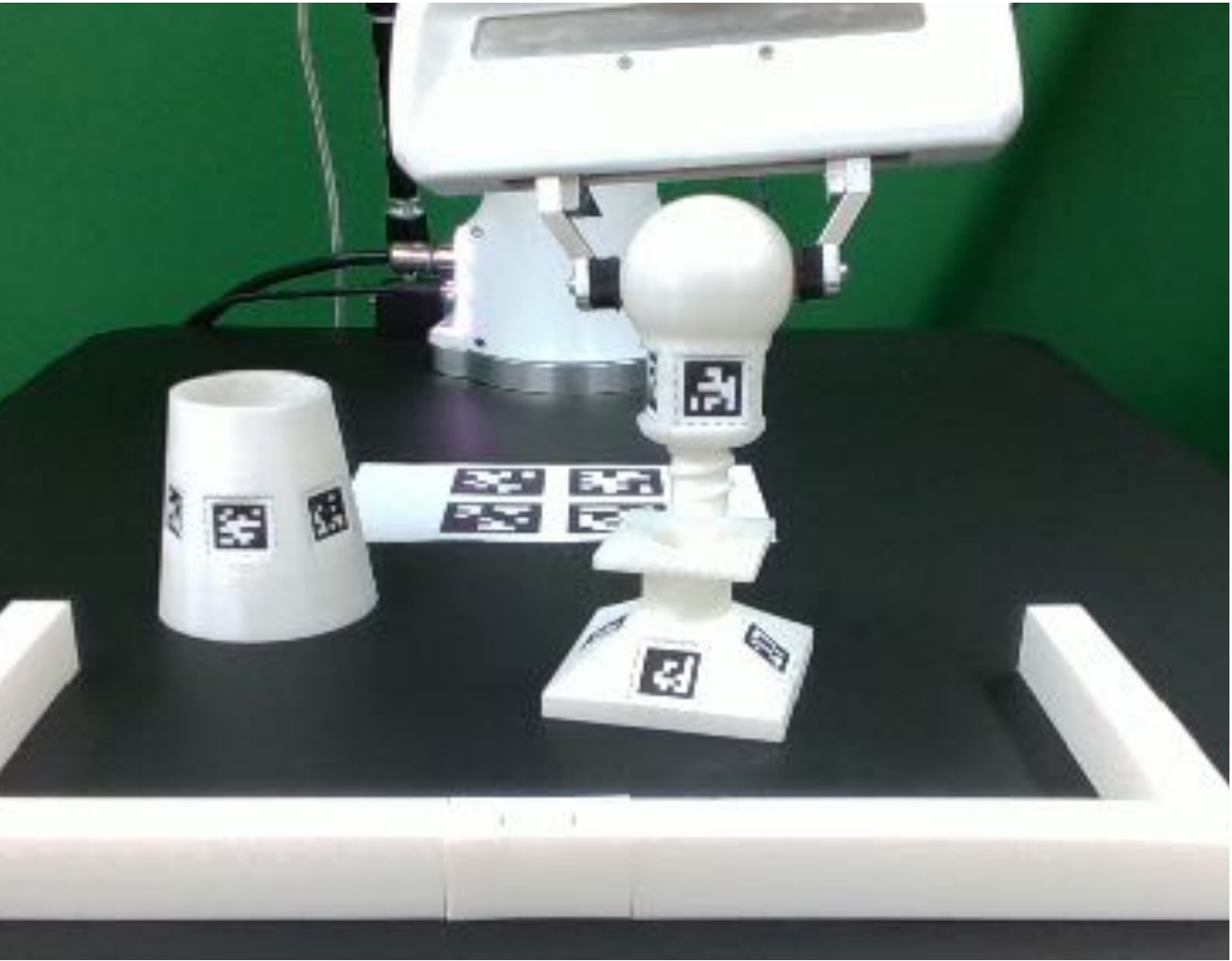}
        \caption{Front camera}
        \vspace{0.6em}
        \includegraphics[width=\linewidth]{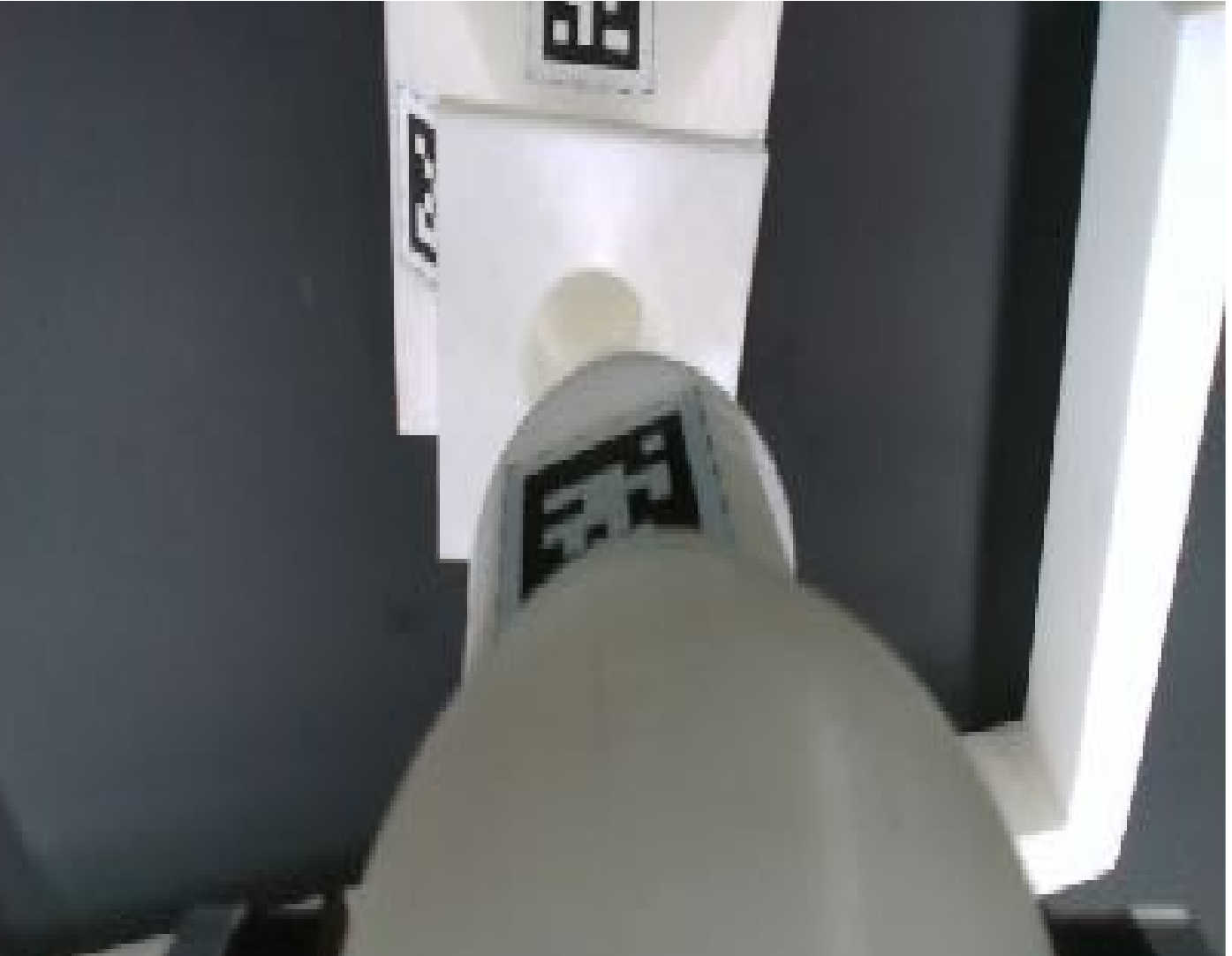}
        \caption{Wrist camera}
    \end{subfigure}
    \caption{\textbf{Real-world furniture assembly environment.} Our reproducible robot system (a) and visual observations from the front-view camera (b) and wrist camera (c).}
    \label{fig:environment}
    \vspace{-1em}
\end{figure}

\begin{figure}[t]
    \centering
    \captionsetup[subfigure]{labelformat=empty}
    \begin{subfigure}[t]{0.24\linewidth}
        \includegraphics[width=\linewidth]{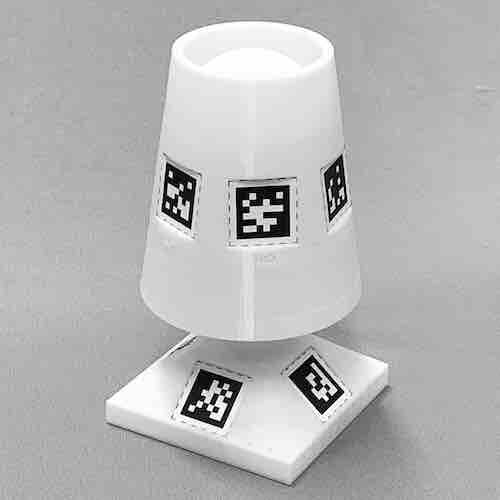}
        \vspace{-1.5em}
        \caption{\texttt{lamp}}
        \label{fig:furniture:lamp}
    \end{subfigure}
    \hfill
    \begin{subfigure}[t]{0.24\linewidth}
        \includegraphics[width=\linewidth]{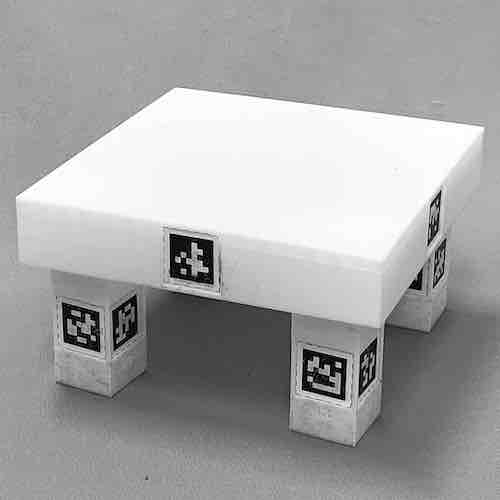}
        \vspace{-1.5em}
        \caption{\texttt{square\_table}}
        \label{fig:furniture:square_table}
    \end{subfigure}
    \hfill
    \begin{subfigure}[t]{0.24\linewidth}
        \includegraphics[width=\linewidth]{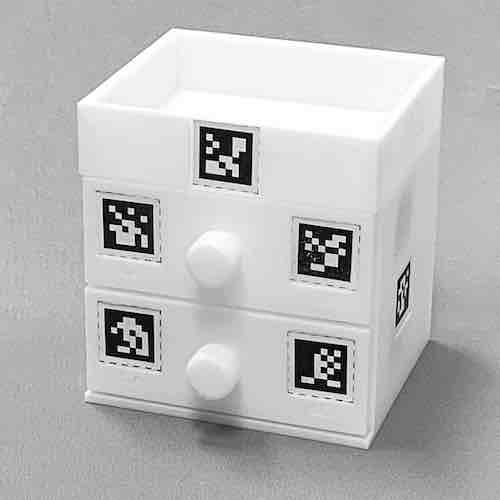}
        \vspace{-1.5em}
        \caption{\texttt{drawer}}
        \label{fig:furniture:drawer}
    \end{subfigure}
    \hfill
    \begin{subfigure}[t]{0.24\linewidth}
        \includegraphics[width=\linewidth]{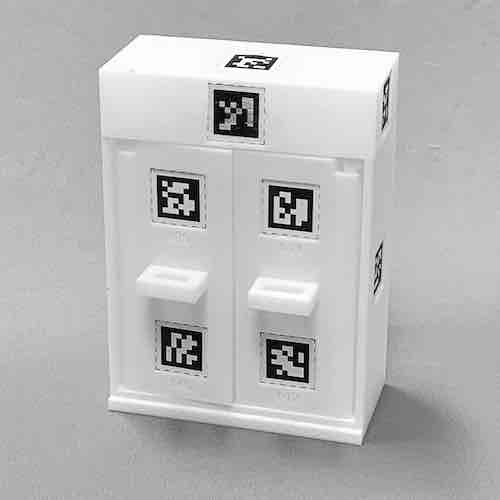}
        \vspace{-1.5em}
        \caption{\texttt{cabinet}}
        \label{fig:furniture:cabinet}
    \end{subfigure}
    \hfill
    \begin{subfigure}[t]{0.24\linewidth}
        \includegraphics[width=\linewidth]{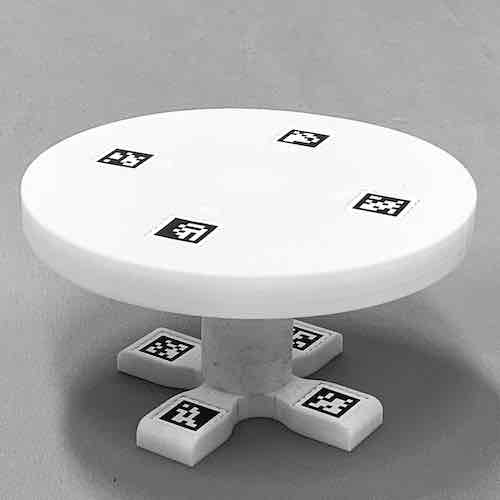}
        \vspace{-1.5em}
        \caption{\texttt{round\_table}}
        \label{fig:furniture:round_table}
    \end{subfigure}
    \hfill
    \begin{subfigure}[t]{0.24\linewidth}
        \includegraphics[width=\linewidth]{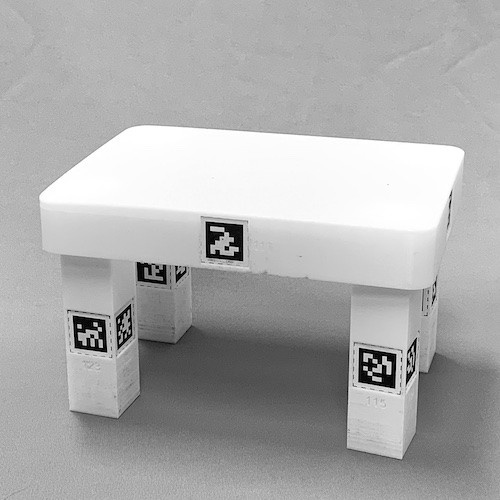}
        \vspace{-1.5em}
        \caption{\texttt{desk}}
        \label{fig:furniture:desk}
    \end{subfigure}
    \hfill
    \begin{subfigure}[t]{0.24\linewidth}
        \includegraphics[width=\linewidth]{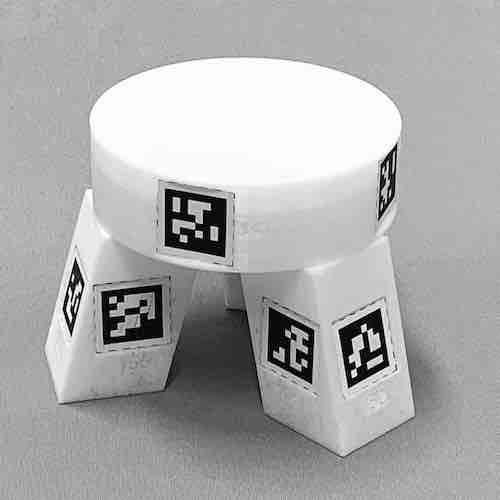}
        \vspace{-1.5em}
        \caption{\texttt{stool}}
        \label{fig:furniture:stool}
    \end{subfigure}
    \hfill
    \begin{subfigure}[t]{0.24\linewidth}
        \includegraphics[width=\linewidth]{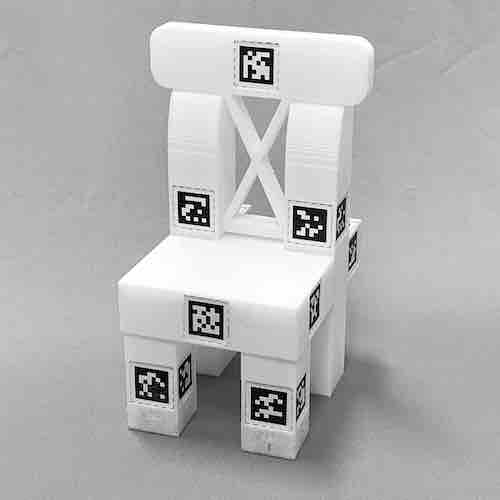}
        \vspace{-1.5em}
        \caption{\texttt{chair}}
        \label{fig:furniture:chair}
    \end{subfigure}
    \caption{\textbf{3D printed furniture models.} Each furniture is designed inspired by IKEA furniture. Due to the limitations imposed by using a single robotic arm, we modify some furniture pieces feasible to be assembled with one hand.}
    \label{fig:furniture}
    \vspace{-1em}
\end{figure}

To further robotics research toward solving people's everyday tasks, it is crucial to tackle challenges in more complex and longer-horizon tasks. This calls for reproducible benchmarks of long-horizon complex robotic manipulation tasks. To this end, we propose to focus on furniture assembly as the next milestone for complex, long-horizon robotic manipulation, and present FurnitureBench, a reproducible real-world furniture assembly benchmark, as shown in \Cref{fig:teaser,fig:environment,fig:furniture}.

Furniture assembly is a proper task suite to benchmark a difficult, long-horizon manipulation task through which many challenges in robotic manipulation must be addressed to solve. It has a hierarchical task structure, requiring reasoning over long horizons (e.g., deciding in which order and how to assemble various furniture pieces). In addition, connecting different furniture pieces together requires complex path planning and dexterous manipulation skills (e.g., robust grasping, accurate alignment of attachment points, and deliberate force control for inserting and screwing).

FurnitureBench is a \textit{reproducible} and \textit{easy-to-use} benchmark that can evaluate such challenges in furniture assembly. It aims to enable any lab to easily set up a standardized robotic environment and evaluate the capability of algorithms on realistic complex long-horizon tasks. 
We believe that FurnitureBench will help robot manipulation researchers identify challenges in solving complex long-horizon tasks, easily compare the performance of their approaches over prior work, and eventually solve diverse complex long-horizon real-world tasks.

The main contributions of this paper are as follows:
\begin{itemize}
  \item We introduce FurnitureBench, a \textbf{real-world} furniture assembly benchmark, which allows robotics researchers to investigate RL, IL, and TAMP algorithms on a \textit{realistic and complex} task beyond pick-and-place (\Cref{sec:task,sec:benchmark}, \Cref{fig:environment}).
  \item We ensure \textbf{reproducibility} by incorporating $3$D printable furniture pieces (\Cref{fig:furniture}) and providing comprehensive environment setup and evaluation guides (\Cref{sec:benchmark:real-world}).
  \item We collected \textbf{$\mathbf{200}$+ hours of large teleoperation demonstrations} that reduce barriers to real-world experiments on offline RL and IL algorithms (\Cref{sec:benchmark:easy-to-use}). 
  \item We evaluate diverse subtasks required for furniture assembly, e.g., grasping, inserting, and screwing skills, in our \textbf{single-skill benchmark} (\Cref{sec:experiments:skill}), which identifies challenges in learning ``inserting'' and ``screwing'' skills. 
  \item In the \textbf{full-assembly benchmark}, we evaluate the full furniture assembly tasks, where IL and offline RL methods achieve up to $2$ out of $12$ subtasks on average (\Cref{sec:experiments:full-task}).
  \item We develop FurnitureSim, a \textbf{simulator} of our real-world benchmark, which can boost fast iteration of experiments for new methods (\Cref{sec:benchmark:simulation,sec:experiments:simulation}).
\end{itemize}

\begin{table*}[t]
    \centering
    \resizebox{\linewidth}{!}{
    \begin{tabular}{|c|c|ccccccccccc|}
        \rot{} & 
        \rot{} & 
        \rot{\textbf{FurnitureBench (Ours)}} & 
        \rot{NIST Assembly Task Boards~\citep{kimble2020benchmarking}} & 
        \rot{RGB-Stacking~\citep{lee2021beyond}} & 
        \rot{REPLAB~\citep{yang2019replab}} &
        \rot{DGBench~\citep{burgess2022dgbench}} & 
        \rot{GRASPA 1.0~\citep{bottarel2020graspa}} & 
        \rot{ROBEL D’Claw~\citep{ahn2020robel}} & 
        \rot{In-Hand Manipulation~\citep{8950086}} & 
        \rot{Bimanual Cloth Manipulation~\citep{8957044}} & 
        \rot{Bimanual Manipulation~\citep{8989777}} & 
        \rot{Real Robot Challenge~\citep{bauer2021real, pmlr-v176-bauer22a}}
        \\ \hline%
        \multirow{2}*{{Reproducibility}} 
        & 
        \cellcolor[rgb]{.8,.8,.8} Robot and workspace & 
        \cellcolor[rgb]{.8,.8,.8} \textcolor{indiagreen}\faCheck & 
        \cellcolor[rgb]{.8,.8,.8} \textcolor{red}\faTimes & 
        \cellcolor[rgb]{.8,.8,.8} \textcolor{indiagreen}\faCheck & 
        \cellcolor[rgb]{.8,.8,.8} \textcolor{indiagreen}\faCheck &
        \cellcolor[rgb]{.8,.8,.8} \textcolor{indiagreen}\faCheck &
        \cellcolor[rgb]{.8,.8,.8} \textcolor{red}\faTimes & 
        \cellcolor[rgb]{.8,.8,.8} \textcolor{indiagreen}\faCheck & 
        \cellcolor[rgb]{.8,.8,.8} \textcolor{red}\faTimes & 
        \cellcolor[rgb]{.8,.8,.8} \textcolor{red}\faTimes & 
        \cellcolor[rgb]{.8,.8,.8} \textcolor{red}\faTimes & 
        \cellcolor[rgb]{.8,.8,.8} \textcolor{indiagreen}\faCheck
        \\
        & 
        \begin{tabular}{@{}c@{}}Availability of objects\end{tabular} & 
        \textcolor{indiagreen}\faCheck &
        \textcolor{indiagreen}\faCheck &
        \textcolor{indiagreen}\faCheck &
        \textcolor{indiagreen}\faCheck &
        \textcolor{indiagreen}\faCheck &
        \textcolor{indiagreen}\faCheck &
        \textcolor{indiagreen}\faCheck &
        \textcolor{indiagreen}\faCheck &
        \textcolor{indiagreen}\faCheck &
        \textcolor{indiagreen}\faCheck &
        \textcolor{red}\faTimes
        
        \\ \cline{1-2}
        \multirow{1}*{{Long horizon}} & 
        \cellcolor[rgb]{.8,.8,.8} \begin{tabular}{@{}c@{}} Task horizon\footnotemark[1] \end{tabular} & 
        \cellcolor[rgb]{.8,.8,.8}  $2300$ steps, $230$ sec  & 
        \cellcolor[rgb]{.8,.8,.8}  - & 
        \cellcolor[rgb]{.8,.8,.8}  $400$ steps & 
        \cellcolor[rgb]{.8,.8,.8}  - & 
        \cellcolor[rgb]{.8,.8,.8}  - & 
        \cellcolor[rgb]{.8,.8,.8}  - & 
        \cellcolor[rgb]{.8,.8,.8}  $80$ steps, $8$ sec &  
        \cellcolor[rgb]{.8,.8,.8}  $2$ sec & 
        \cellcolor[rgb]{.8,.8,.8}  $20$-$140$ sec & 
        \cellcolor[rgb]{.8,.8,.8}  $60$ sec & 
        \cellcolor[rgb]{.8,.8,.8}  $120$ sec       
        \\ \cline{1-2}
        \multirow{2}*{{Complexity}}  & 
         \begin{tabular}{@{}c@{}}DoF\end{tabular} &
         \begin{tabular}{@{}c@{}}6D\end{tabular} & 
         -  & 
         $4$D &
         $6$D &
         $3$D & 
         -  & 
         $6$D &
         -  & 
         -  & 
         $6$D &
         $6$D
        \\
        &
        \cellcolor[rgb]{.8,.8,.8}\begin{tabular}{@{}c@{}} Skills\footnotemark[2] \end{tabular} & 
        \cellcolor[rgb]{.8,.8,.8} PnP, P, F, I, S & 
        \cellcolor[rgb]{.8,.8,.8} PnP, I, S & 
        \cellcolor[rgb]{.8,.8,.8} PnP, F &
        \cellcolor[rgb]{.8,.8,.8} G, R &
        \cellcolor[rgb]{.8,.8,.8} DG &
        \cellcolor[rgb]{.8,.8,.8} G, R& 
        \cellcolor[rgb]{.8,.8,.8} PO, S& 
        \cellcolor[rgb]{.8,.8,.8} IR &
        \cellcolor[rgb]{.8,.8,.8} PnP, F &
        \cellcolor[rgb]{.8,.8,.8} H, I, B &
        \cellcolor[rgb]{.8,.8,.8} P, L, RO 
        
        \\ \cline{1-2}
        \multirow{4}*{{Diversity}}  & 
         \begin{tabular}{@{}c@{}}Number of tasks\end{tabular} &
         $8$ & 
         $4$ & 
         $2$ &
         $2$ &
         $2$ & 
         $3$ & 
         $3$ &
         $10$ & 
         $3$ & 
         $2$ &
         $2$ \\
        &
        \cellcolor[rgb]{.8,.8,.8}\begin{tabular}{@{}c@{}} Different levels of\\ evaluation per task \end{tabular} & 
        \cellcolor[rgb]{.8,.8,.8} $3$ & 
        \cellcolor[rgb]{.8,.8,.8} $1$ & 
        \cellcolor[rgb]{.8,.8,.8} $5$ &
        \cellcolor[rgb]{.8,.8,.8} $1$ &
        \cellcolor[rgb]{.8,.8,.8} $1$ &
        \cellcolor[rgb]{.8,.8,.8} $3$ & 
        \cellcolor[rgb]{.8,.8,.8} $2$-$3$ & 
        \cellcolor[rgb]{.8,.8,.8} $3$ &
        \cellcolor[rgb]{.8,.8,.8} $4$-$5$ &
        \cellcolor[rgb]{.8,.8,.8} $1$ &
        \cellcolor[rgb]{.8,.8,.8} $4$ 
        \\
        & 
        \cellcolor[rgb]{1,1,1}\begin{tabular}{@{}c@{}}Object categories\footnotemark[3]\end{tabular} &
        \cellcolor[rgb]{1,1,1} R, A & 
        \cellcolor[rgb]{1,1,1} R, D & 
        \cellcolor[rgb]{1,1,1} R & 
        \cellcolor[rgb]{1,1,1} R, A, D & 
        \cellcolor[rgb]{1,1,1} R & 
        \cellcolor[rgb]{1,1,1} R & 
        \cellcolor[rgb]{1,1,1} R & 
        \cellcolor[rgb]{1,1,1} R & 
        \cellcolor[rgb]{1,1,1} D & 
        \cellcolor[rgb]{1,1,1} R, D & 
        \cellcolor[rgb]{1,1,1} R
        
        \\ \cline{1-2}
        \multirow{2}*{{Demonstrations}}  & 
        \cellcolor[rgb]{.8,.8,.8} \begin{tabular}{@{}c@{}}Expert teleoperation\end{tabular} & 
        \cellcolor[rgb]{.8,.8,.8} \textcolor{indiagreen}\faCheck & 
        \cellcolor[rgb]{.8,.8,.8} \textcolor{red}\faTimes & 
        \cellcolor[rgb]{.8,.8,.8} \textcolor{red}\faTimes & 
        \cellcolor[rgb]{.8,.8,.8} \textcolor{red}\faTimes &
        \cellcolor[rgb]{.8,.8,.8} \textcolor{red}\faTimes &
        \cellcolor[rgb]{.8,.8,.8} \textcolor{red}\faTimes & 
        \cellcolor[rgb]{.8,.8,.8} \textcolor{red}\faTimes & 
        \cellcolor[rgb]{.8,.8,.8} \textcolor{red}\faTimes & 
        \cellcolor[rgb]{.8,.8,.8} \textcolor{red}\faTimes & 
        \cellcolor[rgb]{.8,.8,.8} \textcolor{indiagreen}\faCheck & 
        \cellcolor[rgb]{.8,.8,.8} \textcolor{red}\faTimes 
        \\
        & 
        \begin{tabular}{@{}c@{}}Dataset size\end{tabular} & 
        $200$hrs / $5$k ep. &
        - &
        - &
        $92$k ep. &
        - &
        - &
        - &
        - &
        - &
        $1200$ ep. &
        $10$k ep.  
        
        \\ \cline{1-2}
        \multirow{1}*{{Simulation}}  & 
        \cellcolor[rgb]{.8,.8,.8} \begin{tabular}{@{}c@{}}Simulator\end{tabular} & 
        \cellcolor[rgb]{.8,.8,.8} \textcolor{indiagreen}\faCheck & 
        \cellcolor[rgb]{.8,.8,.8} \textcolor{red}\faTimes & 
        \cellcolor[rgb]{.8,.8,.8} \textcolor{indiagreen}\faCheck & 
        \cellcolor[rgb]{.8,.8,.8} \textcolor{red}\faTimes &
        \cellcolor[rgb]{.8,.8,.8} \textcolor{red}\faTimes &
        \cellcolor[rgb]{.8,.8,.8} \textcolor{red}\faTimes & 
        \cellcolor[rgb]{.8,.8,.8} \textcolor{indiagreen}\faCheck & 
        \cellcolor[rgb]{.8,.8,.8} \textcolor{indiagreen}\faCheck & 
        \cellcolor[rgb]{.8,.8,.8} \textcolor{red}\faTimes & 
        \cellcolor[rgb]{.8,.8,.8} \textcolor{indiagreen}\faCheck & 
        \cellcolor[rgb]{.8,.8,.8} \textcolor{indiagreen}\faCheck
        \\\cline{1-13}
    \end{tabular}
    }
    \begin{flushleft}\tiny
    \textsuperscript{1} Average or maximum episode lengths or expert execution time \quad 
    \textsuperscript{2}PnP: Pick-and-place / P: Push / F: Flip / I: Insert / S: Screw / G: Grasp / R: Reach / DG: Dynamic grasp / PO: Pose / IR: In-hand re-orientation / H: Hold / B: Bend / L: Lift / RO: Rotate  \quad 
    \textsuperscript{3}R: Rigid objects, A: Articulated objects, D: Deformable objects  
    \end{flushleft}
    \vspace{-0.5em}
    \caption{
      \textbf{Comparison of real-world robotic manipulation benchmarks.} Our furniture assembly benchmark is reproducible, testing a complex, long-horizon task, and providing a large dataset and a simulator for easy benchmarking.
    }  
    \label{tab:benchmarks}
    \vspace{-1em}
\end{table*}

\section{Related Work}
\label{sec:related_work}

\textbf{Simulated manipulation benchmarks.}\quad
Deep reinforcement learning (RL) has made rapid progress with the advent of standardized, simulated benchmarks, such as Atari~\citep{bellemare13arcade} and continuous control~\citep{brockman2016openai, tassa2018deepmind} benchmarks. 
In robotic manipulation, most existing simulated environments have been focused on tasks like picking and placing~\citep{brockman2016openai, lee2019silo, james2019rlbench, zhu2020robosuite, yu2019meta}, in-hand manipulation~\citep{andrychowicz2020learning, rajeswaran2018learning}, door opening~\citep{urakami2019doorgym}, peg insertion~\citep{chebotar2019closing, yamada2020mopa}, and screwing~\citep{narang2022factory}. However, these tasks, which consist of lifting, stacking, and picking and placing, are limited to short-horizon primitive skills. 

Composite manipulation tasks, such as block stacking~\citep{duan2017one-shot}, ball serving~\citep{lee2019composing}, kitchen tasks~\citep{gupta2019relay, pertsch2020spirl}, and table-top manipulation~\citep{mees2022calvin}, have been introduced but are still limited to small variations in shapes and physical properties of objects. 
BEHAVIOR~\citep{srivastava2021behavior, li2022behavior} and Habitat~\citep{savva2019habitat, szot2021habitat} present simulated benchmarks including diverse long-horizon mobile manipulation tasks while focusing on high-level planning and abstracting complex low-level control problems. In contrast, the IKEA furniture assembly environment~\citep{lee2021ikea} simulates a complex manipulation task, \textit{furniture assembly}, to evaluate long-term planning and generalization of learned skills across various shapes of objects. 
Although simulated benchmarks are easily accessible and useful for quickly verifying ideas, they do not ensure that algorithms can directly work on real robots due to the complexity and stochastic nature of the real environment. 

\textbf{Real-world manipulation benchmarks.}\quad
Real-world robot learning involves many challenges that are not considered in simulated environments, such as collision, noisy and delayed sensory inputs, automatic environment resetting, and providing reward signals. As summarized in \Cref{tab:benchmarks}, recent robotic manipulation benchmarks have introduced reproducible benchmarking environments with an affordable robot and easy-to-make cage~\citep{yang2019replab, ahn2020robel}, with $3$D printed objects~\citep{lee2021beyond}, and by evaluating in cloud~\citep{bauer2021real, pmlr-v176-bauer22a}. However, these works focus on much simpler and short-horizon tasks (e.g., stacking blocks, re-orienting objects) than furniture assembly.

\textbf{Robotic assembly benchmarks.}\quad
Robotic furniture assembly has been studied in instrumented and constrained settings~\citep{niekum2013incremental,knepper2013ikeabot, suarez2018can}. Yet, these works are not suitable for benchmarking as they use their own furniture models, robots, and software. 
NIST Assembly Task Board~\citep{kimble2020benchmarking} provides a suite of complex manufacturing tasks and evaluation metrics, which is most similar to ours. However, it mainly focuses on providing task specifications and lacks a standard experimental setup, namely a robot, hardware, observation, and action. This motivates us to implement a \emph{standardized} real-world manipulation benchmark consisting of long-horizon complex tasks with careful designs for reproducibility and usability.
\section{Furniture Assembly: Long-Horizon~Complex~Manipulation~Task}
\label{sec:task}

\begin{figure*}[t]
    \centering
    \includegraphics[width=\linewidth]{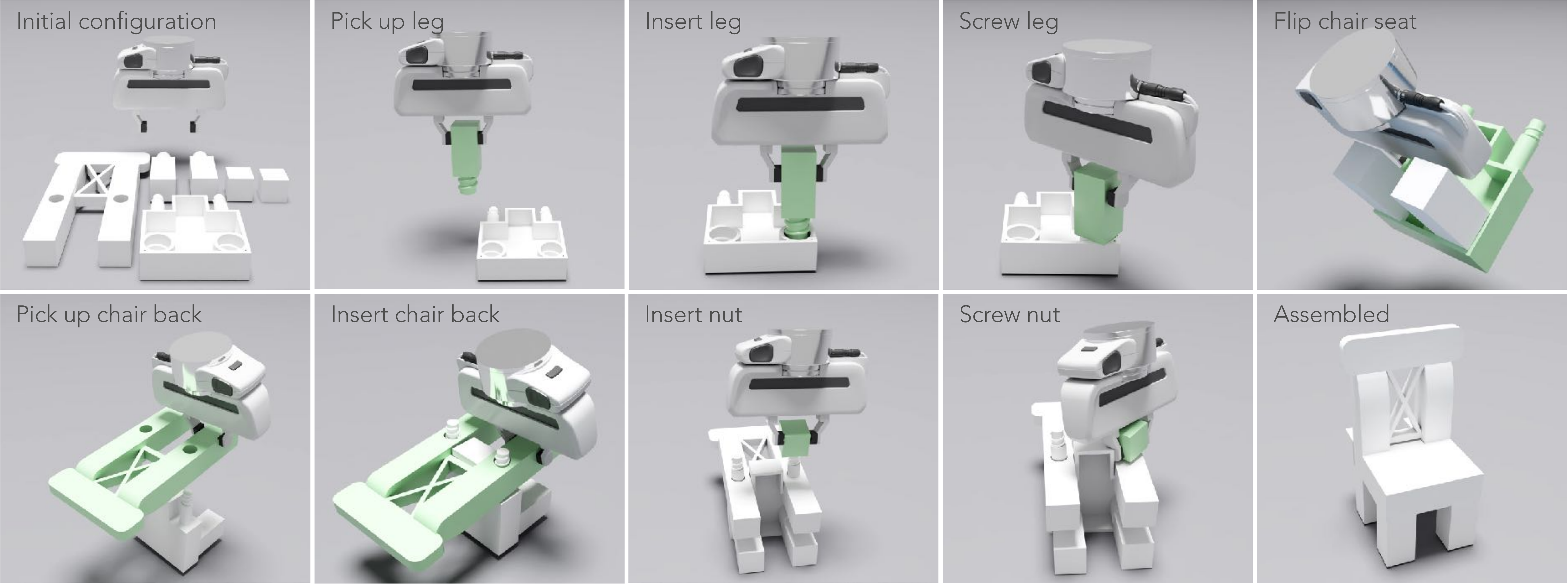}
    \caption{
        \textbf{Overview of a furniture assembly task, \texttt{chair}.} Furniture assembly requires many dexterous skills, including grasping, re-orienting, inserting, and screwing. The part in contact with the robot gripper is highlighted in green and some repetitive steps, e.g., screwing the second leg and nut, are omitted.
    }
    \label{fig:skills}
    \vspace{-1em}
\end{figure*}

To advance robotics research from simple and artificial tasks to complex and realistic tasks, we need a benchmark that helps researchers identify challenges in solving such complex, long-horizon real-world tasks, and allows them to easily evaluate and compare various approaches. In this paper, we propose to benchmark the task of furniture assembly, which is a meaningful, long-horizon, complex, real-world task.

Even for humans, furniture assembly is not a simple task. We need to understand how pieces are assembled into the final configuration and make a plan to assemble them in the right order. Furniture assembly also involves many complex manipulation skills, such as the accurate alignment of two parts and sophisticated control to screw them together. Therefore, furniture assembly is a comprehensive robotic manipulation task~\citep{niekum2013incremental, knepper2013ikeabot, suarez2018can} requiring high-level long-term planning, sophisticated control, and reliable $3$D perception, making it a suitable benchmark for robotic manipulation research.

\textbf{Long-horizon tasks.}\quad 
The furniture assembly task is long-horizon and hierarchical, as illustrated in \Cref{fig:skills}. Furniture assembly can be accomplished by repeating $1$-step assembly as follows: (1)~selecting two parts to be assembled, (2)~grasping one part\footnote{This benchmark focuses on single-arm manipulation, which is the most common robot arm setup in research labs. Using one robotic arm results in the limited dexterity to handle multiple objects simultaneously or a large object. We overcome this issue by adding an obstacle fixed on the table (see \Cref{fig:environment}), which can be used to hold an object, and slightly modifying $3$D furniture models suited for one-hand manipulation. We leave extending this work to multi-arm and mobile manipulation for future work.}, (3)~moving the part towards the other, (4)~aligning the attachable points (e.g., a screw and hole), and (5)~firmly attaching them via screwing or insertion, until all parts are assembled. Thus, furniture assembly has a much longer horizon on average ($60$-$230$~sec, $600$-$2300$~low-level steps) than prior manipulation benchmarks (e.g., $15$-$120$~sec for Roboturk~\citep{mandlekar2018roboturk}).

\textbf{Diverse dexterous skills.}\quad 
This seemingly simple procedure requires diverse, challenging manipulation skills, as illustrated in \myfigref{fig:skills}. First, \textbf{grasping} a furniture piece for assembly requires a specific grasping pose for the following sub-tasks. Attaching two parts sometimes requires re-orienting the furniture pieces both in hand and on the table through picking-and-placing or pushing. As the task proceeds, maneuvering an arm and manipulating furniture parts need more thorough path planning to \textbf{avoid collisions} with irrelevant parts. For example, a robot needs to avoid colliding with already attached table legs; otherwise, it can move irrelevant furniture pieces around. Lastly, \textbf{inserting} and \textbf{screwing}, by themselves, are challenging manipulation skills. Inserting and screwing skills require precise alignment between a screw and a hole, then repeated rotation of the screw while gently pressing it. Moreover, due to the limited wrist joint rotation, screwing (i.e., rotating a part for \ang{540}) requires at least five \ang{90}-screwing and re-grasping motions. 
\section{FurnitureBench: Reproducible Real-World Furniture Assembly Benchmark}
\label{sec:benchmark}

While bringing real-world complexity into the robot manipulation community, our benchmark is also designed to be reproducible and easy-to-use. 
In this section, we first describe our \textbf{reproducible} real-robot experimental setup, including the robotic system design and furniture models. For \textbf{ease of use}, we provide a plug-and-play robot control stack and a large teleoperated demonstration dataset. We then introduce a simulated environment of our real-world benchmark for easy and fast verification of various algorithms.

\subsection{Reproducible Real-World Benchmark Environment}
\label{sec:benchmark:real-world}

\textbf{Reproducible system design.}\quad
To make the real-robot environment easy to reproduce, we opt for widely used products across the world and $3$D-printing objects. As shown in \Cref{fig:environment:setup}, our real-world system mainly consists of one $7$-DoF Franka Emika Panda robot arm with a default parallel gripper and three Intel RealSense D435 RGB-D cameras, which are widely used in robotics community. As a workspace, we propose to use the black IKEA TOMMARYD table, which is available in a wide range of countries, and attach $3$D-printed obstacles to the table. The green photography backdrop is used to provide a consistent background and a single light is used to vary lighting conditions in color temperature (\SI{4600}{\kelvin}-\SI{6000}{\kelvin}) and brightness ($\leq$\SI{4000}{\lumen}). We provide detailed system setup instructions (\Cref{sec:instruction}) and software tools, which allow a user to build a new environment with nearly the same configuration based on furniture and table poses estimated with AprilTag~\citep{olson2011apriltag}. 

\textbf{Reproducible furniture models.}\quad
For reproducibility, we use $3$D-printed furniture models (\cref{fig:furniture}); thus, any lab can make the same furniture parts by simply downloading and $3$D-printing our model files.\footnote{Our furniture models require a $3$D printer with the capacity larger than \SI{25}{\cm}$\times$\SI{25}{\cm}$\times$\SI{15}{\cm} and printing one furniture model takes up to $24$ hours.} Our benchmark features $8$ different furniture models, each  modeled after an existing piece of IKEA furniture, as shown in \Cref{fig:ikea_furniture}. These $8$ furniture models are selected to introduce their own challenges and interactions (e.g., \texttt{drawer} requires precise alignment between the rail and drawer; already assembled legs in \texttt{desk} can interrupt assembly of other legs). More details about the furniture models and their assembly procedures are described in \Cref{sec:furniture_models}.

\begin{figure}[t]
    \centering
    \includegraphics[width=0.9\linewidth]{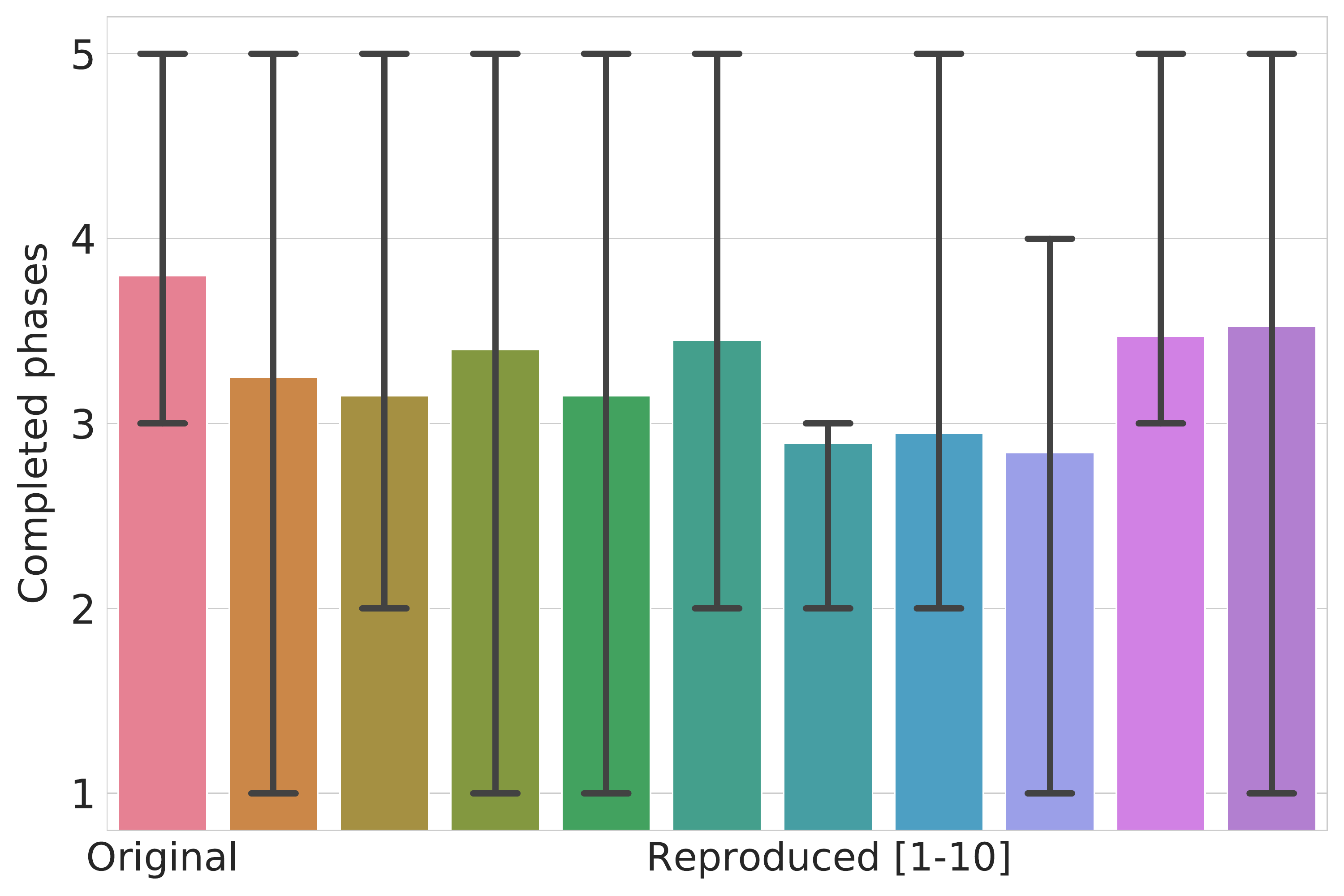}
    \caption{\textbf{Reproducibility analysis.} We evaluate the number of completed subtasks (i.e., phases) of an IQL policy on the \texttt{one\_leg} assembly task in $10$ newly built environments under the low randomness level. The new environments achieve on average $84$\% of the performance in the original environment, which ensures the reproducibility of our benchmark.} 
    \label{fig:reproducibility_performance}
    \vspace{-1em}
\end{figure}

\textbf{Reproducibility analysis.}\quad
To verify the reproducibility of our benchmark, we conduct the reproducibility test by asking $10$ participants to set up an environment from scratch following the instruction in \Cref{sec:instruction}. Each participant independently builds the system, and this takes about $3$ hours. We then evaluate a fully-trained IQL-R3M policy on the \texttt{one\_leg} task in these $10$ new environments. \Cref{fig:reproducibility_performance} shows the average number of completed phases (i.e., subtasks) over the $10$ new environments are consistently within $75$-$93$\% of the performance in the original (best-performing) environment. This ensures that our benchmarking result can be reproduced in a new environment built by a new subject with slightly different camera viewpoints, backgrounds, and lighting conditions. Although the results are reproducible, we found that the pre-trained policy is sensitive to the changes in the background and viewpoints of the front and wrist cameras. Therefore, we highlight how to minimize the errors in camera viewpoints in our step-by-step instruction (\Cref{sec:instruction}). The experimental details can be found in \Cref{sec:experiment_details:reproducibility}.

\textbf{Observation and robot control.}\quad
The observation space for the environment consists of front-view and wrist camera inputs (\Cref{fig:environment}) and a proprioceptive robot state. Two Intel RealSense D435 cameras are used to obtain the RGB inputs. The front-view image ($1280\times720$) is first downsampled to $455\times256$ and then center-cropped to $224\times224$. The wrist camera image is directly downsampled to $224\times224$ for the wider view. Note that estimated poses of furniture parts with AprilTags~\citep{olson2011apriltag} are \textbf{not} used as an observation but used for making the benchmark reproducible. The proprioceptive robot state includes the end-effector position, orientation (in quaternion), velocity, and gripper width. We use end-effector delta-pose control at \SI{10}{\hertz} implemented using Operational Space Control~\citep{khatib1987unified}.

\begin{table}[t]
\centering
\caption{\textbf{Statistics of the tasks and demonstrations.} Each task includes three different levels with respect to the randomness in initialization: low, medium, and high in order.}
\label{tab:data}
\resizebox{\linewidth}{!}{
\begin{tabular}{lccccc}
    \toprule
                 & Parts  & Phases     & Demos &  Avg. length  & Total hrs \\
    \midrule
    \multirow{3}*{{\texttt{lamp}}}  &  \multirow{3}*{$3$}    & \multirow{3}*{$7$}           & $150$       &  $594$                 &  $4.9$   \\
                                    &                      &                            & $150$       &  $598$                 &  $5.0$   \\
                                    &                      &                            & $50$        &  $768$                 &  $2.1$   \\
    \midrule
    \multirow{3}*{{\texttt{square\_table}}}  &  \multirow{3}*{$5$}    & \multirow{3}*{$16$} & $150$       &  $1689$                 &  $14.1$   \\
                                    &                      &                            & $150$       &  $1660$                 &  $13.8$   \\
                                    &                      &                            & $50$        &  $1682$                 &  $4.7$   \\
    \midrule
    \multirow{3}*{{\texttt{desk}}}  &  \multirow{3}*{$5$}    & \multirow{3}*{$16$}          & $100$       &  $1531$                 &  $8.5$   \\
                                    &                      &                            & $100$       &  $1914$                 &  $10.6$   \\
                                    &                      &                            & $50$        &  $1687$                 &  $4.7$   \\
    \midrule
    \multirow{3}*{{\texttt{drawer}}}  &  \multirow{3}*{$3$}    & \multirow{3}*{$8$}         & $250$       &  $571$                 &  $7.9$   \\
                                    &                      &                            & $250$       &  $520$                 &  $7.2$   \\
                                    &                      &                            & $50$        &  $781$                 &  $2.2$   \\
    \midrule
    \multirow{3}*{{\texttt{cabinet}}}  &  \multirow{3}*{$4$} & \multirow{3}*{$11$}          & $150$       &  $883$                  &  $7.4$   \\
                                    &                      &                            & $150$       &  $814$                  &  $6.8$   \\
                                    &                      &                            & $50$        &  $1166$                 &  $3.2$   \\
    \midrule
    \multirow{3}*{{\texttt{round\_table}}}  &  \multirow{3}*{$3$}    & \multirow{3}*{$8$}   & $100$       &  $847$                  &  $4.7$   \\
                                    &                      &                            & $100$       &  $867$                  &  $4.8$   \\
                                    &                      &                            & $50$        &  $1060$                 &  $2.9$   \\
    \midrule
    \multirow{3}*{{\texttt{stool}}} &  \multirow{3}*{$4$}    & \multirow{3}*{$11$}          & $100$       &  $1231$                 &  $6.8$   \\
                                    &                      &                            & $100$       &  $1419$                 &  $7.9$   \\
                                    &                      &                            & $50$        &  $1273$                &   $3.5$   \\
    \midrule
    \multirow{3}*{{\texttt{chair}}} &  \multirow{3}*{$6$}    & \multirow{3}*{$17$}           & $100$       &  $1817$                 &  $10.1$   \\
                                    &                      &                            & $100$       &  $2282$                 &  $12.7$   \\
                                    &                      &                            & $50$        &  $2066$                 &  $5.7$   \\
    \midrule
    \multirow{3}*{{\texttt{one\_leg}}}  &  \multirow{3}*{$2$}    & \multirow{3}*{$5$}       & $1000$      &  $374$                 &  $20.8$   \\
                                    &                      &                            & $1000$      &  $429$                 &  $23.8$   \\
                                    &                      &                            & $500$       &  $461$                 &  $12.8$   \\
    \midrule
    \textbf{Overall} & $2$ - $6$ & $5$ - $17$ & $5100$ & $374$ - $2282$ & $219.6$ \\
    \bottomrule
\end{tabular}
}
\end{table}

\subsection{Easy-to-Use Benchmark}
\label{sec:benchmark:easy-to-use}

Finally, to make our benchmark easy to use with as little human effort as possible, we provide a plug-and-play robot control software, a task initialization tool for evaluation, and a dataset of successful demonstrations for every assembly task.

\textbf{Robot control stack.}\quad
To lower the barrier to entry into a real-robot manipulation setup, our robot control stack is provided as a Docker image. The robot control stack is developed on top of a Python-based controller, Polymetis~\citep{Polymetis2021}, which is easy to install, use, and adapt.

\begin{figure}[t]
    \centering
    \begin{subfigure}[t]{0.485\linewidth}
        \includegraphics[width=\linewidth]{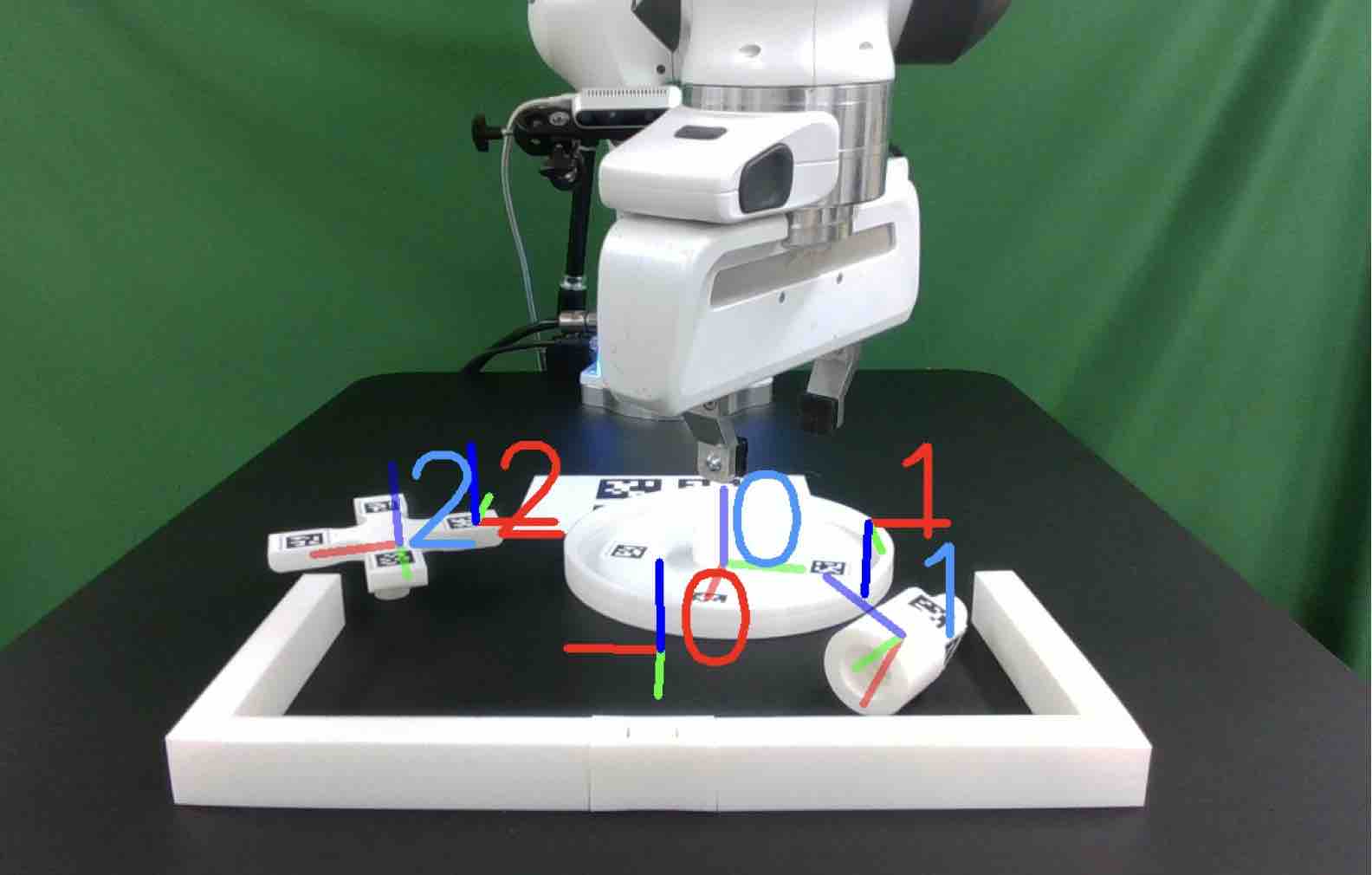}
        \caption{Front-view camera}
        \label{fig:initialization_tool:front_initial}
    \end{subfigure}
    \hfill
    \begin{subfigure}[t]{0.485\linewidth}
        \includegraphics[width=\linewidth]{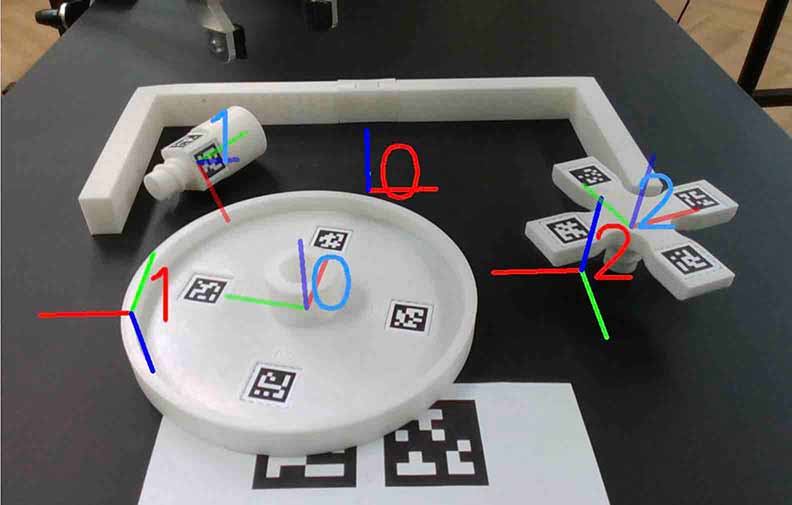}
        \caption{Rear camera}
        \label{fig:initialization_tool:rear_initial}
    \end{subfigure}
    \caption{
        \textbf{Task initialization GUI tool.} To evaluate our benchmarks with the proper distribution of initial poses of furniture parts, we provide a task initialization tool. Through two camera views in (a-b), a user requires to match the poses of furniture parts (transparent axis, blue index) with the target configuration sampled from a pre-defined distribution (solid axis, red index).
    }
    \label{fig:initialization_tool}
    \vspace{-1em}
\end{figure}

\textbf{Task initialization tool.}\quad
Initial configurations set by users without any guidance can be easily biased. This biased task initialization makes experimental results from different users not comparable since evaluation results can vary depending on initial state distributions. To evaluate algorithms under the same initial state distribution, we provide a task initialization GUI tool as shown in \Cref{fig:initialization_tool}. This tool samples an initial configuration from a pre-defined distribution and guides a user to match the sampled initial pose of each furniture part.

\textbf{Teleoperated demonstration dataset.}\quad
Our furniture assembly benchmark proposes the suite of challenging long-horizon manipulation tasks described above. Due to complex manipulation skills and the long task time horizon, learning these tasks from scratch may require millions of real-world interactions for the state-of-the-art RL methods. To make this benchmark more feasible, we collect a large demonstration dataset using teleoperation and evaluate IL and offline RL methods trained on this dataset.

We collected $219.6$ hours of successful demonstrations using an Oculus~Quest~2 controller and a keyboard. The VR controller makes general data collection easier and faster while a keyboard is used to rotate the wrist without moving the position of the gripper for fine-grained control, such as screwing. We summarize the statistics of the collected demonstrations in \Cref{tab:data}. For each 
model and task initialization level, $2$ to $24$ hours of demonstration data is collected. Each demonstration is around $300$-$3000$ steps long due to the long-horizon nature of the task.

\subsection{FurnitureSim: Simulated Environment}
\label{sec:benchmark:simulation}

\begin{figure}[t]
    \centering
    \begin{subfigure}[t]{0.325\linewidth}
        \includegraphics[width=\linewidth]{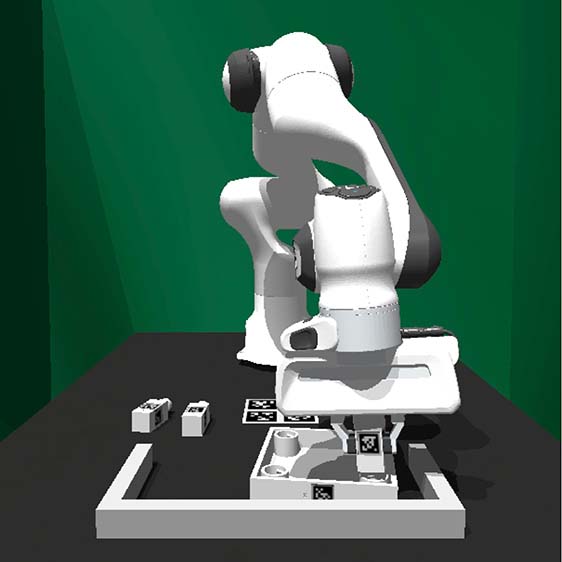}
        \caption{Fast rendering}
        \label{fig:simulator_screw}
    \end{subfigure}
    \hfill
    \begin{subfigure}[t]{0.325\linewidth}
        \includegraphics[width=\linewidth]{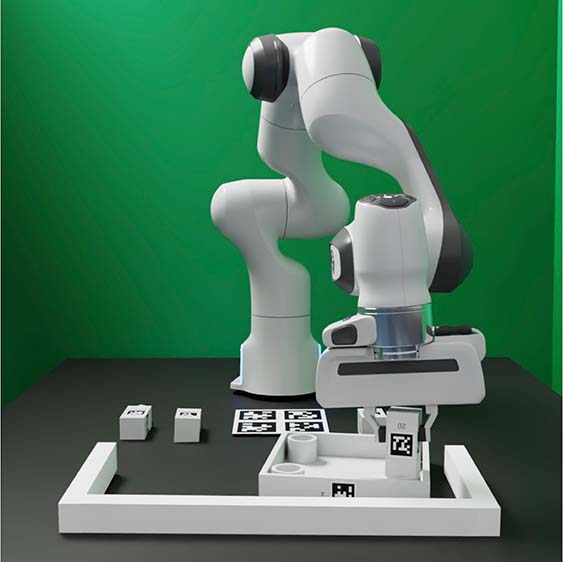}
        \caption{Ray tracing}
        \label{fig:rendering_screw}
    \end{subfigure}
    \hfill
    \begin{subfigure}[t]{0.325\linewidth}
        \includegraphics[width=\linewidth]{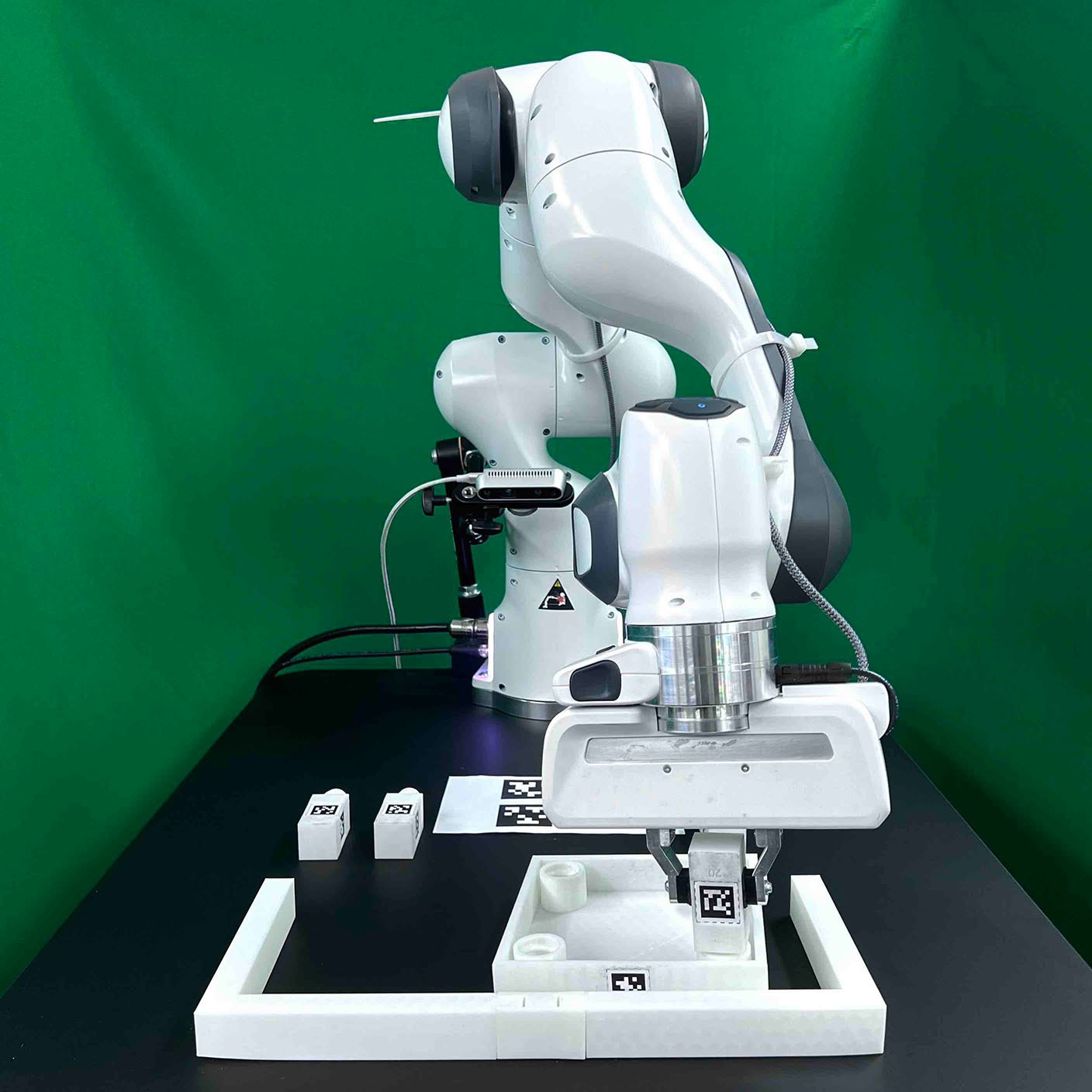}
        \caption{Real-world image}
        \label{fig:real_screw}
    \end{subfigure}
    \caption{
        \textbf{FurnitureSim: simulated version of FurnitureBench.} FurnitureSim provides realistic physics simulation and rendering: (a)~fast online rendering and (b)~photorealistic offline rendering with ray tracing similar to the real-world image (c).
    }
    \label{fig:simulator}
    \vspace{-1em}
\end{figure}

Slow and costly evaluation is one of the major drawbacks of real-robot benchmarks. Introducing a simulator can help researchers quickly understand the tasks and test their ideas. To this end, we provide FurnitureSim, a simulated version of FurnitureBench, as shown in \Cref{fig:simulator}. For realistic simulation, FurnitureSim is built upon Isaac Gym~\citep{makoviychuk2021isaac} and Factory~\citep{narang2022factory}, which provides a fast and accurate simulation of screws along with realistic rendering. To closely resemble the real-world environment, we use the same $3$D furniture models and robot controller used for the real-world environment.

\begin{figure}[b]
    \vspace{-1em}
    \centering
    \includegraphics[width=\linewidth]{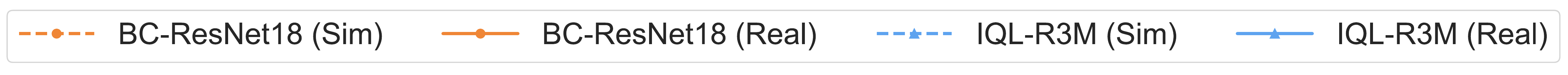} \\
    \includegraphics[width=0.485\linewidth]{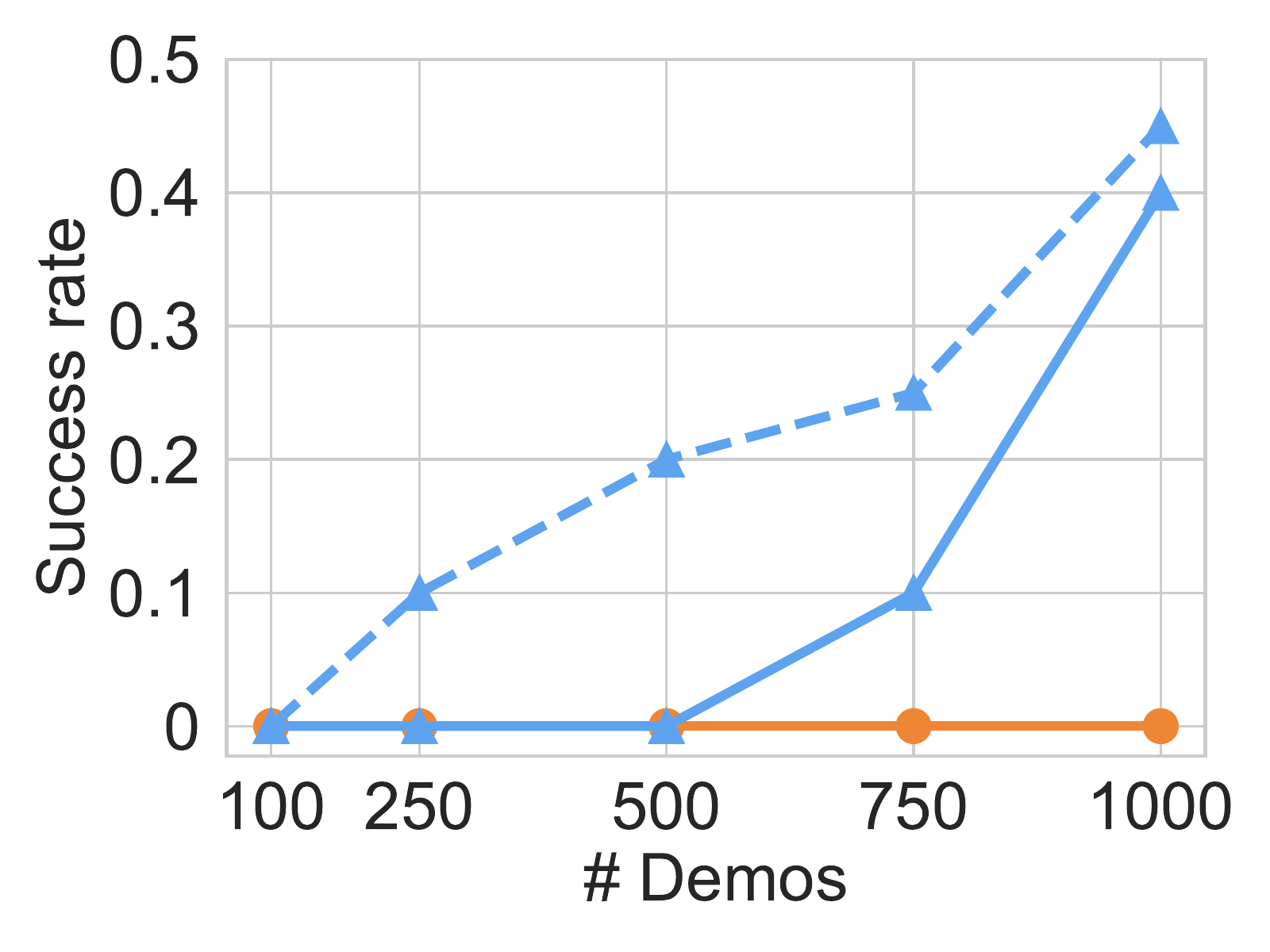}
    \includegraphics[width=0.485\linewidth]{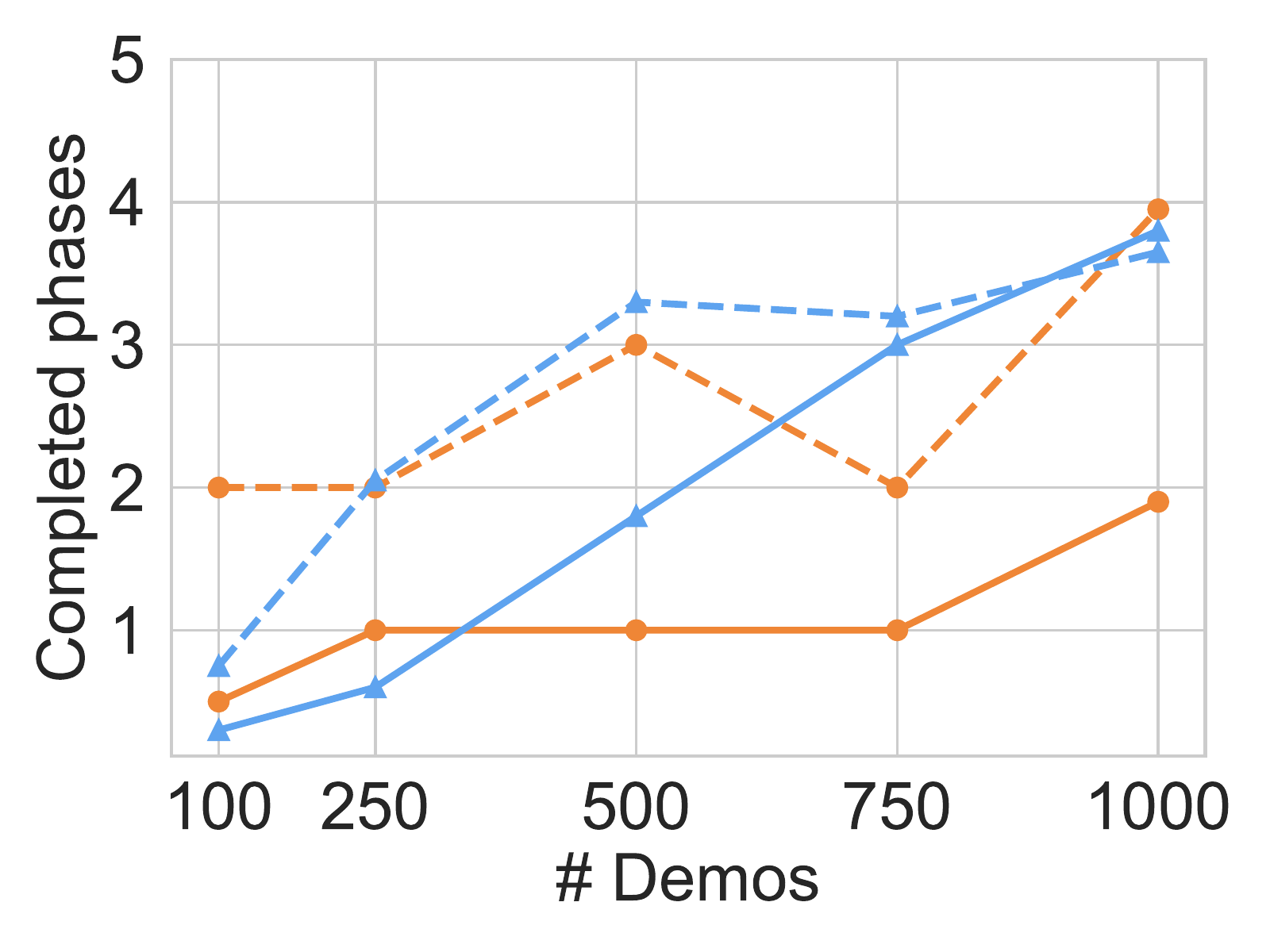}
    \caption{
        \textbf{Correlation between FurnitureBench and FurnitureSim.} We compare the performance of IL and offline RL methods with respect to the dataset size between simulation and the real world on the \texttt{one\_leg} assembly task under the low randomness level. We compare both the success rate (left) and the number of completed phases (right). The results in simulation and the real world show similar trends with respect to the data size as well as the choice of algorithm.
    }
    \label{fig:correlation_simulation}
\end{figure}

\textbf{Correlation between FurnitureBench and FurnitureSim.}\quad
Note that there still exist domain gaps between the real-world and simulated environments; yet, the simulator can be still useful for quickly verifying the correctness of algorithms and providing a proxy for real-world performance. 
In \Cref{fig:correlation_simulation}, we evaluate IL (BC~\citep{pomerleau1989alvinn} with the ResNet-18 encoder~\citep{he2016identity}) and offline RL (IQL~\citep{kostrikov2022offline} with R3M features~\citep{nair2022r3m}) in both the real world and simulation. The result clearly shows the positive correlation between the simulation and real-world performances in success rates. The performances in the simulation are generally better than the ones from the real world, which implies additional challenges in the real-world benchmark.
We leave closing simulation-to-real domain gaps as future work.

\section{Experimental Setup}
\label{sec:experimental_setup}

\textbf{Baselines.}\quad
We evaluate our benchmark with imitation learning (BC) and the state-of-the-art offline RL (IQL) methods. Please refer to \Cref{sec:training_details} for implementation details.
\begin{itemize}
    \item \textbf{BC} (Behavioral Cloning~\citep{pomerleau1989alvinn}) fits a policy to the demonstration state-action pairs $(\mathbf{s}, \mathbf{a})$ with supervised learning.
    \item \textbf{IQL} (Implicit Q Learning~\citep{kostrikov2022offline}) is the state-of-the-art offline RL, performing advantage-weighted BC with a value function trained with expectile regression loss. 
\end{itemize}

\textbf{Evaluation metric.}\quad 
Our benchmark first measures the number of completed part assembly. However, due to its difficulty, most results are $0$ with the tested algorithms. To provide more fine-grained measures of progress, we report the number of completed \textbf{phases} (i.e., subtasks defined in \Cref{sec:environment_details:reward}) in each episode. 
A trained model is evaluated for 10 episodes, where their initial states are set following the provided task initialization guide tool.

\begin{figure}[b]
    \centering
    \vspace{-1em}
    \begin{subfigure}[t]{\linewidth}   
        \includegraphics[width=\linewidth]{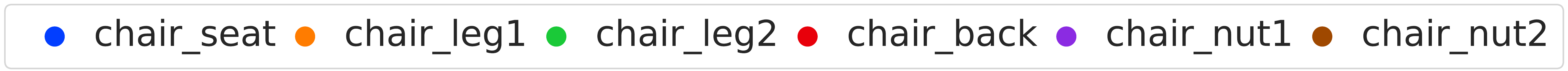}
    \end{subfigure}
    \\
    \vspace{5pt}
    \begin{subfigure}[t]{0.32\linewidth}
        \centering
        \includegraphics[width=\linewidth]{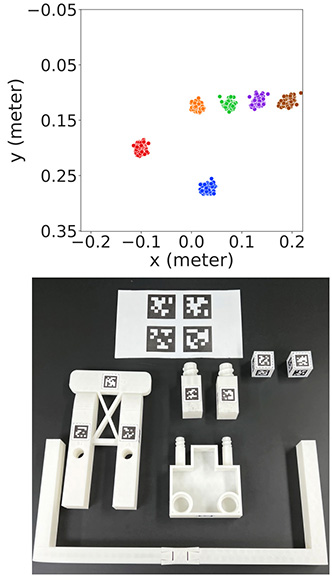}
        \caption{Low}
        \label{fig:low_perturbation}
    \end{subfigure}
    \hfill
    \begin{subfigure}[t]{0.32\linewidth}
        \centering
        \includegraphics[width=\linewidth]{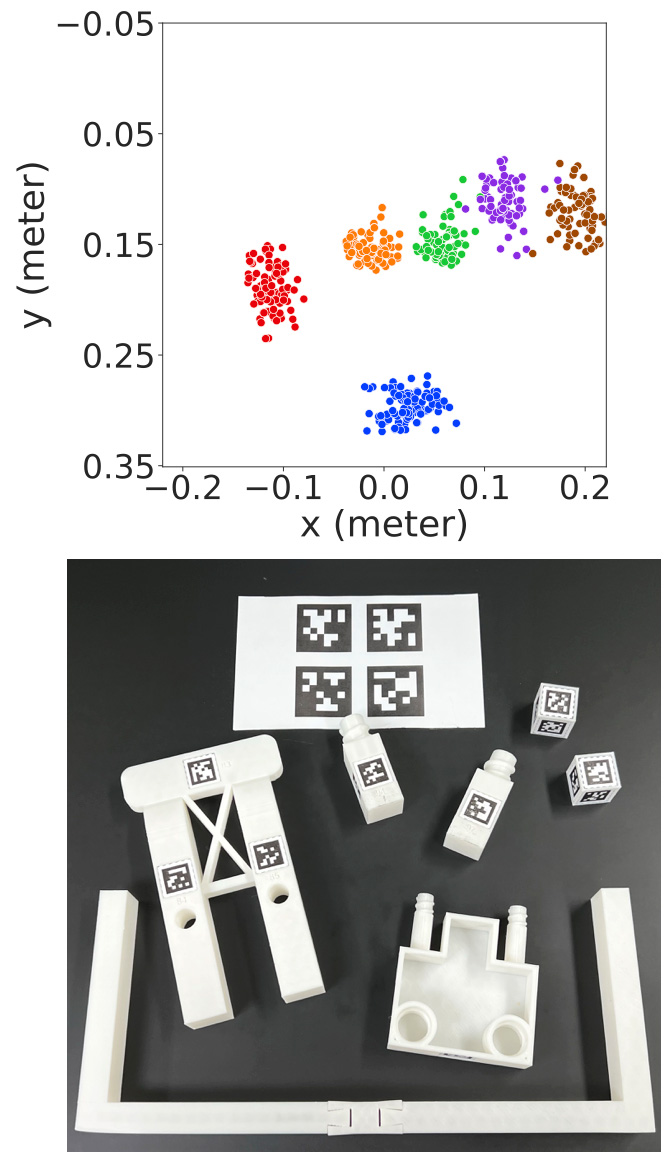}
        \caption{Medium}
        \label{fig:medium_perturbation}
    \end{subfigure}
    \hfill
    \begin{subfigure}[t]{0.32\linewidth}
        \centering
        \includegraphics[width=\linewidth]{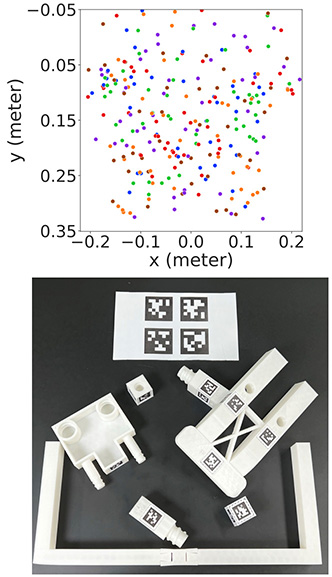}
        \caption{High}
        \label{fig:high_perturbation}
    \end{subfigure}
    \caption{
        \textbf{Three initialization randomness levels.} The higher the randomness in the initial state is, the more generalization capability is required for an agent. We provide three different levels of randomness in the initial states: (a)~\textbf{low} fixes part poses and allows only a little human error, (b)~\textbf{medium} randomizes the part poses around the given arrangement, and (c)~\textbf{high} randomly places all pieces. We visualize the initial part position distributions in our data (top) and example instances (bottom). 
    }
    \label{fig:initial_states}
\end{figure}

\textbf{Environment initialization randomness levels.}\quad
Although the ultimate goal of autonomous robotic manipulation is to handle all possible configurations, a wide variety of initial configurations makes the problem challenging.
Thus, we benchmark both tractable and challenging scenarios in three different levels with respect to the randomness in the initial furniture part configuration, as illustrated in \Cref{fig:initial_states}:
\begin{itemize}
    \item \textbf{Low}: The pose of each furniture piece is fixed; but, there can be a small noise during resetting.
    \item \textbf{Medium}: Based on the pre-defined poses in \textbf{Low}, we perturb each part with translation noise between $[\SI{-5}{\cm}, \SI{5}{\cm}]$ and rotational noise between $[\ang{-45}, \ang{45}]$.
    \item \textbf{High}: Furniture parts are randomly initialized on the workspace.
\end{itemize}

\textbf{Evaluation procedure.}\quad
Each individual trial is conducted according to the following procedure:
\begin{enumerate}
    \item The robot arm moves to the initial pose.
    \item The furniture parts are rearranged using our provided tool according to the initialization level (low, med, high).
    \item A policy controls the robot until it completes the task, stops motions for $5$~sec, shows unsafe movements, exceeds $350$ steps per skill, or exceeds $3000$ steps in total.
    \item The success rate and phase completion score are recorded.
\end{enumerate}

\section{Benchmarking Results}
\label{sec:experiments}

We evaluate an IL algorithm (BC~\citep{pomerleau1989alvinn} with the ResNet-18 encoder~\citep{he2016identity}) and the state-of-the-art offline RL algorithm (IQL~\citep{kostrikov2022offline} with R3M features~\citep{nair2022r3m}) on our reproducible, real-world furniture assembly benchmark.\footnote{This paper focuses on benchmarking end-to-end learning approaches since engineering furniture assembly procedures using TAMP without having access to state information is beyond the scope of this paper. But, this benchmark environment and tasks can be also used for research in TAMP.} We first benchmark the performance of selected individual skills in our single-skill benchmark (\Cref{sec:experiments:skill}). We then benchmark the full furniture assembly tasks in \Cref{sec:experiments:full-task}, where none of the algorithms make meaningful progress. To validate our benchmark for the full-assembly evaluation, we additionally benchmark a simpler task, \texttt{one\_leg} assembly, with $1000$ demonstrations. Finally, we benchmark our simulated environment in \Cref{sec:experiments:simulation}.

\subsection{Benchmark I: Single-Skill Benchmark}
\label{sec:experiments:skill}

Furniture assembly is a challenging manipulation task due to its long horizon and complex control. Thus, we investigate the difficulty of each individual subtask of furniture assembly, such as grasping, pushing, inserting, and screwing, as defined in \Cref{fig:assembly_procedures}. For the single-skill benchmark, we train a skill policy using manually segmented trajectories corresponding to the skill. Then, we measure the success rate of each skill by initializing the robot and objects to one of the skill's initial states following Appendix, \Cref{sec:experiment_details:single_skill}. We benchmark the first five skills of each furniture model. 

\begin{table}[t]
\centering
\caption{\textbf{Single-skill benchmark results.} First $5$ skills of each furniture task are independently evaluated and the average success rates over $10$ trials are reported in percentage. Each skill is evaluated on two different initialization levels: low and medium randomnesses.}
\label{tab:evaluation_skill}
\resizebox{\linewidth}{!}{
\begin{tabular}{l|cccccccccc}
    \toprule
     & \multicolumn{2}{c}{Skill 1} & \multicolumn{2}{c}{Skill 2} & \multicolumn{2}{c}{Skill 3} & \multicolumn{2}{c}{Skill 4} & \multicolumn{2}{c}{Skill 5} \\
     & low & med & low & med & low & med & low & med & low & med \\
     
    \midrule
    \texttt{lamp} & \multicolumn{2}{c}{Grasping\,1} & \multicolumn{2}{c}{Placing} & \multicolumn{2}{c}{Grasping\,2} & \multicolumn{2}{c}{Inserting} & \multicolumn{2}{c}{Screwing} \\
    \midrule
    BC                              & 0\% & 0\% & 40\%  & 30\%  & 30\% & 20\% & 0\% & 0\%  & 10\%  & 0\% \\
    IQL                             & 70\% & 40\% & 90\%  & 20\%  & 0\% & 20\% & 20\% & 0\%  & 10\%  & 0\% \\
    \midrule
    \multicolumn{2}{l}{\texttt{square\_table}}  & & & & & & & &  \\
    \midrule
    BC                              & 40\% & 0\% & 40\%  & 80\% & 0\% & 20\% & 0\% & 0\%  & 60\%  & 0\% \\
    IQL                             & 80\% & 70\% & 100\%  & 70\%  & 60\% & 40\% & 10\%  & 0\%  & 90\%  & 30\% \\
    \midrule
    \multicolumn{2}{l}{\texttt{desk}}   & & & & & & & & & \\
    \midrule
    BC                              &   60\% & 50\%  & 80\%  & 60\%  & 0\% & 0\% & 0\% & 0\%  & 0\% & 0\%  \\
    IQL                             &  100\% & 60\% & 100\%  & 60\%  & 60\% & 20\% & 20\% & 20\%  & 70\%  & 20\% \\
    \midrule
    \multicolumn{2}{l}{\texttt{round\_table}} & & & & & & & & & \\
    \midrule
    BC                              &  0\% & 50\% & 50\%  & 30\%  & 30\% & 0\% & 0\% & 0\%  & 0\%  & 0\% \\
    IQL                             &  80\% & 90\% & 30\%  & 40\%  & 50\% & 10\% & 10\% & 10\%  & 0\%  & 10\%  \\
    \midrule
    \multicolumn{2}{l}{\texttt{stool}}   & & & & & & & & & \\
    \midrule
    BC                              &   70\% & 0\% & 50\%  & 30\%  & 60\% & 0\% & 0\% & 0\%  & 0\%  & 0\% \\
    IQL                             &  90\% & 60\% & 80\%  & 40\%  & 90\% & 10\% & 0\% & 0\%  & 60\%  & 0\%  \\
    \midrule
    \multicolumn{2}{l}{\texttt{chair}}   & & & & & & & & & \\
    \midrule
    BC                              &  30\% & 10\% & 60\%  & 30\%  & 20\% & 0\% & 10\% & 0\%  & 20\%  & 0\%  \\
    IQL                             &  60\% & 50\% & 100\%  & 40\%  & 40\% & 10\% & 0\% & 0\%  & 50\%  & 10\%  \\
    \midrule
    \multicolumn{2}{l}{\texttt{drawer}} & & & & & & & & \multicolumn{2}{c}{Pushing} \\
    \midrule
    BC                              & 50\% & 60\% & 60\%  & 20\% & 90\% & 60\% & 0\%  & 0\%  & 10\% &  30\%  \\
    IQL                             &  80\% & 50\% & 70\%  & 0\%  & 70\% & 20\% & 0\% & 0\%  & 80\%  & 50\%  \\
    \midrule
    \multicolumn{2}{l}{\texttt{cabinet}} & & & & & & & & \multicolumn{2}{c}{Grasping\,3} \\
    \midrule
    BC                              &  70\% & 30\% & 70\%  & 30\%  & 0\% & 20\% & 0\% & 0\%  & 0\%  & 0\%  \\
    IQL                             &  60\% & 70\% & 70\%  & 60\%  & 20\% & 50\% & 0\% & 0\%  & 30\%  & 50\% \\
    \bottomrule
\end{tabular}
}
\vspace{-1em}
\end{table}

The single-skill benchmark results in \Cref{tab:evaluation_skill} demonstrate that individual skill policies can successfully learn ``grasping'' and ``placing'' skills. The ``pushing'' skill in \texttt{drawer} achieves $30$\% success rate, which is slightly worse than that of the ``grasping'' skill ($60$\%), with BC. This is mainly due to the stochastic dynamics of non-prehensile interactions. On the other hand, both algorithms struggle at ``inserting'' skill, which shows from $0$\% to $20$\% success rates. ``Inserting'' requires precise control to correctly align a screw and a hole, and when the agent fails to align them, it often gets stuck or easily goes into an out-of-distribution state from its training data. 

Interestingly, the ``screwing'' skill turns out to be not too difficult for IQL, e.g., achieving $90$\% and $70$\% success rates for \texttt{square\_table} and \texttt{desk} on the low randomness level, respectively. However, in \texttt{lamp} and \texttt{round\_table}, where the round-shaped parts need to be screwed, IQL struggles and achieves only $10$\% and $0$\% success rates, respectively.

\begin{figure*}[t]  
    \centering
    \includegraphics[width=\linewidth]{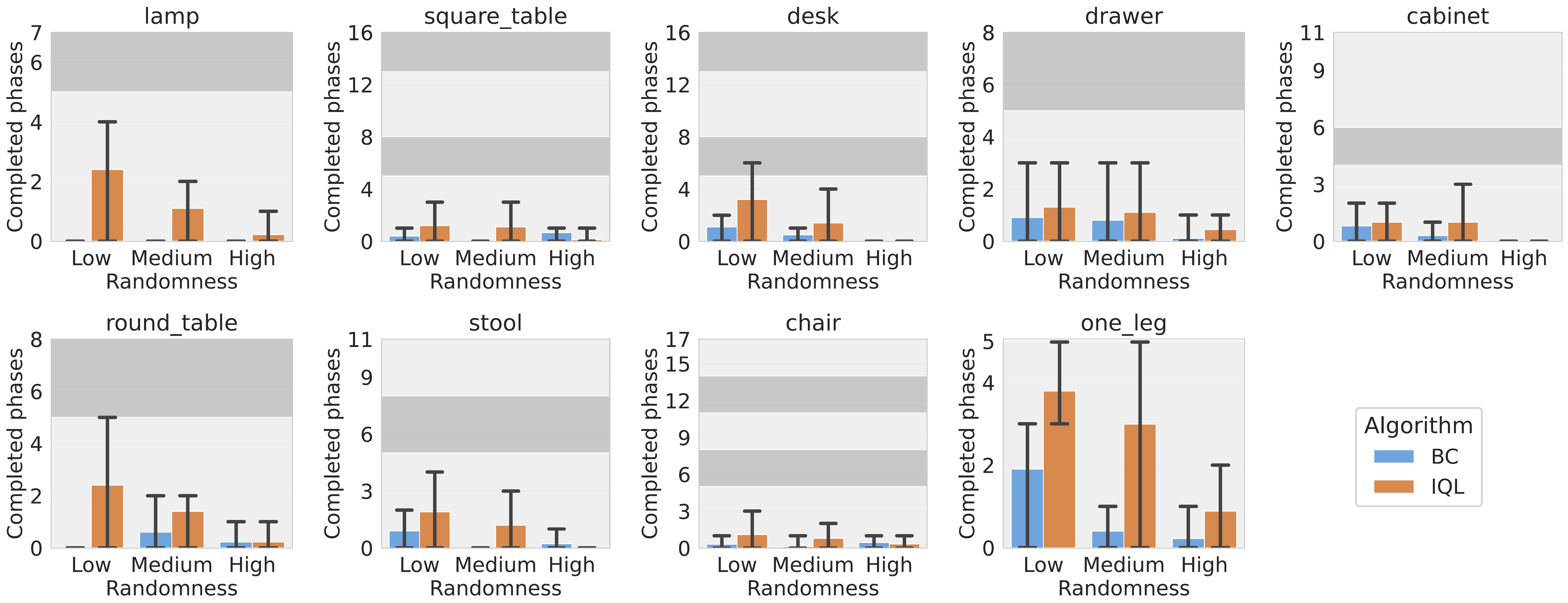}
    \caption{
        \textbf{Full-assembly benchmark results.}  We report the number of completed phases averaged over $10$ episodes and the error bars indicating the minimum and maximum completed phases. The background color indicates each part assembly and the maximum phase in each plot represents the task success. Each furniture model is evaluated on three different initialization levels (low, medium, and high).
    }
    \label{fig:evaluation_full_assembly}
    \vspace{-1em}
\end{figure*}

\subsection{Benchmark II: Full-Assembly Benchmark}
\label{sec:experiments:full-task}

We present full furniture assembly evaluation results in \Cref{fig:evaluation_full_assembly}. Overall, neither BC nor IQL achieve a single part assembly on the full-assembly tasks. Across all furniture and initialization levels, IQL performs better than BC, and lower randomness in the initial configuration leads to higher performance.

Qualitatively, the learned policies grasp the first target furniture part in most cases, and often place it in the designated position and grasp the second target part. However, the policies mostly fail to achieve more difficult phases, such as inserting and screwing, as can be observed in the single-skill benchmark.

The failure of these algorithms to even attach a pair of furniture parts despite the high-quality demonstration dataset highlights the need for further algorithmic improvements and suggests a good utility of our FurnitureBench for such algorithmic progression. More qualitative results are shown in Appendix, \Cref{fig:policy_sequences} and \href{https://clvrai.com/furniture-bench}{our website}.

\textbf{Is this benchmark tractable?}\quad 
To verify the feasibility of the proposed benchmark and provide a more tractable benchmark, we devise an easier benchmarking task, \texttt{one\_leg} assembly, which is a part of \texttt{square\_table} consisting of five phases: (1)~grasping a tabletop, (2)~placing it in the corner, (3)~picking up a table leg, (4)~inserting the leg into the hole on the tabletop, and (5)~screwing it, as illustrated in Appendix, \Cref{fig:assembly:square_table}. The task horizon is approximately $500$ timesteps. We collected $1000$ demonstrations for each of the low and medium initialization randomness levels.

\Cref{fig:correlation_simulation,fig:evaluation_full_assembly} show that IQL-R3M achieves $4$ phases on average and $40$\% success rate on the low randomness level. It always achieves the phase $3$ (grasping the leg) but fails at inserting $60$\% of the time. Once it succeeds in inserting, it mostly succeeds in screwing. This result reassures that ``inserting'' is the most challenging skill as it involves stochastic and frequent collisions.

\textbf{Does more data improve the performance?}\quad
In \Cref{fig:correlation_simulation}, we show that both BC-ResNet18 and IQL-R3M improve their performances with more demonstration data. We further scale the data by combining trajectories collected from the low and medium randomness levels. In \Cref{tab:ablation}, the IQL-R3M policies trained with $2000$ demonstrations in the mixed dataset (\textbf{Mixed data}) show significant improvements, $4.6$ and $3.7$ completed phases on average on the low and medium randomness levels, compared to the policies trained only with $1000$ demonstrations (\textbf{Original}), $3.8$ and $3.0$, respectively. These results suggest the need for learning generalizable policies with limited data.

\textbf{How critical the diversity of training data is?}\quad
To see the benefit of diverse data, we compare IQL-R3M policies trained using the low randomness level data (\textbf{Original low}) and the medium randomness level data (\textbf{Original med}). In \Cref{tab:ablation}, \textbf{Original low} shows a higher performance, $3.8$, than \textbf{Original med}, $3.0$, which implies that the benchmarking algorithms are not good at utilizing diverse data but require a large number of data on the same distribution for evaluation.

\textbf{Is a wrist camera essential for furniture assembly?}\quad
In our benchmark, we use visual observations from the front-view camera and wrist camera. We test whether the wrist camera is helpful for furniture assembly by training a policy only with the front-view camera input (\textbf{Front camera only}). Without the wrist camera input, the performance drops significantly from $3.8$ and $3.0$ to $2.0$ and $1.3$ on the low and medium randomness levels, respectively. This result means that the policies mostly fail to grasp the table leg without the wrist camera. This suggests that (1) the wrist camera provides information about the occluded area from the front camera and (2) some manipulation skills are easier to train and generalize better with the wrist-camera observation.

\textbf{Does a vision-based policy cheat on AprilTag?}\quad
One can concern about a policy learns to identify and exploit AprilTags for learning manipulation skills. Thus, we evaluate the IQL-R3M policy with the furniture models after removing AprilTags (\textbf{No AprilTag}) and after attaching random AprilTag markers different from the original ones (\textbf{Random AprilTag}) in \Cref{tab:ablation}. Due to the visual domain gaps in AprilTags, \textbf{Random AprilTag} shows slightly worse performance, $3.4$ and $2.7$ than \textbf{Original}, $3.8$ and $3.0$, under the low and medium randomness levels, respectively. \textbf{No AprilTag} has a little more performance drop and achieves $3.1$ and $2.4$, respectively. Despite these inferior results, the policy performs reasonably well in both cases, which ensures that the vision-based policies do not exploit the AprilTag marker information.

In summary, the \texttt{one\_leg} assembly experiments verify that our benchmark is \textit{tractable} but the algorithms require a huge amount of data to solve a part of a full furniture assembly task.

\begin{table}[b]
    \vspace{-1em}
    \caption{\textbf{Analysis on one-leg assembly.} We evaluate IQL-R3M under diverse settings and report the average completed phases over 10 episodes.}
    \label{tab:ablation}
    \centering
    \begin{tabular}{lcc}
        \toprule
        & low & med \\
        \midrule
        Original low (low randomness data) & $3.8$ & - \\ 
        Original med (medium randomness data) & $3.0$ & $3.0$ \\ 
        \midrule
        Mixed data & $4.6$ & $3.7$ \\
        Front camera only & $2.0$ & $1.3$ \\
        No AprilTag & $3.1$ & $2.4$  \\
        Random AprilTag & $3.4$ & $2.7$  \\
        \bottomrule
    \end{tabular}
\end{table}

\subsection{Benchmark III: Simulation Benchmark}
\label{sec:experiments:simulation}

For easy testing and debugging of algorithms, we provide the simulated environment, FurnitureSim, as described in \Cref{sec:benchmark:simulation}. Although direct transfer of knowledge from the simulator to the real world is difficult due to the simulation-to-real domain gap, the simulated results can provide a good proxy for how well an algorithm may work in the real world.

\begin{figure}[t]
    \centering
    \begin{subfigure}[t]{\linewidth}
        \includegraphics[width=\linewidth]{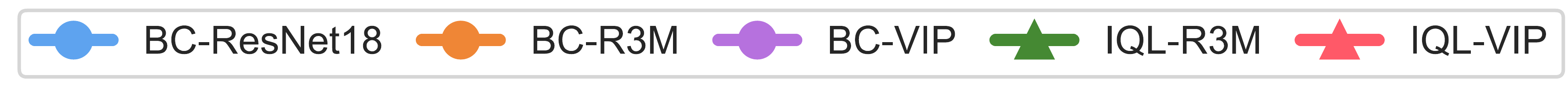}
    \end{subfigure}
    \includegraphics[width=0.485\linewidth]{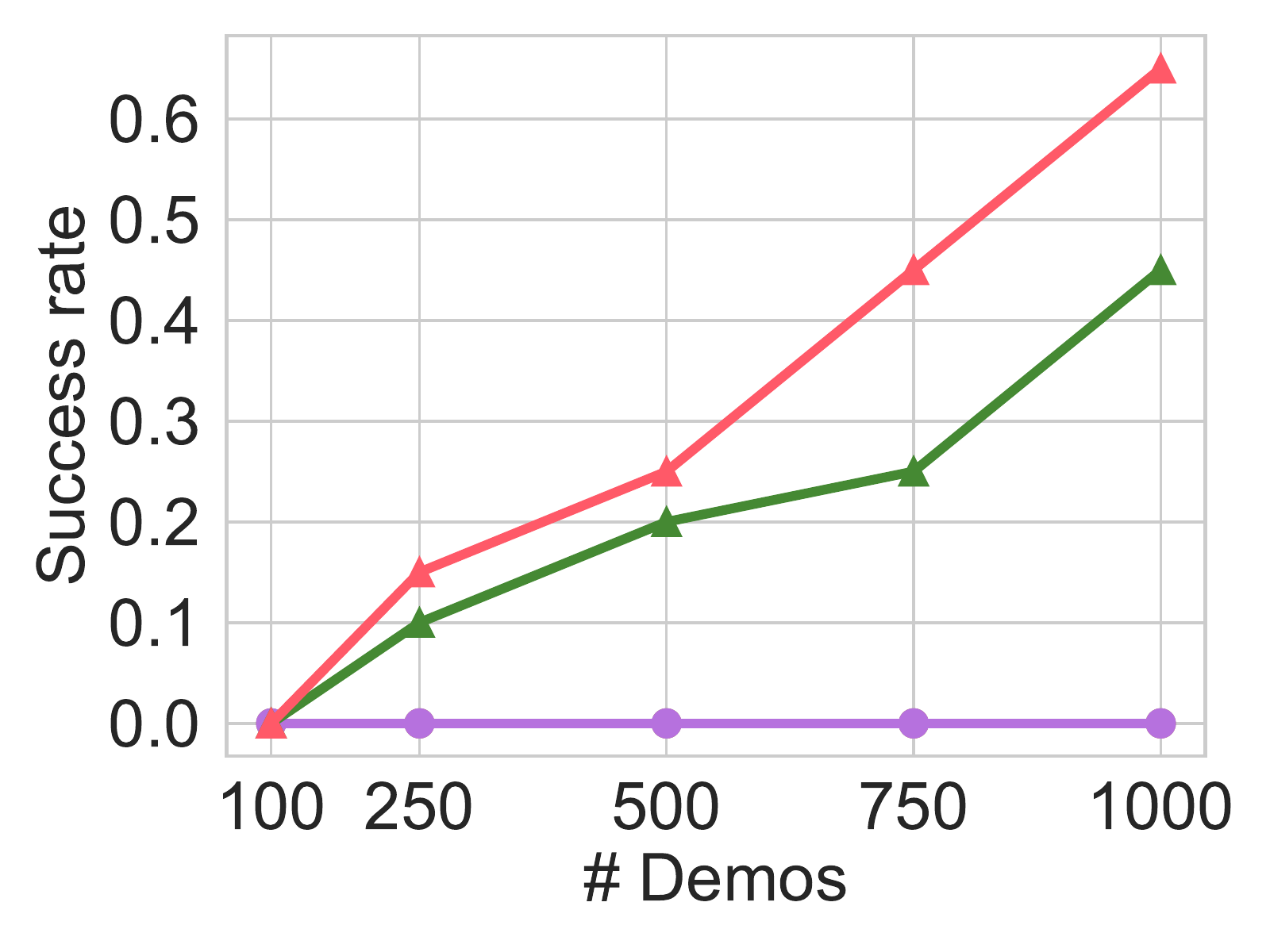}
    \includegraphics[width=0.485\linewidth]{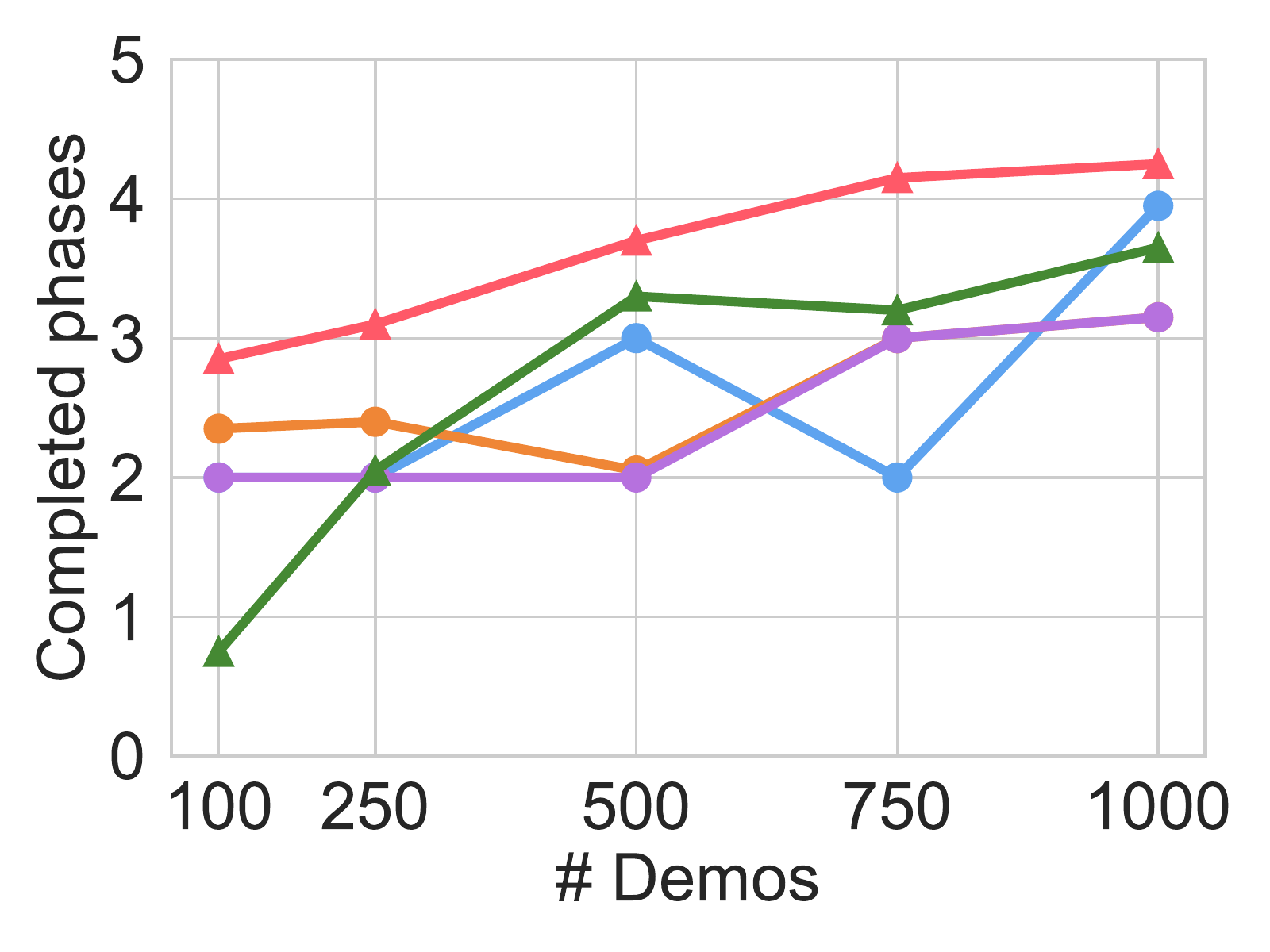}
    \caption{\textbf{Simulation benchmark results.} Using our FurnitureSim simulator, we evaluate BC and IQL with diverse visual encoders in the \texttt{one\_leg} assembly task on the low randomness level. We measure both success rates and average completed phases.} 
    \label{fig:evaluation_simulation}
\end{figure}

\Cref{fig:evaluation_simulation} shows the performance of simulated \texttt{one\_leg} assembly of BC and IQL with various visual encoders: ResNet-18~\citep{he2016identity}, R3M~\citep{nair2022r3m}, and VIP~\citep{ma2022vip}. For the ResNet-18 encoder, we randomly initialize its parameters and train the encoder together with the policy. For R3M and VIP, we use the pre-trained parameters without fine-tuning. Similar to the real-world benchmark, IQL shows higher success rates than BC. In terms of average completed phases, BC achieves comparable performance with IQL but BC fails at accurate insertion and screwing as these contact-rich interactions quickly bring an environment out of distribution. In the simulation, IQL-VIP shows higher performance than IQL-R3M; however, we found that IQL-R3M is slightly more robust in the real-world experiments and thus, we use R3M for the real-world benchmarks.

\section{Conclusion}
\label{sec:conclusion}

We propose FurnitureBench, a \textit{reproducible, real-world} furniture assembly benchmark, as a novel benchmark for testing complex long-horizon manipulation tasks on real robots. To serve a reproducible and easy-to-use benchmarking environment, we provide $3$D printable furniture part models, a step-by-step environment setup guide, software tools, large demonstration data, and a simulator, FurnitureSim. Our benchmarking results show that common imitation learning and the state-of-the-art offline reinforcement learning methods struggle at solving furniture assembly due to its long-horizon nature and complex manipulation skills. Therefore, it is a well-suited benchmark to encourage robot learning researchers to tackle complex long-horizon real-world manipulation tasks with better sample and demonstration efficiency. 

\textbf{Limitations and Future work.}\quad
While FurnitureBench includes many perspectives of real-world furniture assembly, the $3$D furniture models are still tailored to common robotic arms for research, e.g., all pieces have widths larger than \SI{2}{\cm} for easy grasping, which is larger than the tiny screws used in real-world IKEA furniture. Furthermore, our furniture models are much smaller in scale compared to the real-world furniture, which requires robots with a wider workspace and higher payload. Despite such simplification, FurnitureBench still serves most of common challenges in solving complex long-horizon manipulation tasks in the real world, such as temporal credit assignment, exploration, perception, and dexterous manipulation; thus, we believe solving this benchmark will make us a step closer to solving real-world furniture assembly.

Our benchmark is mainly evaluated using single-task criteria with a single robot. Leveraging shared information across different furniture assembly tasks in a multi-task RL perspective (e.g., diverse furniture categories, shapes, materials) or a multi-embodiment setup is worth to be examined.

FurnitureBench uses a single Franka Emika Panda arm and using only a single hand limits the dexterity of possible manipulation skills, resulting in the limited interactions tested in the benchmark. Extending our benchmark to multi-arm or multi-robot collaboration can endow robots with more dexterity.

Lastly, FurnitureSim has simulation-to-real gaps in both visual and physical domains. Especially, we found that system identification is very challenging. For example, the same torque command does not lead to the same robot trajectories in simulation and the real world due to inaccurate robot modeling (e.g., mass of each robot part, friction of joints). We leave closing the simulation-to-real gaps as future work. Simulation-to-real policy transfer under an imperfect simulator could be another interesting direction.

\section*{Acknowledgments}
This work is supported by Institute of Information \& communications Technology Planning \& Evaluation (IITP) grant (No.2019-0-00075, Artificial Intelligence Graduate School Program, KAIST) and National Research Foundation of Korea (NRF) grant (NRF-2021H1D3A2A03103683), funded by the Korea government (MSIT).
We would like to thank Shivin Dass for the initial investigation on the robotic setup, Jesse Zhang and Ayush Jain for their thoughtful reviews and suggestions, and all other CLVR lab members for constructive feedback. We also would like to thank Yashraj Narang for providing advice on Isaac Gym and Factory, and Suraj Nair and Ajay Mandlekar for sharing their robot control stack code.

\bibliography{bib/conference, bib/deep_learning, bib/drl, bib/env, bib/hrl, bib/imitation, bib/lee, bib/modular, bib/rl, bib/robotics, bib/sim2real, bib/real-world_benchmark, bib/representation}
\bibliographystyle{plainnat}

\clearpage
\appendices

\section{Overview}

To enhance the main paper, we additionally provide the following supplementary materials: 
\begin{itemize}
    \item \textbf{Environment details}:
    We elaborate on the hardware configurations, software details, simulator, reward function, action and observation space, and AprilTag pose estimation in \Cref{sec:environment_details}.

    \item \textbf{Evaluation details}: 
    We elaborate on the detailed evaluation protocols of our benchmarks in \Cref{sec:experiment_details}.

    \item \textbf{Training details}: 
    We elaborate on how we train BC and IQL for our benchmarks in \Cref{sec:training_details}.

    \item \textbf{Dataset details} reveal details of the data collection process and characters of datasets \Cref{sec:dataset_details}.

    \item \textbf{Furniture model details}:
    We describe a suite of furniture models and their assembly processes in \Cref{sec:furniture_models}.

    \item \textbf{Qualitative results}:
    Videos of the qualitative results of BC and IQL can be found in \url{https://clvrai.com/furniture-bench}.

    \item \textbf{Code} and \textbf{data}: \url{https://github.com/clvrai/furniture-bench}
    

    \item \textbf{Step-by-step environment setup guide}:
    We describe our reproducible environment setup instruction in \Cref{sec:instruction}.
\end{itemize}

\section{Environment Details}
\label{sec:environment_details}

\subsection{Hardware Specifications}
\label{sec:environment_details:hardware_details}

FurnitureBench aims to provide a reproducible real-robot system by adopting products widely available across the world and $3$D printing objects. Here is the list of products used to build our benchmark system:
\begin{itemize}
    \item Franka Emika Panda~\href{https://www.franka.de/}{(link)} and bench clamp~\href{https://download.franka.de/Bench_Clamp.pdf}{(link)}
    \item 3x Intel RealSense D435~\href{https://store.intelrealsense.com/buy-intel-realsense-depth-camera-d435.html}{(link)}
    \item 2x Manfrotto variable friction arm with camera bracket~\href{https://www.manfrotto.com/us-en/photo-variable-friction-arm-with-bracket-244/}{(link)}
    \item 2x Manfrotto super clamp~\href{https://www.manfrotto.com/global/super-clamp-w-lt-stud-1-4-2900-035rl/}{(link)}
        \begin{itemize}
            \item For the front camera arm, 1x Ulanzi ULS01 camera mount~\href{https://www.amazon.com/Flexible-Adjustable-Articulated-Rotatable-Aluminum/dp/B08LV7GZVB?th=1}{(link)} can be used  instead of one Manfrotto variable friction arm and super clamp.\footnote{We offer two options for the front camera mount. In most of our evaluations, we used the Manfrotto variable friction arm and super clamp for the front camera. However, we observed that some inexperienced participants struggled at manipulating the Manfrotto variable friction arm in high precision. We found that Ulanzi ULS01 camera mount is more intuitive to set up due to its ability to independently move the camera arm along different axes, and we could reproduce our original results with the new camera arm. Both these options are detailed in our step-by-step setup guide.
            }   
        \end{itemize}
    \item Oculus Quest 2 (optional for data collection)~\href{https://store.facebook.com/quest/products/quest-2/}{(link)}
    \item Any keyboard and mouse
    \item Black IKEA TOMMARYD table~\href{https://www.ikea.com/us/en/p/tommaryd-table-anthracite-s99304804/}{(link)}
    \item Green photography backdrop (larger than $w \times h$: \SI{280}{cm}$\times$\SI{400}{cm} for wide-angle coverage)
    \item Background frame (larger than $w \times h$: \SI{190}{cm}$\times$\SI{150}{cm})
    \item 2x Standing pole to hang the green backdrop on the sides of the table    
    \item LED light panel supporting color temperature (\SI{4600}{\kelvin}-\SI{6000}{\kelvin}) and brightness ($\sim$\SI{4000}{\lumen})
    \item 4x 10 feet USB Type-C to Type-A 3.1 cables~\href{https://www.amazon.com/AmazonBasics-Double-Braided-Nylon-Type-C/dp/B07D7NNJ61}{(link)}
    \item Server computer (an Intel 8th generation i7 or AMD Ryzen 5 5600X processor, or better)
    \item Client computer (an Intel 8th generation i7 or AMD Ryzen 5 5600X processor, four USB 3 Type-A ports with high bandwidth, and an NVIDIA RTX 3070 GPU, or better)
    \item White PLA $3$D printer filament
    \item Double-sided rubber tape
    \item 2x M3 x \SI{10}{\milli\meter} hex socket screw for the wrist camera
    \item 1x M6 x \SI{25}{\milli\meter} hex socket screw for the wrist camera
\end{itemize}

Our system requires two computers, Server and Client machines. The server is directly connected to the robot and dedicated to real-time robot control. The client manages all other aspects, including taking sensory inputs from cameras as well as the robot, computing a deep neural network policy, and sending action commands to the robot. In this paper, we use the server machine with an AMD Ryzen 5 5600X CPU and the client machine with an AMD Ryzen 9 5950X CPU and an NVIDIA RTX 3090 GPU.

\begin{figure}[t]
    \centering
    \includegraphics[width=\linewidth]{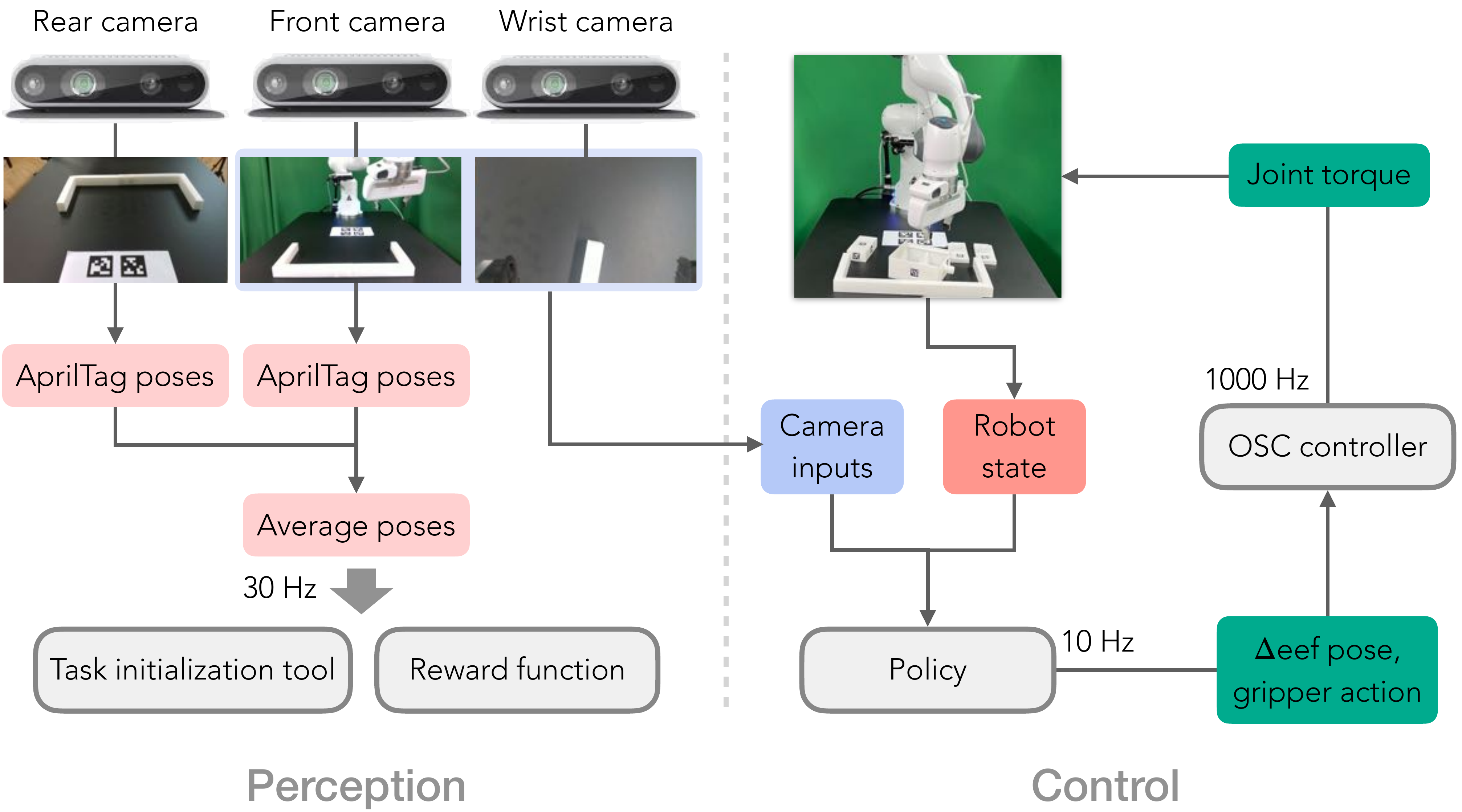}
    \caption{
        \textbf{Overview of our robot system design.} The agent receives proprioceptive robot states and images from the front and wrist cameras and takes action (delta end-effector pose and a binary gripper action) at a frequency of \SI{10}{\hertz}. For the reward function and initialization tool, we use images collected via the front and rear cameras to estimate the poses of the furniture parts using AprilTag. Note that the estimated furniture part poses are used only for environment functionalities, not for the learning agent.
        We choose the front and rear camera poses that can provide holistic views of the workspace as well as the high AprilTag detection accuracy.
    }
    \label{fig:system}
    \vspace{-1em}
\end{figure}

\subsection{Software Details}
\label{sec:environment_details:software_details}

Our robot control stack is built on top of Polymetis~\citep{Polymetis2021}, which provides a Python interface for the robot control stack, including basic robot controllers and network communication interfaces with the robot arm and gripper. 

We implement Operational Space Control (OSC~\citep{khatib1987unified}) in PyTorch~\citep{paszke2019pytorch} and optimize its performance with just-in-time compilation. Although the controller is optimized, OSC's computation may be slower than \SI{1000}{\hertz}, which is the control frequency of Franka Emika Panda. Thus, we compute a joint torque every three low-level control steps, which is approximately \SI{333}{\hertz} by repeating the same torque three times. Finally, as illustrated in \Cref{fig:system}, a policy action in delta end-effector pose (\SI{10}{\hertz}) is converted into approximately 33 joint torques, and each of these torques repeats three times.

\subsection{FurnitureSim Implementation}
\label{sec:environment_details:simulator}

We develop FurnitureSim, a simulated version of FurnitureBench, for easy and fast evaluation of algorithms. To minimize gaps between simulation and the real world, we adopt the code from FurnitureBench as much as we can. Specifically, we simply replace the real-world interactions with fast contact simulation supported by Factory~\citep{narang2022factory} and IsaacGym~\citep{makoviychuk2021isaac} with the PhysX physics engine. 
For example, the $3$D furniture models of FurnitureBench are used for FurnitureSim after converting their formats to the Signed Distance Function format (SDF). The simulator provides camera observations and robot proprioceptive states. An agent then receives this information and predicts its action just as the input comes from the real-world sensors. The robot controller code used in FurnitureBench takes input from the action command by the agent and computes low-level action signals.

Although the same $3$D models are used, estimating their exact physical properties is not trivial. We estimate the density of each $3$D model by first measuring its mass using a $3$D printing slicing tool and then measuring its volume using Blender $3$D Print Toolbox. With the mass and volume of a $3$D part, we can estimate its density.

3D-printed objects with PLA may have diverse friction coefficients~\citep{csirin2023effects}. We found that a friction coefficient between $0.15$ and $0.42$ works robustly for diverse furniture configurations in FurnitureSim. In this paper, we use $0.15$ as a friction coefficient.

\subsection{Reward Function}
\label{sec:environment_details:reward}

The reward function is essential for reinforcement learning and, more generally, evaluating the task's success. Our reward function is simple: $+1$ when a pair of parts are assembled and $0$, otherwise. Thus, a successful assembly will achieve a total reward of $N-1$, where $N$ is the number of parts. Note each pair of parts is rewarded only once, meaning disassembling and reassembling will not provide a reward again. 

The reward function is computed based on the relative poses between parts, which are estimated using AprilTags. To discriminate the success of assembly, we pre-define ground truth relative poses between matching parts when they are assembled. Two parts are considered assembled when the distance between the current relative pose and ground truth pose remain lower than a threshold (cosine similarity of each column vector of a rotation matrix is larger than $0.96$, and the absolute distance in each of x, y, z axes is smaller than \SI{7}{\mm}).

\Cref{fig:assembly_procedures} illustrates example assembly procedures for all furniture models. The images with red boundaries indicate that two parts are assembled, i.e., a $+1$ reward is provided. These assembly orders are not the only solutions. An agent can achieve rewards with different assembly orders, e.g., assembling the table leg to either the top left corner or the bottom right corner of the table top will lead to a reward of $+1$.

\textbf{Phase.}\quad
Furniture assembly tasks are complex and long-horizon -- even getting the first reward requires a large number of environmental steps.
Therefore, instead of benchmarking with reward, we propose to measure more fine-grained progress, which we refer to as a ``phase''. The phase score indicates the number of completed subtasks. The human demonstrators always accomplished each phase in the same order during data collection. During the evaluation, in contrast to the automated reward function, the success of each phase is judged by a human operator. Note that this phase information is not used for training an agent but for evaluating and analyzing a learned agent. Phase numbers for each furniture model can be found in \Cref{fig:assembly_procedures} on the top left corners.

\subsection{Action Space}

We use an 8D action space, which consists of delta end-effector $(x,y,z)$-position in meter ($3$D), delta orientation (quaternion, $4$D), and gripper action ($1$D). The action space is bounded between $-1$ and $+1$.

Although the gripper action is binary ($-1$ for opening and $+1$ for closing the gripper), a policy can output a continuous value between $-1$ to $1$. To prevent the gripper from repeating opening and closing actions, the gripper action is only activated when its absolute value is greater than a threshold of $0.019$.

The robot arm is controlled using OSC~\citep{khatib1987unified}, which computes joint torques to reach a given end-effector goal pose ($3$D Cartesian coordinate and $4$D quaternion). As shown in \Cref{fig:system}, the robot arm takes an end-effector control command at \SI{10}{\hertz}, and the OSC controller produces joint torques in real-time (\SI{1000}{\hertz}). The resulting joint torques are damped to prevent unstable movements due to high DoFs~\citep{kang1992joint}, and these conservative motions help maintain easy and safe robot control. 

Setting the desired end-effector goal too far away from the current end-effector position can generate excessive joint torques since the torques are computed based on the difference between the current and the goal positions. We use two strategies to avoid such instability: interpolation to the desired end-effector goal and clipping the delta action to a fixed amount. The interpolation is done by equally dividing the distance between the current end-effector position and the desired end-effector position into several steps. The total number of steps is defined by the controller frequency divided by the action command frequency. And the clipping range of the delta end-effector position is set to be $\pm$\SI{10}{\centi\meter}.

\subsection{Observation Space}

The observations available from our environment are listed in \Cref{tab:observation_space}. Although other sensory inputs are available via environmental steps and stored while data collection, only a portion of those are used for the agent's input. The observation space for the policy consists of two RGB images from the front and wrist camera and robot proprioceptive states. The front camera image is first downsampled to $455\times256$ (smaller edge match 256 while keeping the image ratio) and then center-cropped to $224\times224$. The wrist camera is directly downsampled to $224\times224$ to keep the wider view. We use INTER\_AREA interpolation in OpenCV~\citep{opencv_library} for downsampling for all the images. The proprioceptive robot state consists of the end-effector position, orientation (in quaternion), velocity, and gripper width.

\begin{table}[ht]
\centering
\caption{Observations available in FurnitureBench. The observations used for the input to the policy are check marked.}
\label{tab:observation_space}
\begin{tabular}{llll}
    \toprule
    Name & Shape & Unit & Policy \\
    \midrule
    Front camera RGB & (1280, 720, 3) & pixel value (0 - 255) & \checkmark \\ 
    Wrist camera RGB & (1280, 720, 3) & pixel value (0 - 255) & \checkmark \\ 
    Rear camera RGB & (1280, 720, 3) & pixel value (0 - 255)  &   \\ 
    Front camera depth & (1280, 720) & (0 - 65536) \SI{}{\milli\meter} &  \\
    Wrist camera depth & (1280, 720) & (0 - 65536) \SI{}{\milli\meter} &  \\
    Rear camera depth & (1280, 720) & (0 - 65536) \SI{}{\milli\meter}  &  \\
    Object poses & (\# parts $\times$ 7) & \SI{}{\meter} and quaternion &   \\
    EE position & (3) & \SI{}{\meter} & \checkmark \\
    EE orientation & (4) & quaternion & \checkmark \\
    EE linear velocity & (3) & \SI{}{\meter/\second} & \checkmark \\
    EE rotational velocity & (3) & \SI{}{\radian/\second}  & \checkmark \\
    Joint positions & (7) & \SI{}{\radian}  & \\
    Joint velocities & (7) & \SI{}{\radian/\second} &  \\
    Joint torques & (7) & \SI{}{\newton\cdot\meter} &  \\
    Gripper width & (1) & \SI{}{\meter} & \checkmark \\
    \bottomrule
\end{tabular}

\end{table}

\subsection{Multi-camera pose estimation using AprilTag}
\label{sec:environment_details:apriltag}

AprilTag pose estimation using a single camera is often inaccurate and can even fail to detect object poses when markers are occluded or angled. To compensate for this detection failure and inaccurate pose estimation, we use two parallel AprilTag detection modules using two cameras (front and rear cameras as illustrated in \Cref{fig:environment:setup}) and aggregate the estimated poses. Thus, the resulting pose estimations can be more accurate. We choose the original poses of the front and rear cameras in that the viewpoints provide holistic views of the workspace as well as clearly observe the  AprilTags in multiple orientations. To handle the noise (e.g., flipping axis) in AprilTag pose estimation, we filter outliers that are different from the last five detection results for each camera.

\begin{figure}[t]
    \centering
    \includegraphics[width=0.4\linewidth]{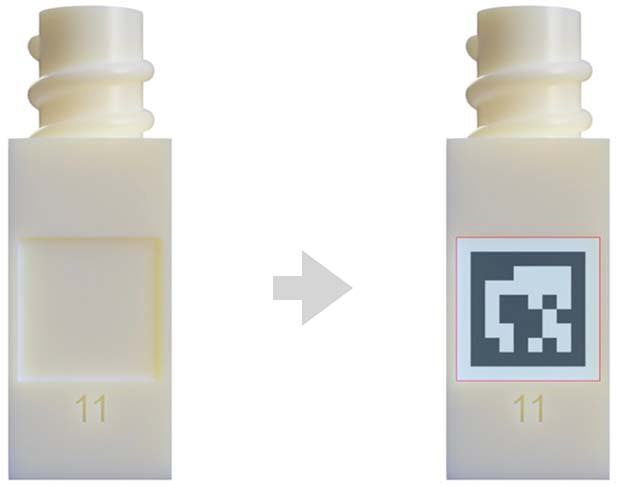}
    \caption{\textbf{AprilTag placeholder.} For easy and accurate marker placement, all $3$D models have AprilTag placeholders on their surfaces with corresponding AprilTag IDs.}
    \label{fig:marker_placeholder}
    \vspace{-1em}
\end{figure}

For accurate and easy marker placement, all $3$D models have placeholders for markers with their AprilTag IDs, as shown in \Cref{fig:marker_placeholder}. We further improve the reliability and accuracy of pose estimation using multiple markers for each furniture part. For each furniture part, the poses of the detected markers are converted to the canonical furniture part pose, where the position is its center of mass and the orientation follows the marker with the smallest ID. Then, the furniture pose is estimated by averaging the canonical poses computed from all detected markers.

To sum it up, we improve the stability and accuracy of pose estimation by (1) averaging over estimated poses from two cameras, (2) filtering outliers, and finally, (3) averaging over estimated poses of multiple markers on each furniture part.

\section{Experiment Details}
\label{sec:experiment_details}

\subsection{Reproducibility Experiment Details}
\label{sec:experiment_details:reproducibility}

In the reproducibility experiments, we investigate the difficulty of our environment setup and confirm the reproducibility of our evaluation results. We ask participants to build the environment from scratch, providing all the necessary components for our benchmarking environment.

\begin{figure}[ht]
    \centering
    \includegraphics[width=1.0\linewidth]{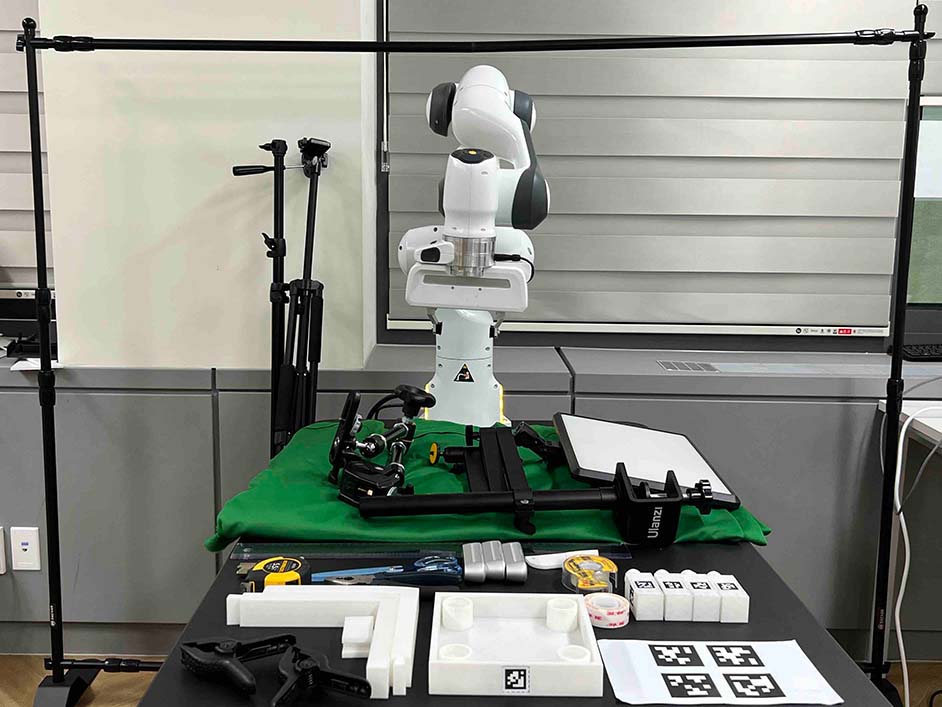}
    \caption{\textbf{Initial configuration for reproducibility experiments.} Each participant is asked to recreate the environment with the provided materials by following our instruction.}
    \label{fig:environment:reproducibility_init}
    \vspace{-1em}
\end{figure}

At the starting point, the robot's pose differs from our specified arrangement, so participants need to adjust its position according to the provided instruction. Since lifting and relocating the robot alone is physically demanding, an additional person assisted during the robot's mounting process. Additionally, the required software has been pre-installed, as our primary goal is to assess the reproducibility of the real-world environment setup rather than software installation. The initial configuration is shown in \Cref{fig:environment:reproducibility_init}. 

After finishing the hardware installation, the pre-trained IQL-R3M policy is evaluated 20 rollouts for the \texttt{one\_leg} assembly task, and the average number of completed phases is measured. For evaluation, we use the low initialization level, and the policy is trained with $1000$ demonstrations.

\begin{figure*}[t]
    \begin{subfigure}[t]{0.3\linewidth}
        \includegraphics[width=0.32\linewidth]{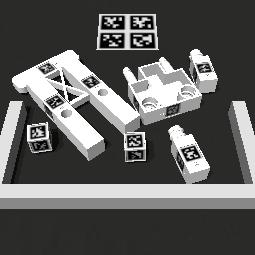} \hfill
        \includegraphics[width=0.32\linewidth]{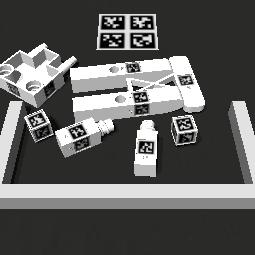} \hfill
        \includegraphics[width=0.32\linewidth]{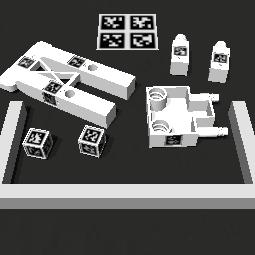}
        \caption{\texttt{chair}}
    \end{subfigure}
    \hspace{0.3em}
    \begin{subfigure}[t]{0.3\linewidth}
        \includegraphics[width=0.32\linewidth]{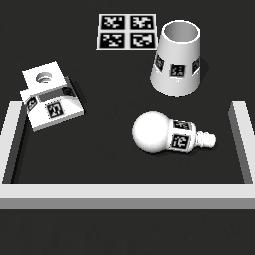} \hfill
        \includegraphics[width=0.32\linewidth]{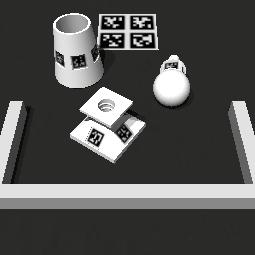} \hfill
        \includegraphics[width=0.32\linewidth]{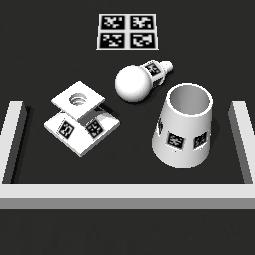}
        \caption{\texttt{lamp}}
    \end{subfigure}
    \hspace{0.3em}
    \begin{subfigure}[t]{0.3\linewidth}
        \includegraphics[width=0.32\linewidth]{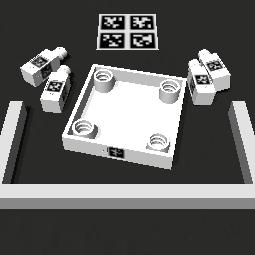} \hfill
        \includegraphics[width=0.32\linewidth]{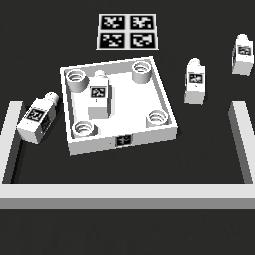} \hfill
        \includegraphics[width=0.32\linewidth]{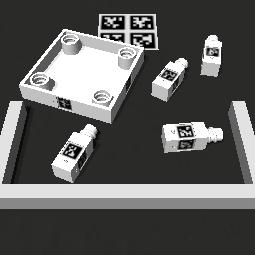}
        \caption{\texttt{square\_table}}
    \end{subfigure}
    \vspace{0.5em}
    \\
    \begin{subfigure}[t]{0.3\linewidth}
        \includegraphics[width=0.32\linewidth]{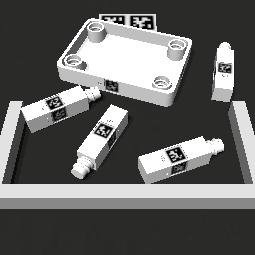} \hfill
        \includegraphics[width=0.32\linewidth]{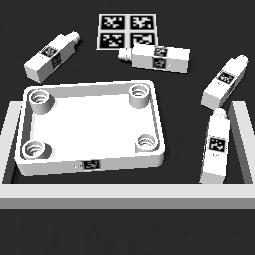} \hfill
        \includegraphics[width=0.32\linewidth]{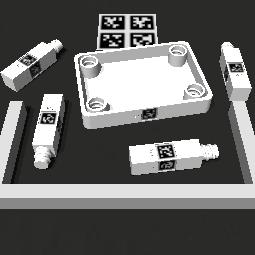}
        \caption{\texttt{desk}}
    \end{subfigure}
    \hspace{0.3em}
    \begin{subfigure}[t]{0.3\linewidth}
        \includegraphics[width=0.32\linewidth]{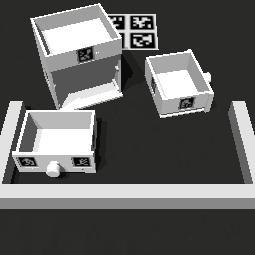} \hfill
        \includegraphics[width=0.32\linewidth]{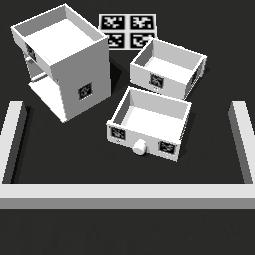} \hfill
        \includegraphics[width=0.32\linewidth]{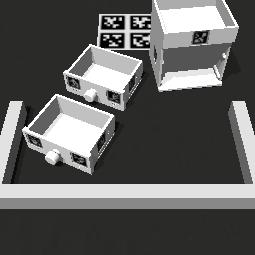}
        \caption{\texttt{drawer}}
    \end{subfigure}
    \hspace{0.3em}
    \begin{subfigure}[t]{0.3\linewidth}
        \includegraphics[width=0.32\linewidth]{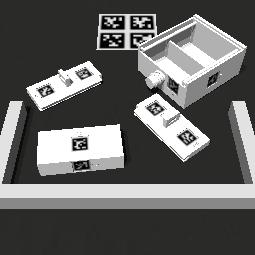} \hfill
        \includegraphics[width=0.32\linewidth]{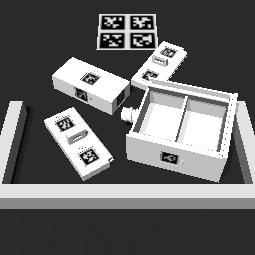} \hfill
        \includegraphics[width=0.32\linewidth]{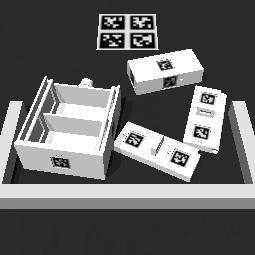}
        \caption{\texttt{cabinet}}
    \end{subfigure}
    \vspace{0.5em}
    \\    
    \begin{subfigure}[t]{0.3\linewidth}
        \includegraphics[width=0.32\linewidth]{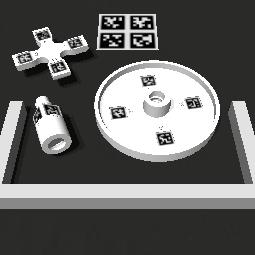} \hfill
        \includegraphics[width=0.32\linewidth]{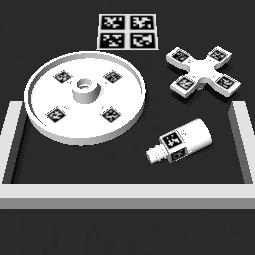} \hfill
        \includegraphics[width=0.32\linewidth]{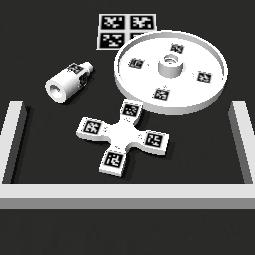}
        \caption{\texttt{round\_table}}
    \end{subfigure}
    \hspace{0.3em}
    \begin{subfigure}[t]{0.3\linewidth}
        \includegraphics[width=0.32\linewidth]{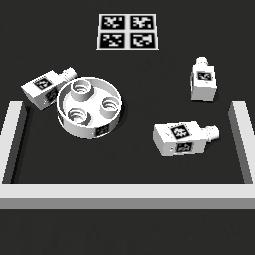} \hfill
        \includegraphics[width=0.32\linewidth]{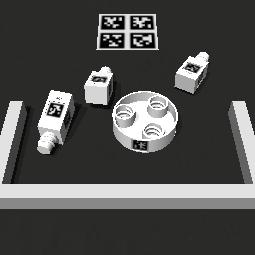} \hfill
        \includegraphics[width=0.32\linewidth]{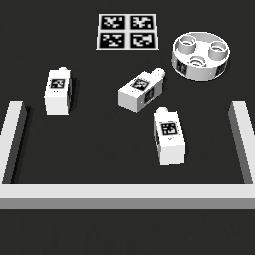}
        \caption{\texttt{stool}}
    \end{subfigure}
    \caption{\textbf{Three pre-defined initial states for evaluation in the high task initialization randomness level}.}
    \label{fig:high_randomness}
    \vspace{-1em}
\end{figure*}

\subsection{Single-Skill Benchmark Experiment Details}
\label{sec:experiment_details:single_skill}

For the single-skill benchmark, we need to define an initial state for each skill evaluation: we randomly sample one starting state of the skill from a dataset and fix this state for the entire evaluation. For the medium initialization randomness level, we add different noises to this pre-defined state for skill evaluation rollouts. To prevent an agent from repeating the same episodes during evaluation, we also randomize the end-effector pose ($\pm$\SI{0.5}{\cm}, $\pm$\ang{5} and $\pm$\SI{5}{\cm}, $\pm$\ang{15} for the low and medium randomness levels, respectively).

\subsection{Full-Assembly Benchmark Experiment Details}
\label{sec:experiment_details:full_assembly}

Three initialization randomness levels, namely low, medium, and high, introduce the different levels of perturbation on the initial poses of furniture parts, as illustrated in \Cref{sec:experimental_setup}. The task initialization GUI tool helps configure the environment to align with the specified poses, as illustrated in \Cref{sec:benchmark:easy-to-use}. 

For the high randomness level, we collected data from completely random initial configurations. However, for consistent evaluation results in the high randomness level, we use three distinct predetermined poses as initialization states, as shown in \Cref{fig:high_randomness}.

\section{Training Details}
\label{sec:training_details}

The observation space of our environment consists of the front-view and wrist-view camera inputs, and proprioceptive robot states (end-effector position, orientation, velocity, and gripper width). For efficient visual policy learning, we use the pre-trained visual encoder, R3M~\citep{nair2022r3m} and VIP~\citep{ma2022vip}, which is trained with large-scale egocentric human videos. We concatenate $2048$-D feature vector extracted from each camera image with proprioceptive states and feed this into $3$-layer MLP consisting of ($512$, $256$, $256$) hidden units for the policy network. Hyperparameters for our baseline implementations are listed in \Cref{tab:hyperparameters}.

\textbf{Behavioral Cloning (BC~\citep{pomerleau1989alvinn}).}\quad 
We train the policy using a batch of size $64$ for $50$ epochs. The policy is optimized using the Adam optimizer~\citep{kingma2014adam} with a learning rate of $0.0003$.

\textbf{Implicit Q Learning (IQL~\citep{kostrikov2022offline}).}\quad
We use the official JAX~\citep{jax2018github} implementation of IQL with minor changes to use R3M~\citep{nair2022r3m} and VIP~\citep{ma2022vip} features. The policy is trained with sparse rewards, where rewards in each demonstration are manually marked as $+1$ when each furniture part is assembled.

\begin{table}[ht]
  \caption{Hyperparameters for BC and IQL.}
  \label{tab:hyperparameters}
  \centering
  \begin{tabular}{lll}
    \toprule
    Hyperparameter                  & BC                    & IQL               \\
    \midrule                   
    Learning rate                   & 3e-4                  & 3e-4              \\ 
    Learning rate decay             & Step decay            & Cosine decay      \\
    Learning rate decay factor      & 0.5                   & -                 \\
    \# Mini-batches                 & 64                    & 256               \\
    Policy \# Hidden units          & (512, 256, 256)       & (512, 256, 256)   \\
    Q/V \# Hidden units             & -                     & (512, 256, 256)   \\
    Activation                      & ReLU                  & ReLU              \\
    Image encoder                   & R3M, VIP, ResNet-18   & R3M, VIP               \\
    Discount factor ($\gamma$)      & -                     & 0.996             \\
    Expectile ($\tau$)              & -                     & 0.8               \\
    Inverse temperature ($\beta$)   & -                     & 0.5               \\
    \bottomrule
  \end{tabular}
\end{table}

\section{Dataset Details}
\label{sec:dataset_details}

\subsection{Data Collection}
\label{sec:data_details:data_collection}

\label{sec:training_details:teleoperation}
\begin{figure}[h]
    \centering
    \includegraphics[width=\linewidth]{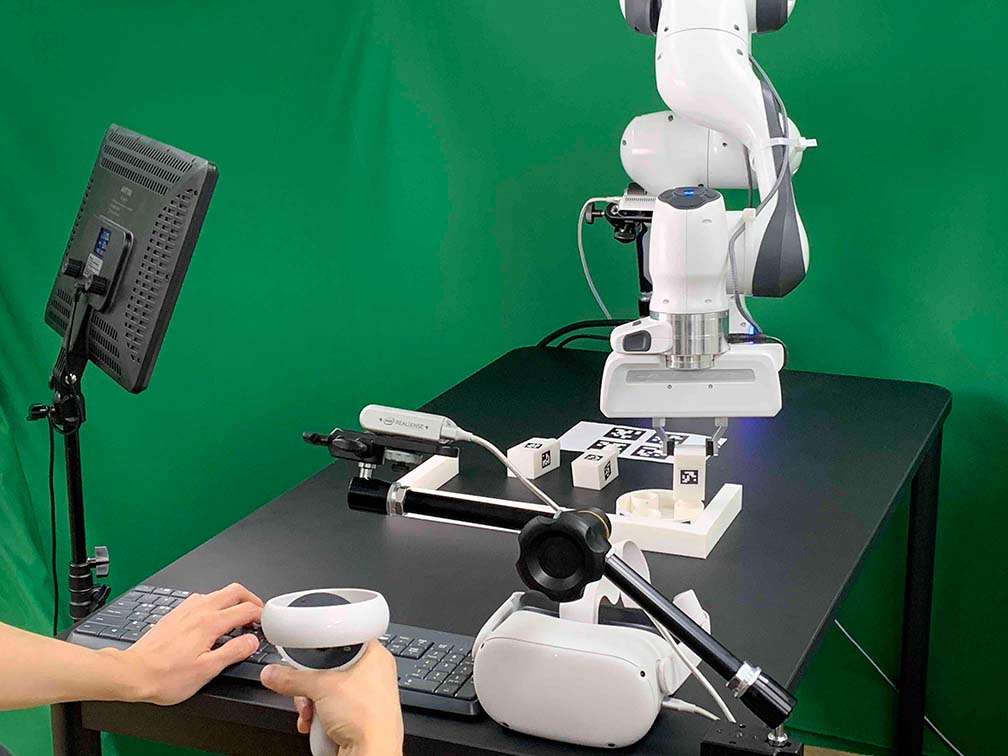}
    \caption{
        \textbf{Teleoperation setup.} Our system supports Oculus Quest 2 controller and keyboard to control a $7$-DoF robotic arm for demonstration collections.
    }
    \label{fig:system:teleoperation}
    \vspace{-1em}
\end{figure}

We collected $219$ hours of demonstration data over eight furniture models from two human operators. For each of low and medium initialization levels, we collected $100$ demonstrations for \texttt{desk}, \texttt{round\_table}, \texttt{stool}, \texttt{chair}; $150$ demonstrations for \texttt{cabinet}, \texttt{lamp}, \texttt{square\_table}; and $250$ demonstrations for \texttt{drawer}. For the high initialization level, we collected $50$ demonstrations for each furniture model.

Each timestep of a demonstration consists of RGB-D frames from three cameras, robot action, reward, and AprilTag poses for all furniture parts. In the case of low and medium randomness, a phase completion mask is provided. In addition, demonstrations include metadata, such as furniture model id, errors during data collection, and episode success. The statistics of the collected data can be found in \Cref{tab:data}.

To cover diverse configurations, we use a single light panel with various color temperatures and vary their positions and directions over time. We also randomize the front-view camera pose every episode. We plan to extend the diversity of data by collecting demonstrations from multiple different laboratories. The teleoperation setup is illustrated in \Cref{fig:system:teleoperation}.

\subsection{Data Annotation}
\label{sec:data_details:data_annotation}

During data collection, the human operators corrected wrong reward signals from the automatic reward function, possibly caused by a detection error or occlusion. In addition, the number of completed phases in the low and medium randomness levels are manually annotated. The phase of high randomness is not annotated since the assembly processes are inconsistent across the trajectories depending on the initialization.

\newpage
\onecolumn

\section{Furniture Models}
\label{sec:furniture_models}

\begin{figure*}[ht]
    \begin{subfigure}[t]{0.3\linewidth}
        \includegraphics[width=0.32\linewidth]{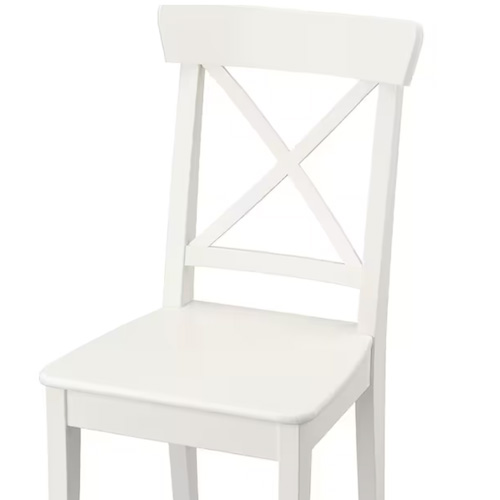} \hfill
        \includegraphics[width=0.32\linewidth]{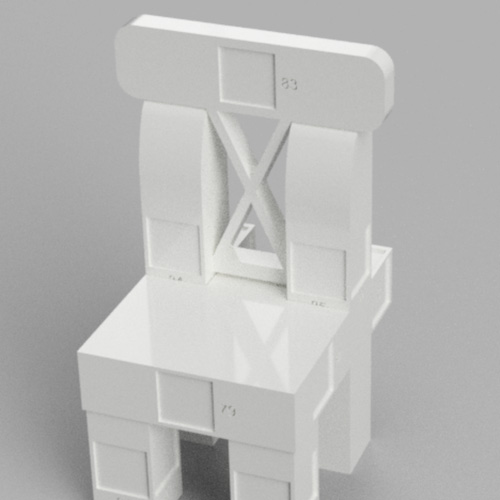} \hfill
        \includegraphics[width=0.32\linewidth]{fig/3d_printed/chair_silver.jpg}
        \caption{IKEA INGOLF \texttt{chair}}
    \end{subfigure}
    \begin{subfigure}[t]{0.3\linewidth}
        \includegraphics[width=0.32\linewidth]{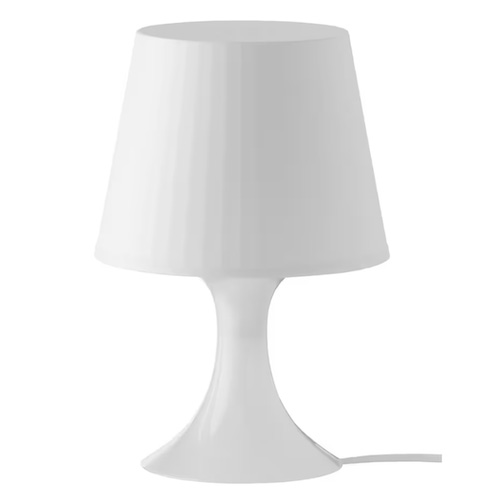} \hfill
        \includegraphics[width=0.32\linewidth]{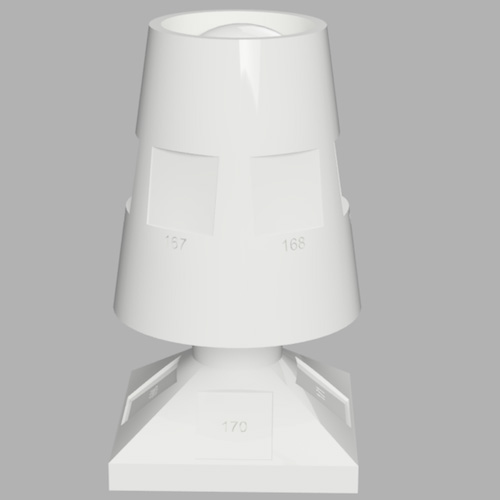} \hfill
        \includegraphics[width=0.32\linewidth]{fig/3d_printed/lamp_silver.jpg}
        \caption{IKEA LAMPAN \texttt{lamp}}
    \end{subfigure}
    \begin{subfigure}[t]{0.3\linewidth}
        \includegraphics[width=0.32\linewidth]{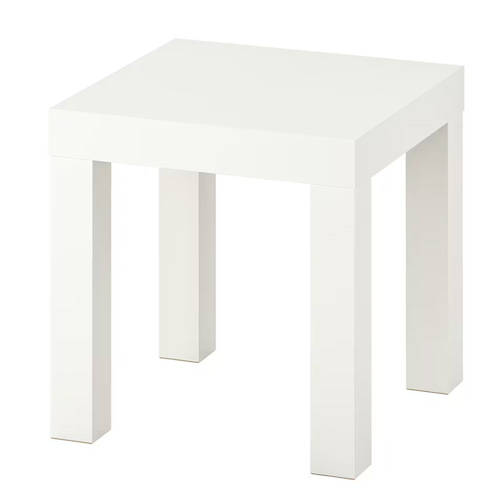} \hfill
        \includegraphics[width=0.32\linewidth]{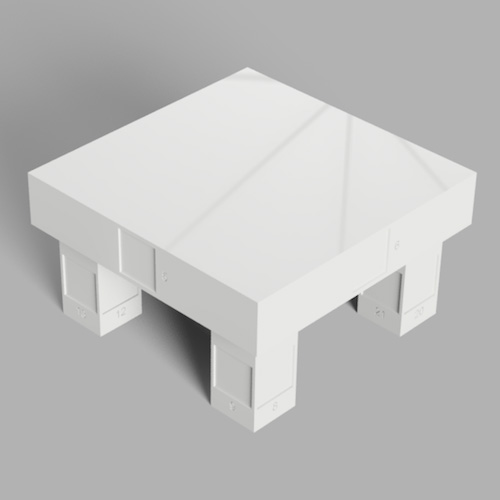} \hfill
        \includegraphics[width=0.32\linewidth]{fig/3d_printed/square_table_silver.jpg}
        \caption{IKEA LACK \texttt{square\_table}}
    \end{subfigure}
    \vspace{1em}
    \\
    \begin{subfigure}[t]{0.3\linewidth}
        \includegraphics[width=0.32\linewidth]{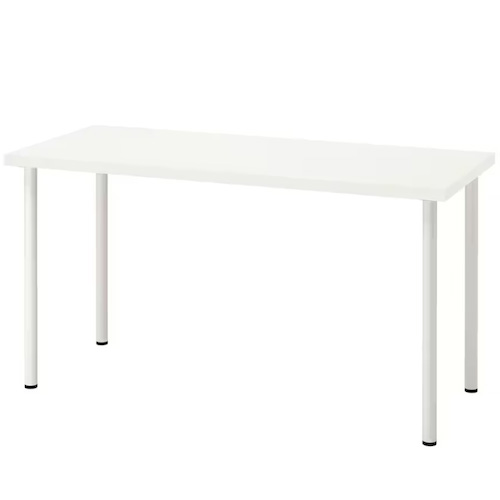} \hfill
        \includegraphics[width=0.32\linewidth]{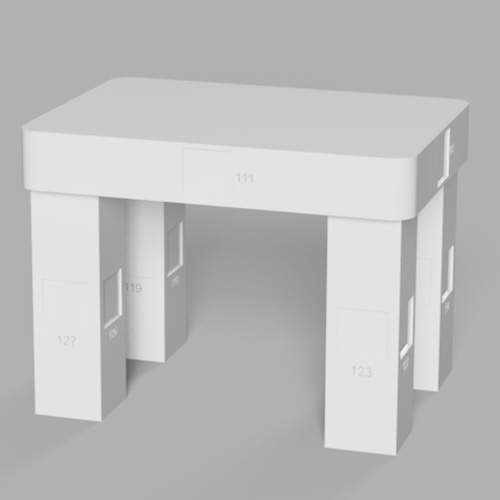} \hfill
        \includegraphics[width=0.32\linewidth]{fig/3d_printed/desk_silver.jpg}
        \caption{IKEA LAGKAPTEN \texttt{desk}}
    \end{subfigure}
    \begin{subfigure}[t]{0.3\linewidth}
        \includegraphics[width=0.32\linewidth]{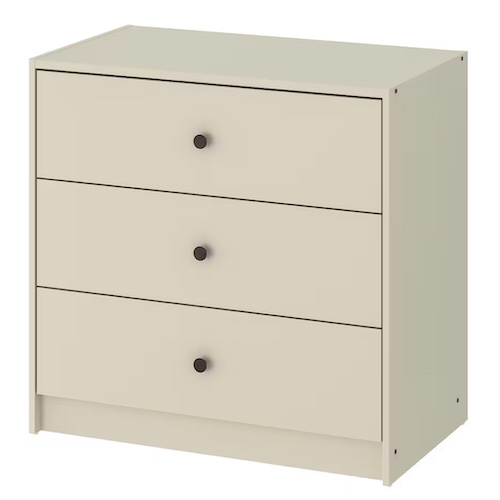} \hfill
        \includegraphics[width=0.32\linewidth]{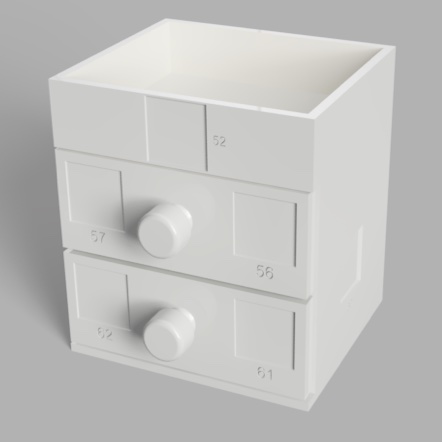} \hfill
        \includegraphics[width=0.32\linewidth]{fig/3d_printed/drawer_silver.jpg}
        \caption{IKEA GURSKEN \texttt{drawer}}
    \end{subfigure}
    \begin{subfigure}[t]{0.3\linewidth}
        \includegraphics[width=0.32\linewidth]{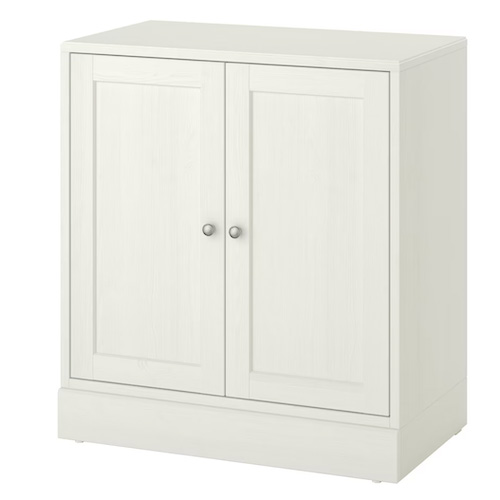} \hfill
        \includegraphics[width=0.32\linewidth]{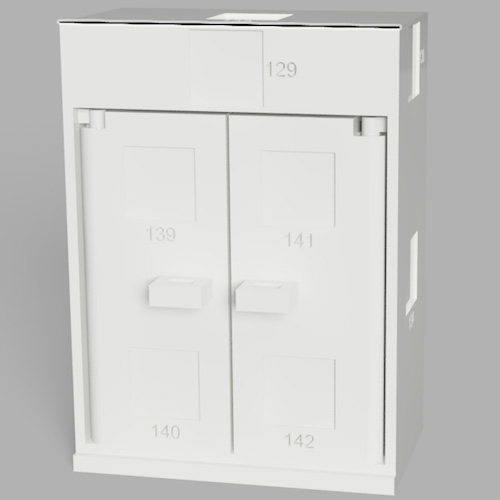} \hfill
        \includegraphics[width=0.32\linewidth]{fig/3d_printed/cabinet_silver.jpg}
        \caption{IKEA HAVSTA \texttt{cabinet}}
    \end{subfigure}
    \vspace{1em}
    \\    
    \begin{subfigure}[t]{0.3\linewidth}
        \includegraphics[width=0.32\linewidth]{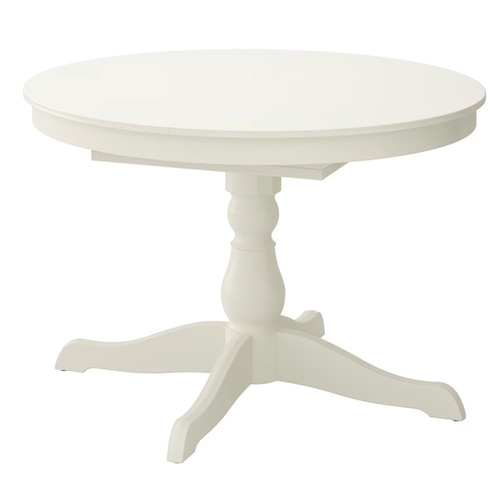} \hfill
        \includegraphics[width=0.32\linewidth]{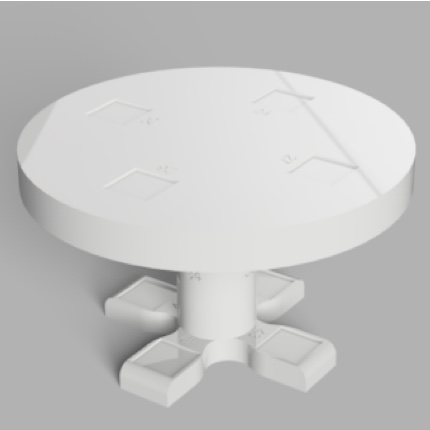} \hfill
        \includegraphics[width=0.32\linewidth]{fig/3d_printed/round_table_silver.jpg}
        \caption{IKEA INGATORP \texttt{round\_table}}
    \end{subfigure}
    \begin{subfigure}[t]{0.3\linewidth}
        \includegraphics[width=0.32\linewidth]{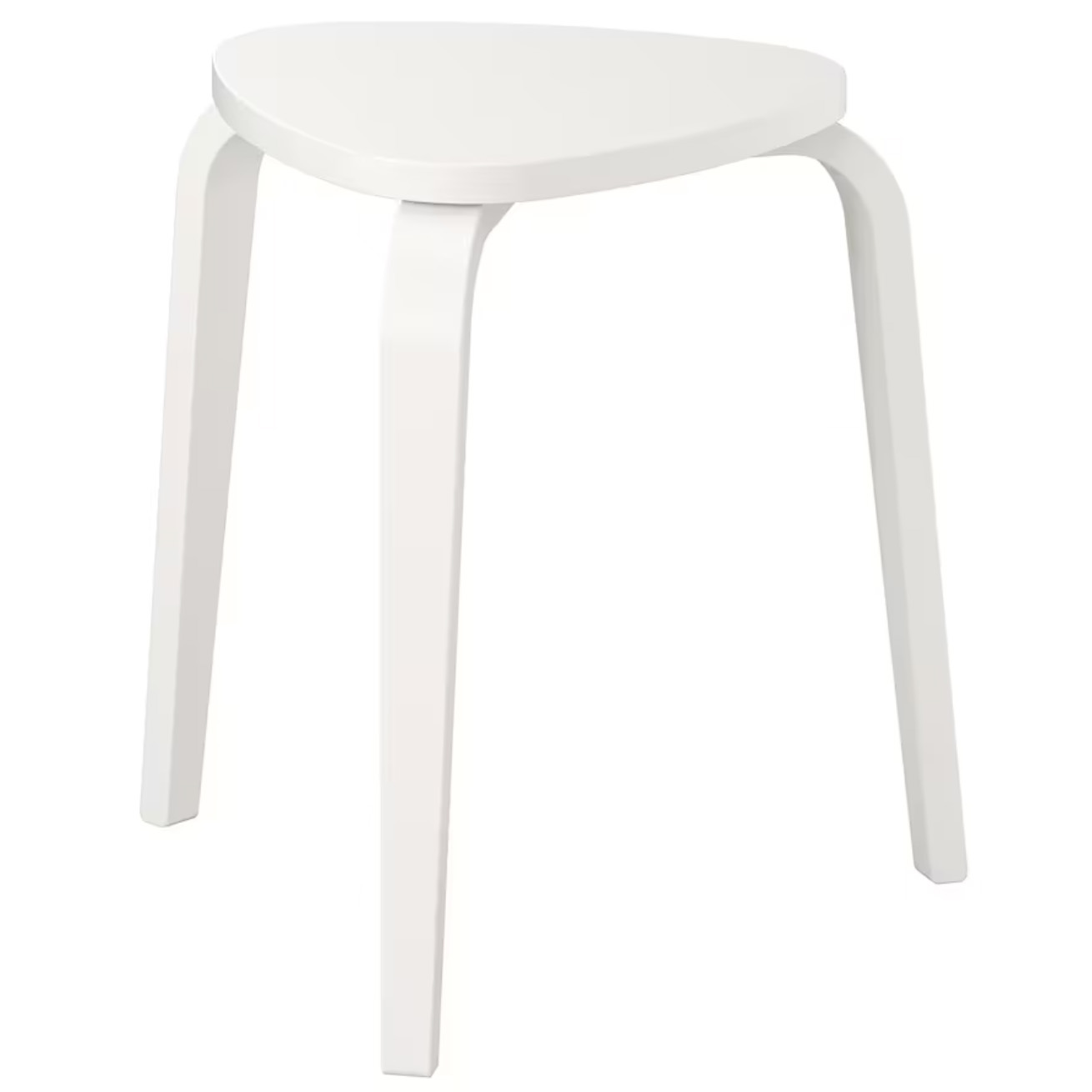} \hfill
        \includegraphics[width=0.32\linewidth]{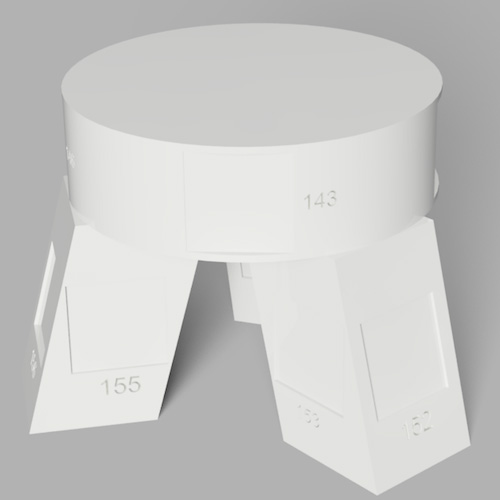} \hfill
        \includegraphics[width=0.32\linewidth]{fig/3d_printed/stool_silver.jpg}
        \caption{IKEA KYRRE \texttt{stool}}
    \end{subfigure}
    \caption{\textbf{Furniture $3$D models.} IKEA model furniture~(left), $3$D furniture model~(middle), and $3$D printed furniture model~(right).}
    \label{fig:ikea_furniture}
\end{figure*}

Our benchmark features $8$ different furniture models, each of which is modeled after an existing piece of IKEA furniture, as shown in \Cref{fig:ikea_furniture}. Due to the feasibility of a single robotic arm, we modify some furniture pieces so that a robot can assemble them with one hand. We manually $3$D modeled all furniture using \textsc{Autodesk Fusion 360}.

Each furniture model introduces unique interactions and different levels of challenges. \Cref{fig:assembly_procedures} illustrates detailed assembly processes, rewards, and phases for all furniture models. The summaries of the assembly processes and challenges are explained below:
\begin{itemize}
    \item \texttt{lamp}: The robot needs to screw in a light bulb and then place a lamp hood on top of the bulb. The robot should perform sophisticated grasping since the bulb can be easily slipped when grasping and screwing due to the rounded shape.
    \item \texttt{square\_table}: The robot needs to screw four table legs to a tabletop. As more legs are attached, they are more likely to collide with assembled table legs. Thus, careful control to avoid collisions is necessary to solve the task.
    \item \texttt{desk}: This task is similar to \texttt{square\_table}; but the legs are larger and longer, making it possibly more challenging as the robot needs to more precise positioning and orienting its hand before grasping.
    \item \texttt{drawer}: The robot needs to insert and push two drawer boxes into a drawer. The careful alignment of the rails on the drawer is necessary to push the boxes successfully.
    \item \texttt{cabinet}: The robot must first insert two doors into the poles on each side of the body. Next, it must lock the top, so the doors do not slide out. This task requires careful control to align the doors and slide into the pole. Moreover, diverse skills such as flipping the cabinet body, and screwing the top are also needed to accomplish the task.
    \item \texttt{round\_table}: The robot should assemble one rounded tabletop, rounded leg, and cross-shaped table base. The robot should handle an easily movable round leg and cross-shaped table base, which requires finding a careful grasping point.
    \item \texttt{stool}: The task consists of one round seat and three tilted legs. Since the legs are not straight, the screwing requires a good grasping pose to assemble the parts robustly. Moreover, finding a collision-free path is necessary since the legs are close to each other when they are assembled.
    \item \texttt{chair}: The task consists of one chair seat, chair back, table legs, and chair nuts (6 parts total). This task requires an accurate and complex interaction with many parts (e.g., screwing, insertion) and diverse strategies in every assembly stage.
\end{itemize}

\begin{figure*}[ht]
\centering

\begin{subfigure}[t]{\linewidth}
    \centering
    \includegraphics[width=\linewidth]{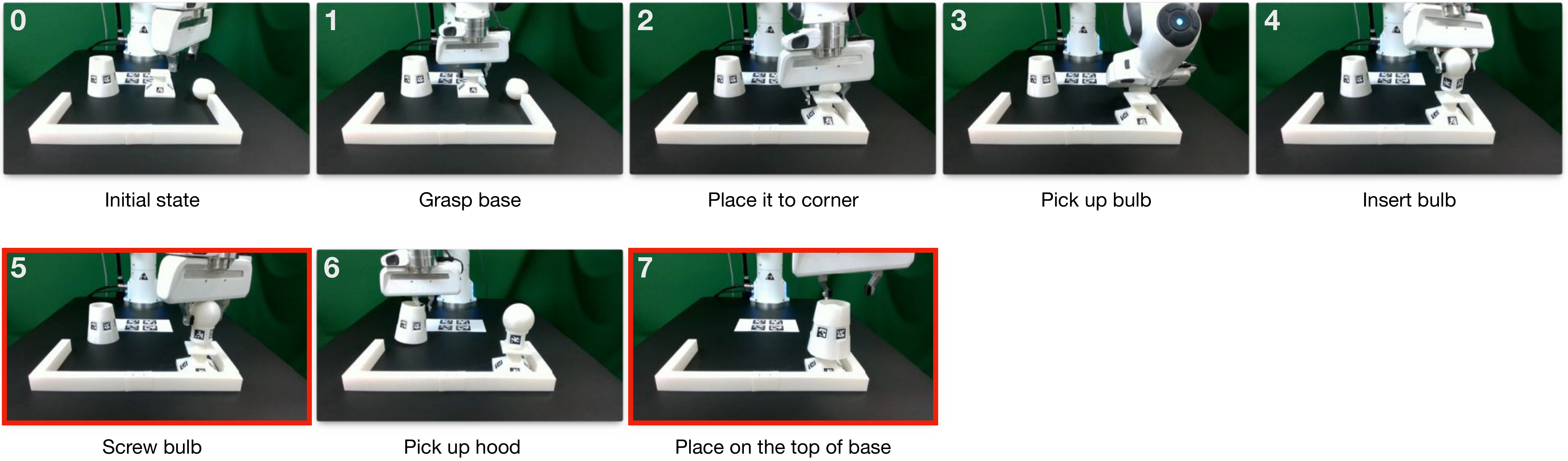}
    \caption{Assembly procedure of \texttt{lamp}}
    \label{fig:assembly:lamp}
\end{subfigure}
\vspace{1em}
\\
\begin{subfigure}[t]{\linewidth}
    \centering
    \includegraphics[width=\linewidth]{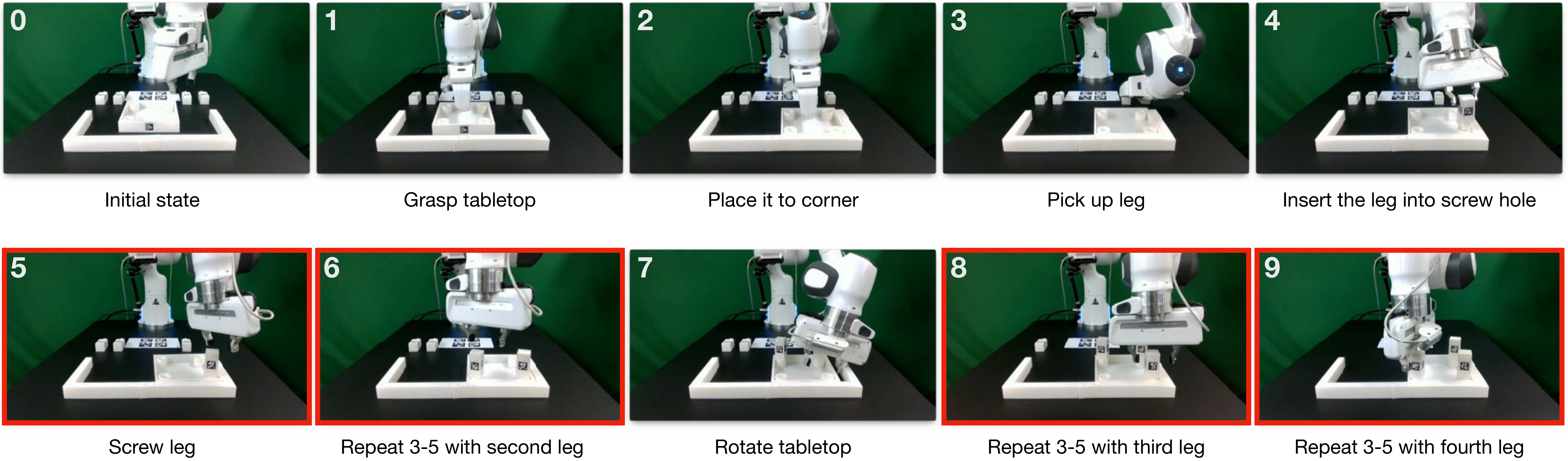}
    \caption{Assembly procedure of \texttt{square\_table}}
    \label{fig:assembly:square_table}
\end{subfigure}
\vspace{1em}
\\
\begin{subfigure}[t]{\linewidth}
    \centering
    \includegraphics[width=\linewidth]{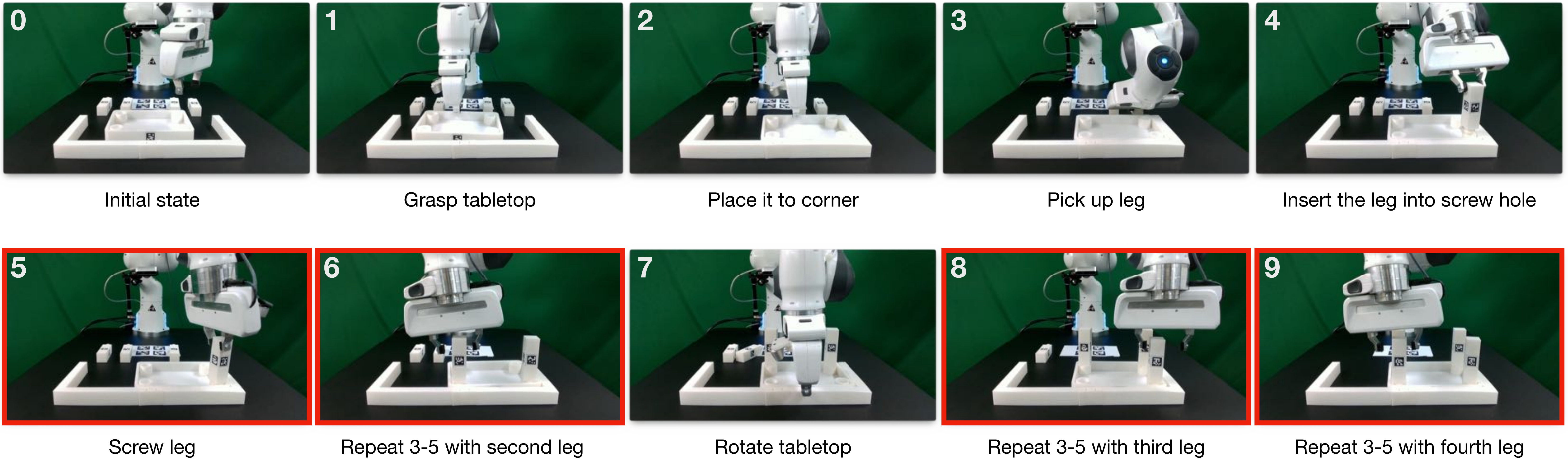}
    \caption{Assembly procedure of \texttt{desk}}
    \label{fig:assembly:desk}
\end{subfigure}

\caption{\textbf{Assembly procedures}: (a)~\texttt{lamp} (b)~\texttt{square\_table} (c)~\texttt{desk}. Numbers on the top left corners specify the phase, and red boxes indicate when a reward signal is provided.}
\end{figure*}%
\begin{figure*}[ht]\ContinuedFloat

\begin{subfigure}[t]{\linewidth}
    \centering
    \includegraphics[width=\linewidth]{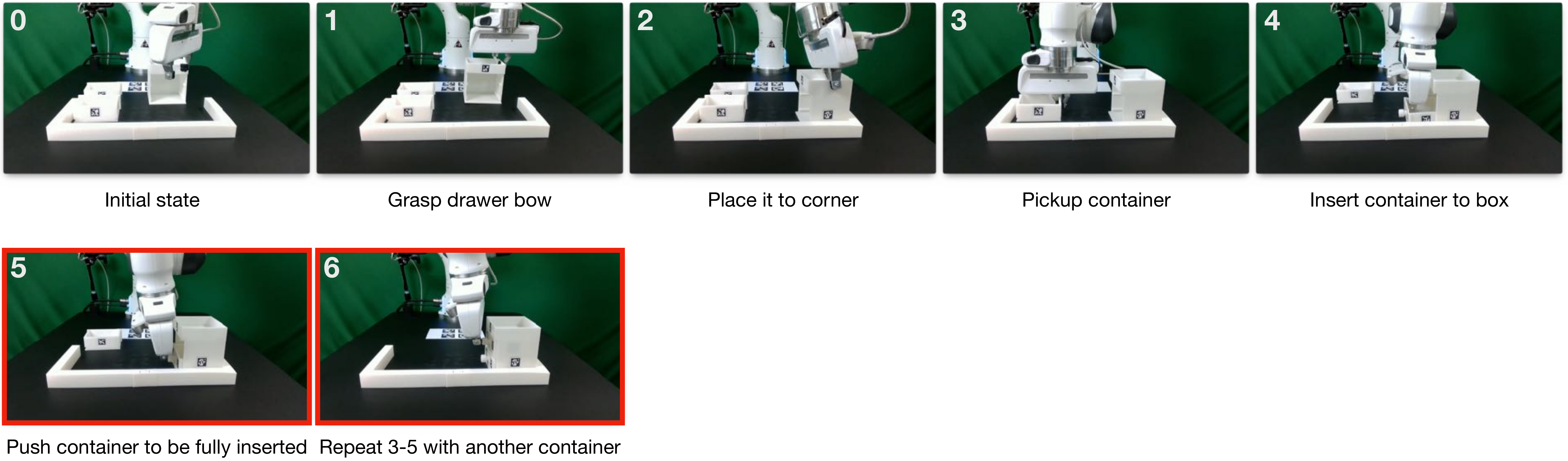}
    \caption{Assembly procedure of \texttt{drawer}}
    \label{fig:assembly:drawer}
\end{subfigure}
\vspace{1em}
\\
\begin{subfigure}[t]{\linewidth}
    \centering
    \includegraphics[width=\linewidth]{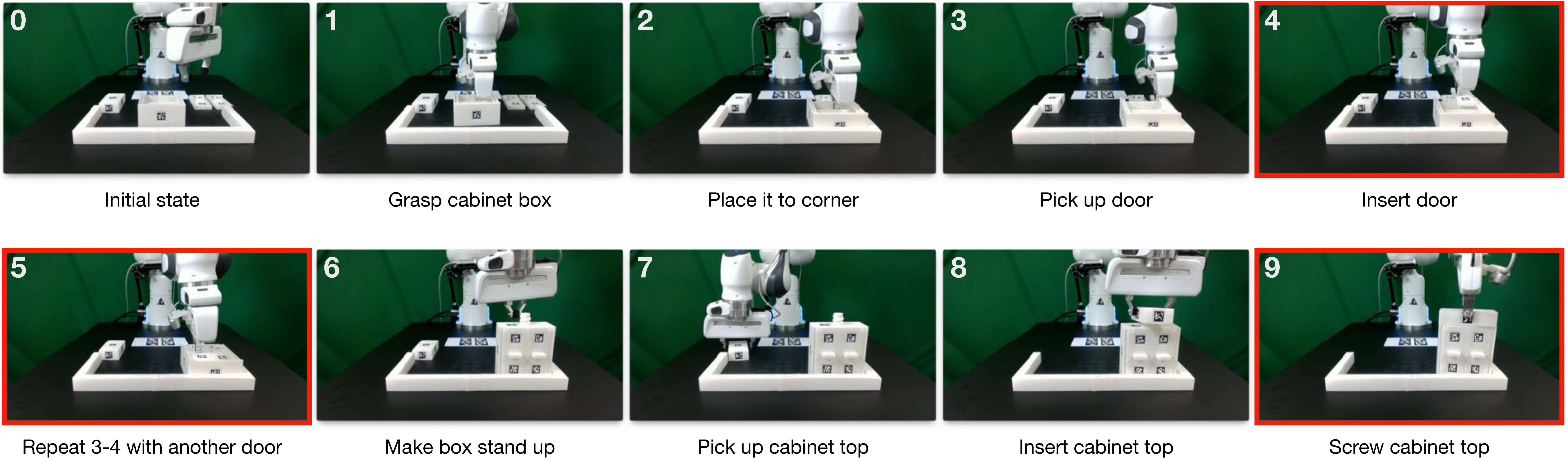}
    \caption{Assembly procedure of \texttt{cabinet}}
    \label{fig:assembly:cabinet}
\end{subfigure}
\vspace{1em}
\\
\begin{subfigure}[t]{\linewidth}
    \centering
    \includegraphics[width=\linewidth]{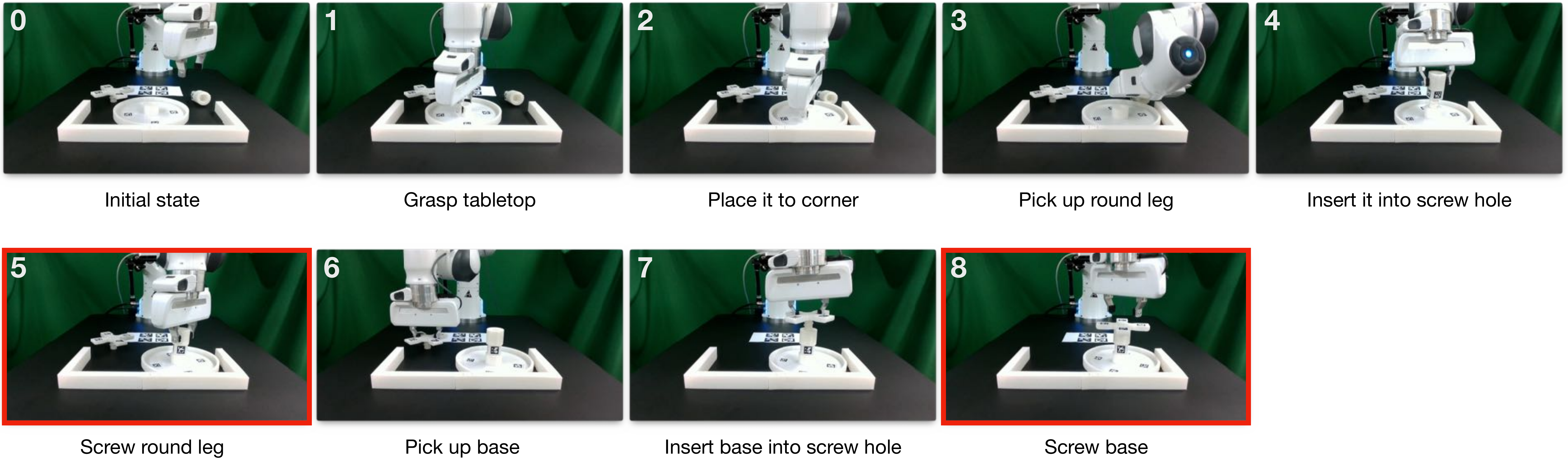}
    \caption{Assembly procedure of \texttt{round\_table}}
    \label{fig:assembly:round_table}
\end{subfigure}

\caption{\textbf{Assembly procedures}: (d)~\texttt{drawer} (e)~\texttt{cabinet} (f)~\texttt{round\_table}. Numbers on the top left corners specify the phase, and red boxes indicate when a reward signal is provided.}
\end{figure*}%
\begin{figure*}[ht]\ContinuedFloat

\begin{subfigure}[t]{\linewidth}
    \centering
    \includegraphics[width=\linewidth]{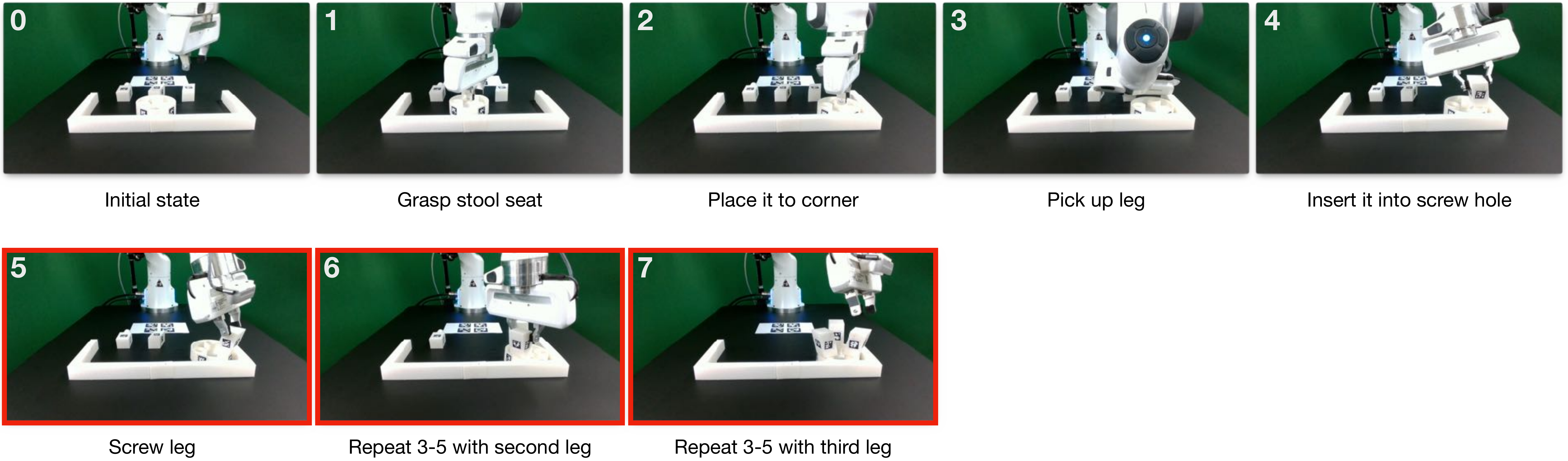}
    \caption{Assembly procedure of \texttt{stool}}
    \label{fig:assembly:stool}
\end{subfigure}
\vspace{1em}
\\
\begin{subfigure}[t]{\linewidth}
    \centering
    \includegraphics[width=\linewidth]{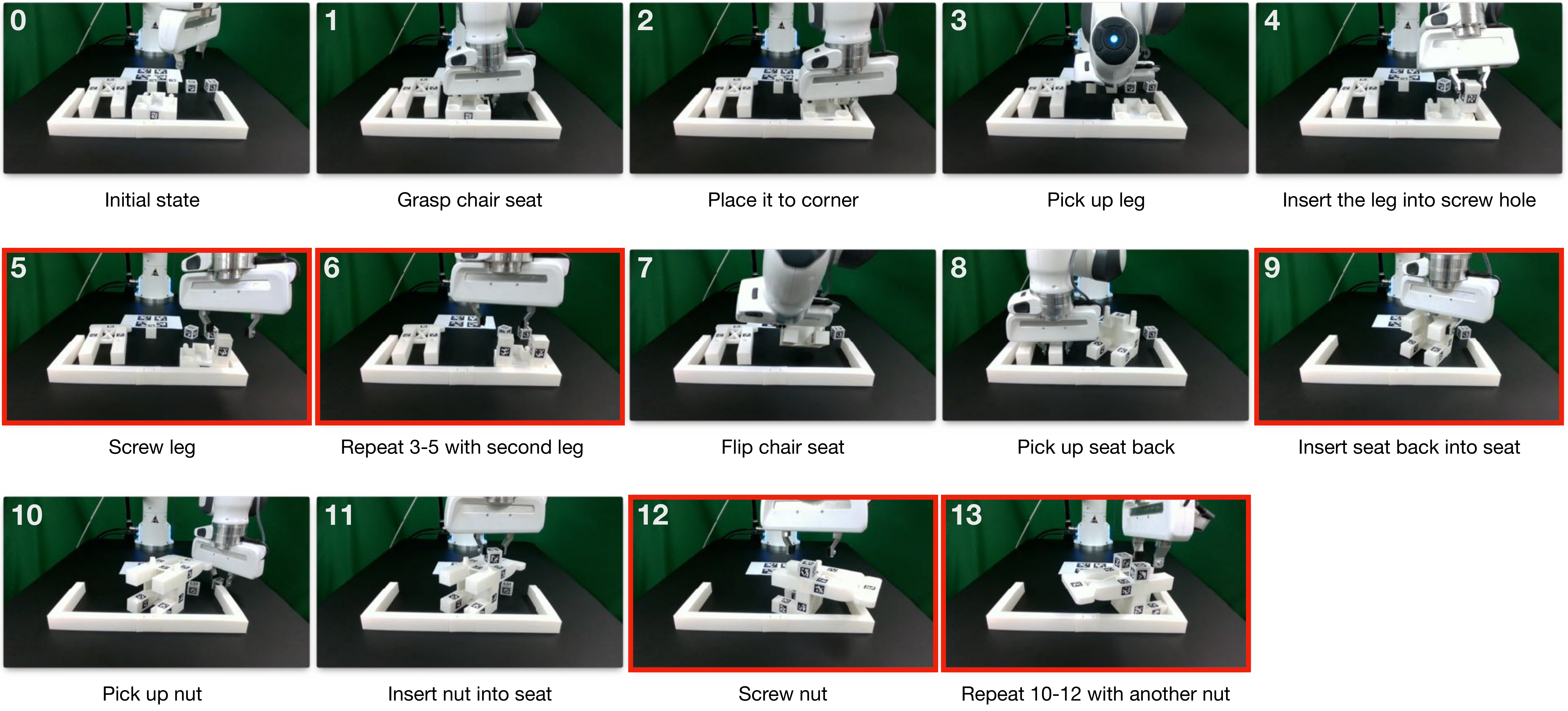}
    \caption{Assembly procedure of \texttt{chair}}
    \label{fig:assembly:chair}
\end{subfigure}

\caption{\textbf{Assembly procedures}: (g)~\texttt{stool} (h)~\texttt{chair}. Numbers on the top left corners specify the phase, and red boxes indicate when a reward signal is provided.}
\label{fig:assembly_procedures}
\end{figure*}

\clearpage

\begin{figure*}[ht]
    \begin{subfigure}[t]{0.33\linewidth}
        \includegraphics[width=\linewidth]{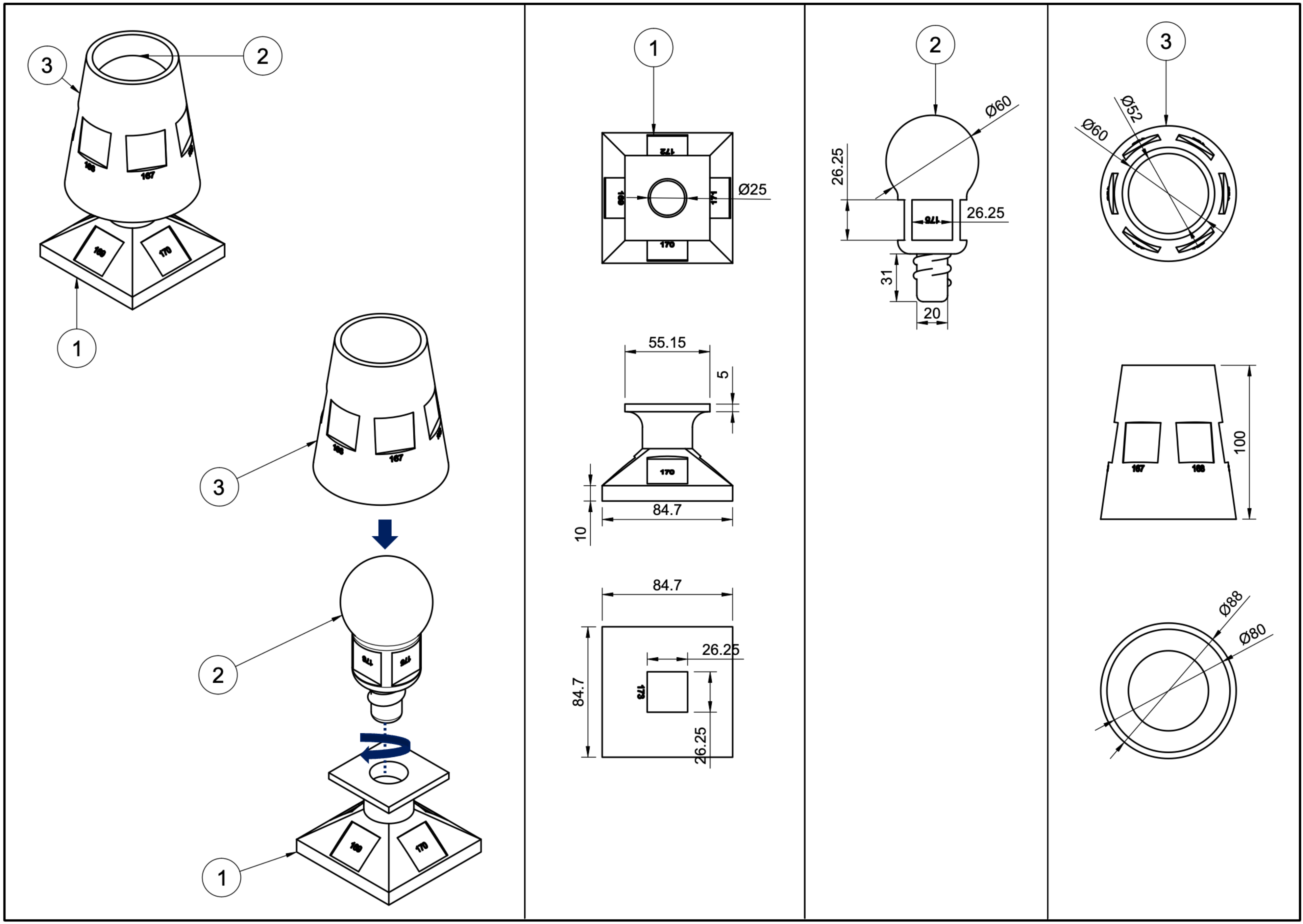}
        \vspace{-1.5em}
        \caption{\texttt{lamp}}
    \end{subfigure}
    \begin{subfigure}[t]{0.33\linewidth}
        \includegraphics[width=\linewidth]{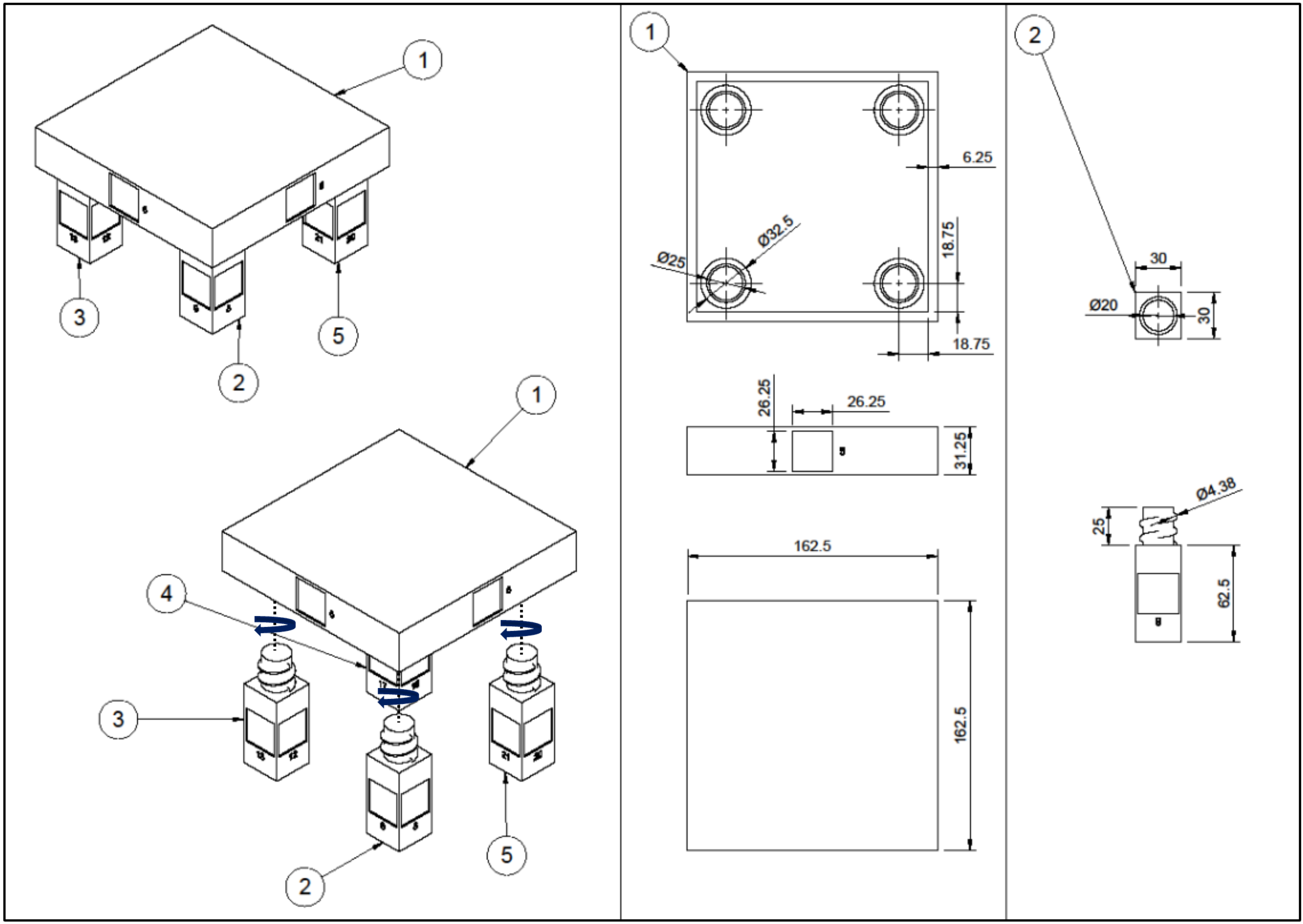}
        \vspace{-1.5em}
        \caption{\texttt{square\_table}}
    \end{subfigure}
    \begin{subfigure}[t]{0.33\linewidth}
        \includegraphics[width=\linewidth]{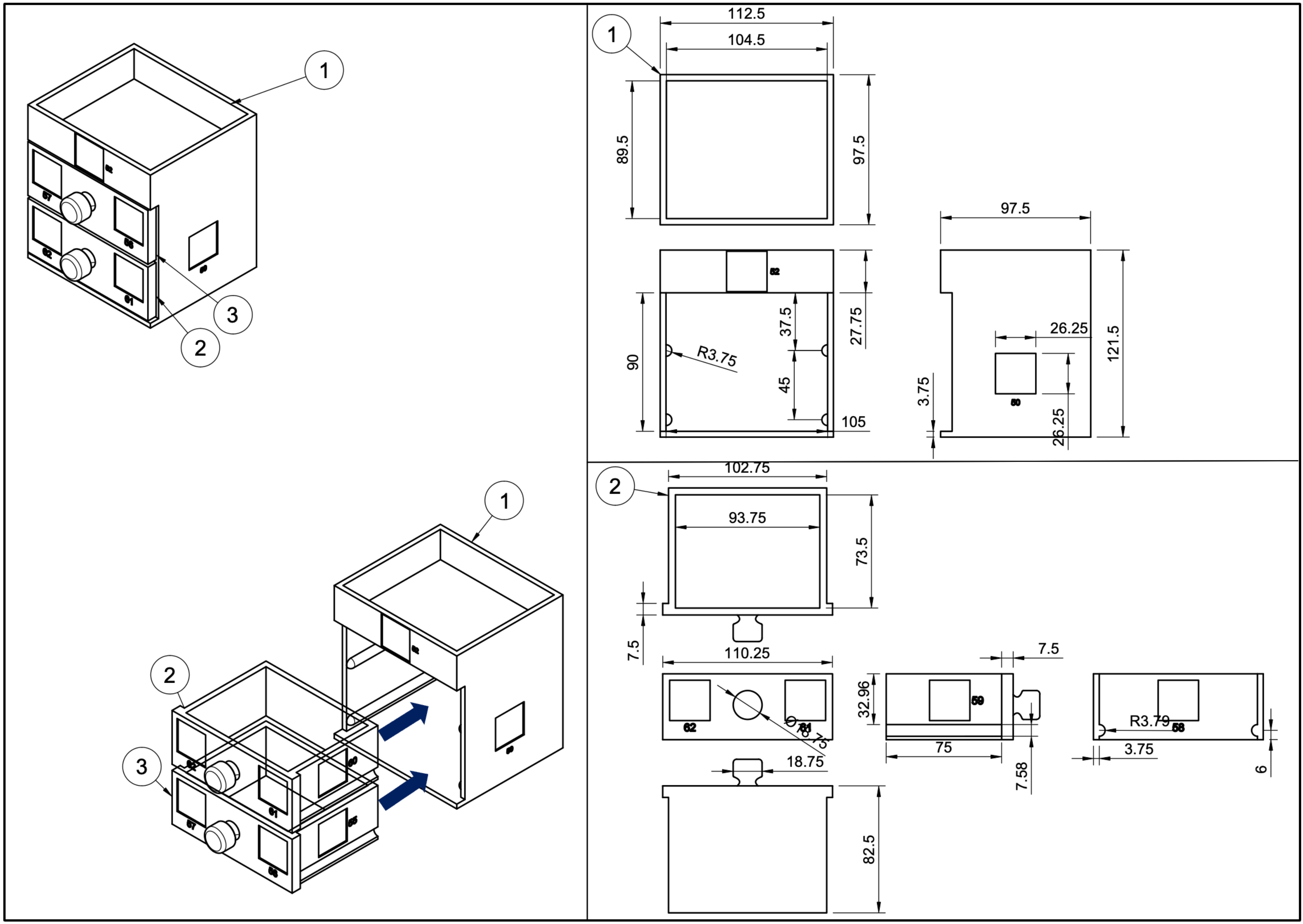}
        \vspace{-1.5em}
        \caption{\texttt{drawer}}
    \end{subfigure}
    \vspace{1em}
    \\    
    \begin{subfigure}[t]{0.33\linewidth}
        \includegraphics[width=\linewidth]{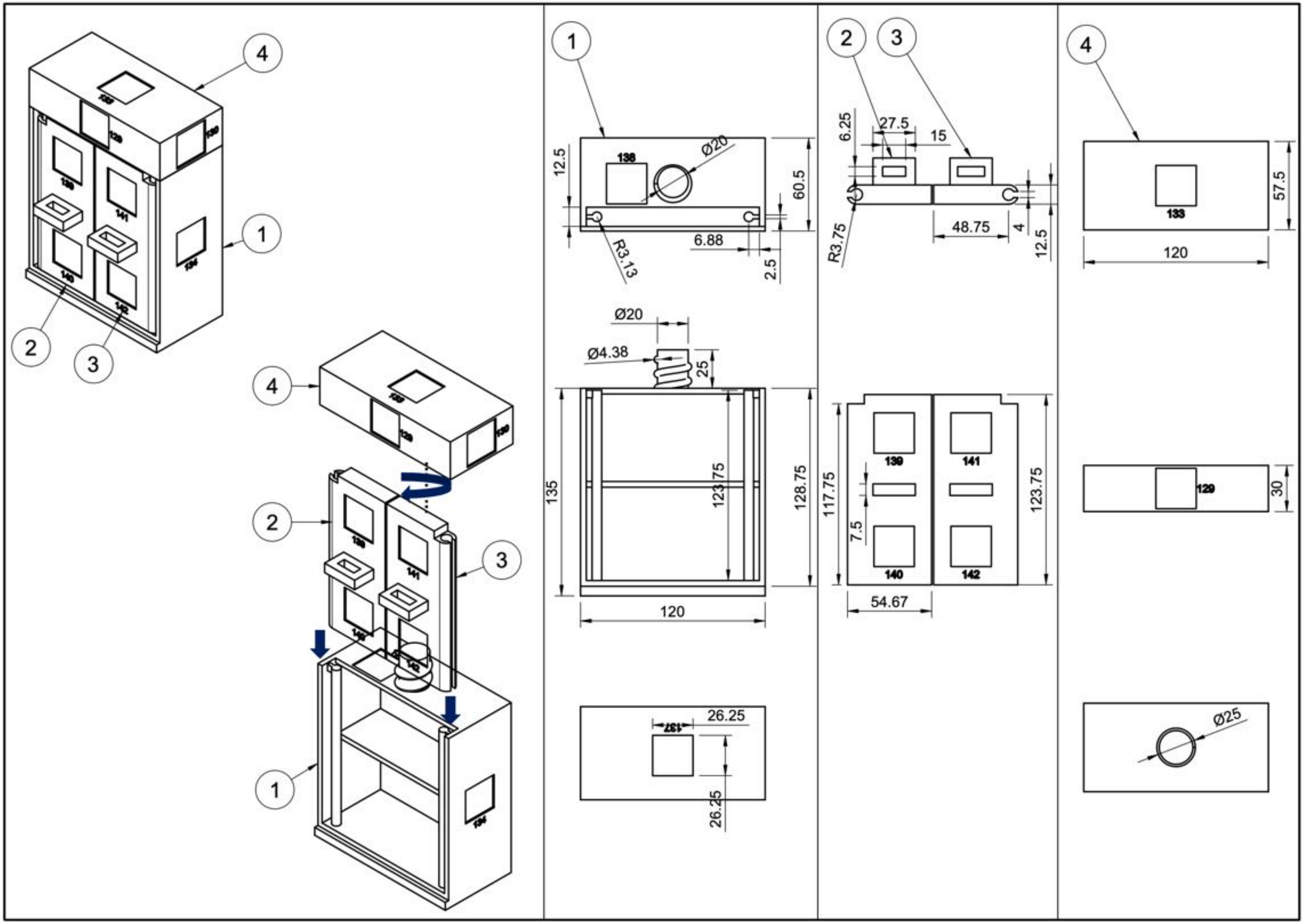}
        \vspace{-1.5em}
        \caption{\texttt{cabinet}}
    \end{subfigure}
    \begin{subfigure}[t]{0.33\linewidth}
        \includegraphics[width=\linewidth]{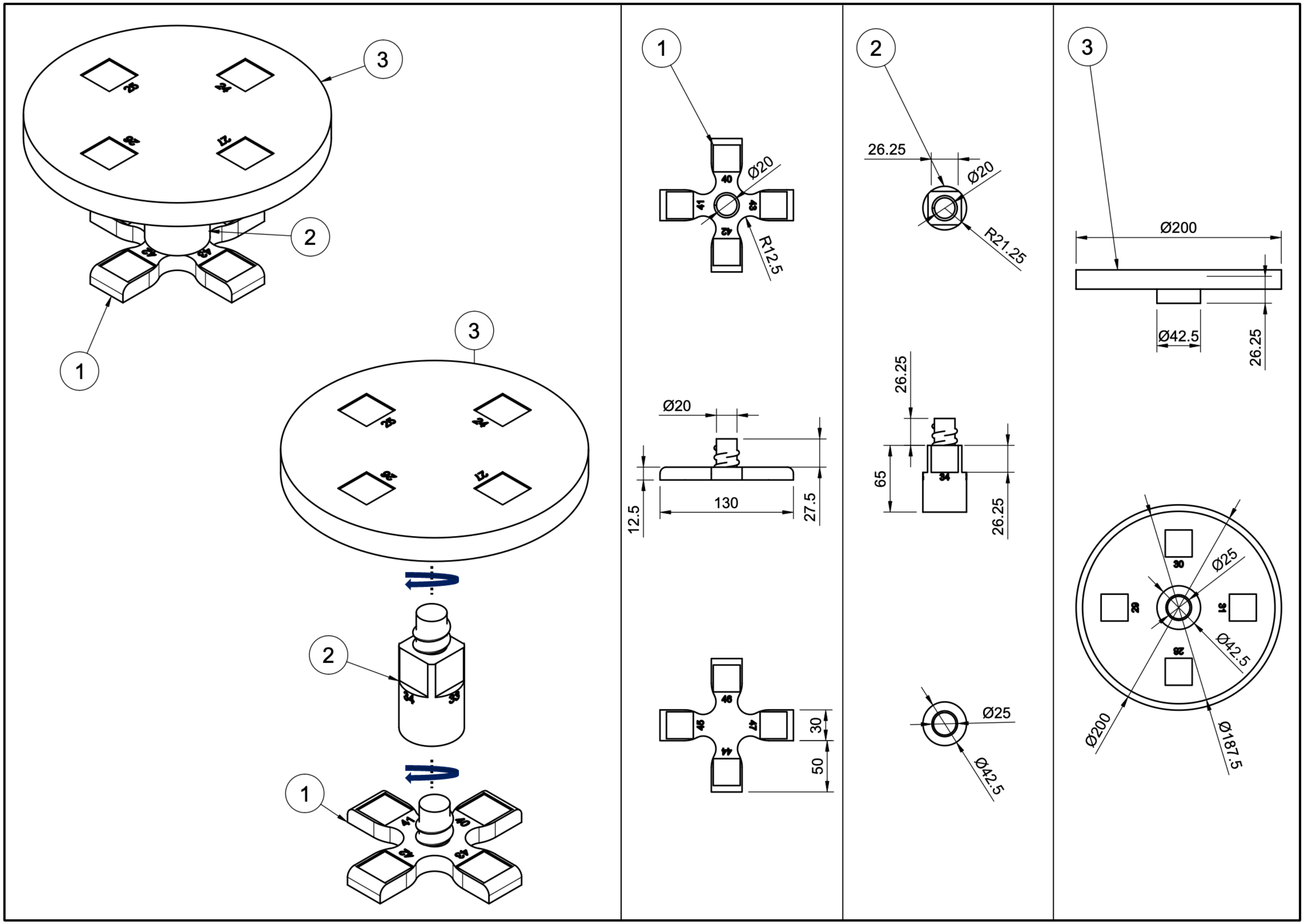}
        \vspace{-1.5em}
        \caption{\texttt{round\_table}}
    \end{subfigure}
    \begin{subfigure}[t]{0.33\linewidth}
        \includegraphics[width=\linewidth]{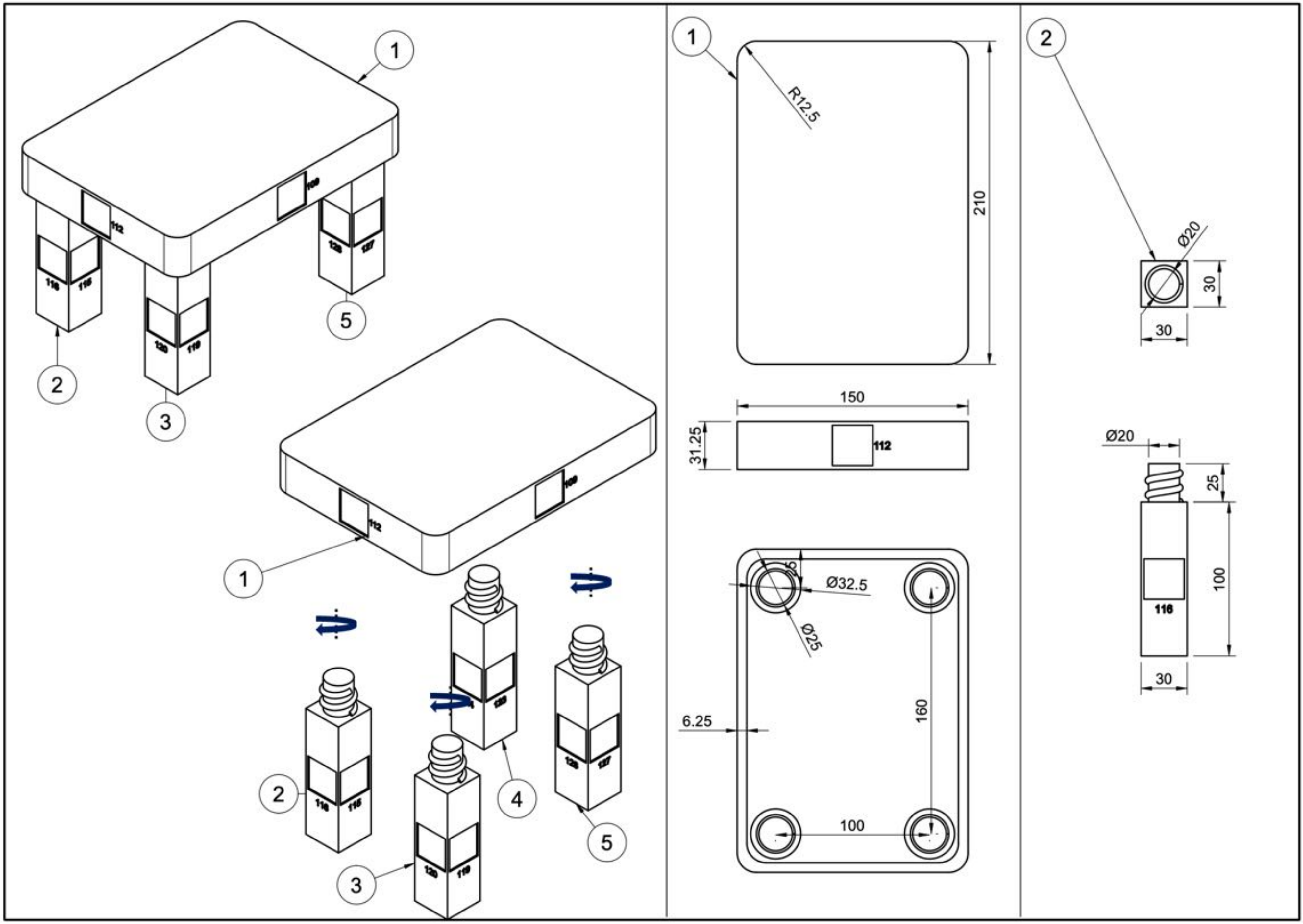}
        \vspace{-1.5em}
        \caption{\texttt{desk}}
    \end{subfigure}
    \vspace{1em}
    \\
    \begin{subfigure}[t]{0.33\linewidth}
        \includegraphics[width=\linewidth]{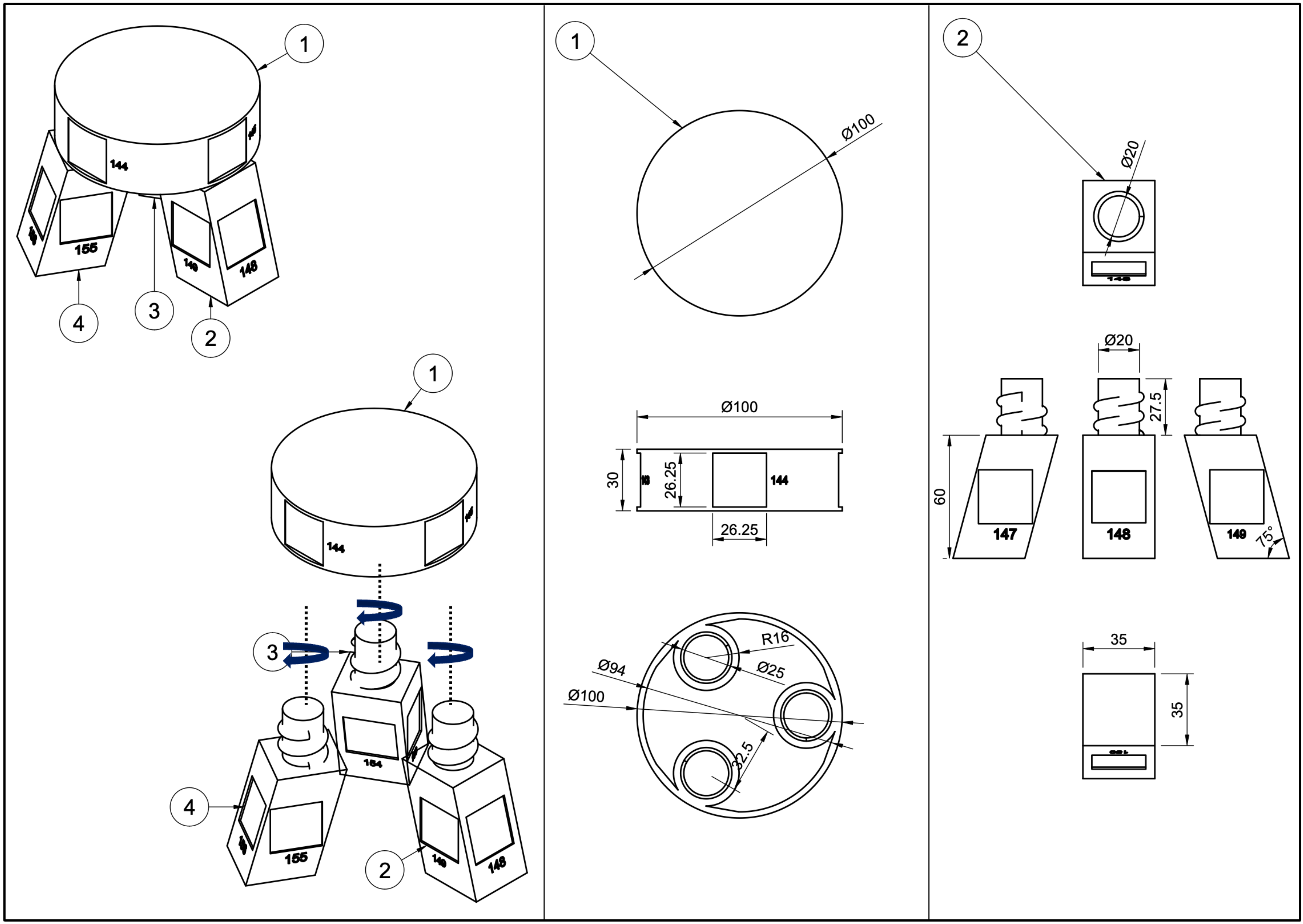}
        \vspace{-1.5em}
        \caption{\texttt{stool}}
    \end{subfigure}
    \begin{subfigure}[t]{0.33\linewidth}
        \includegraphics[width=\linewidth]{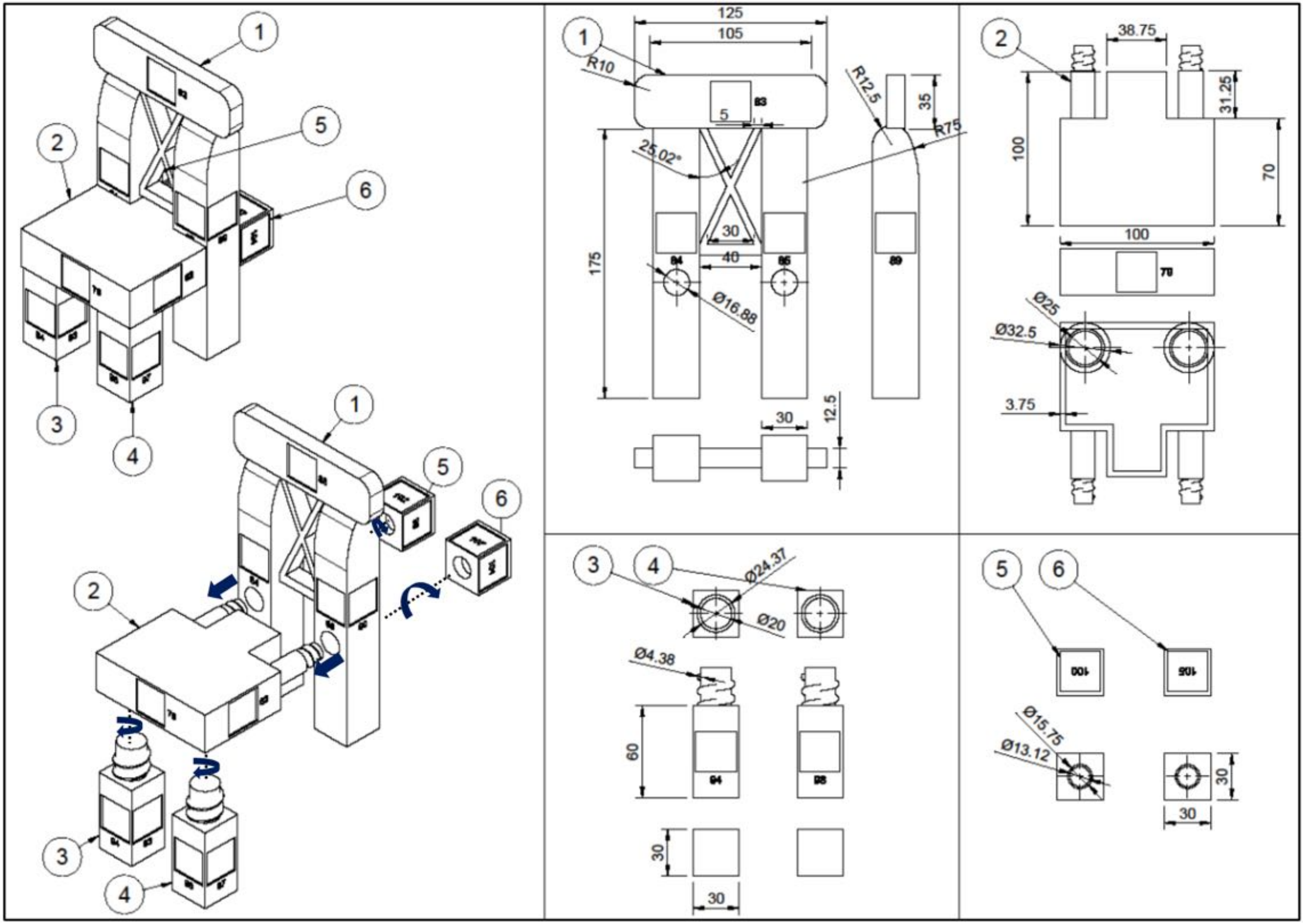}
        \vspace{-1.5em}
        \caption{\texttt{chair}}
    \end{subfigure}
    \caption{\textbf{Blueprints of furniture models.} The leftmost column shows the final configuration (top) and how furniture parts are assembled (bottom). The rest of the columns illustrate the dimensions of all furniture parts.}
    \label{fig:blueprints}
\end{figure*}

The blueprints of furniture models are shown in \Cref{fig:blueprints}. Some small or thin parts are designed to be thicker than \SI{3}{\cm} to afford space for an AprilTag marker. The largest dimension is smaller than \SI{21}{\cm}, so that our furniture models are compatible with most $3$D printers. The potential grasping regions are at most \SI{6}{\cm} (\texttt{lamp}) wide, which is smaller than the open gripper width (\SI{7.0}{\cm} for \texttt{lamp} and \SI{6.5}{\cm} for other furniture). In our experiments, the $3$D models are printed using a \textsc{FlashForge Guider IIs} $3$D printer and white PLA filaments. Printing a small furniture model takes approximately 10 hours, while large furniture models, such as \texttt{desk}, need two passes of $3$D printing, which takes about 24 hours.

\clearpage

\section{Qualitative Results}
\label{sec:qualitative_results}

In \Cref{fig:policy_sequences}, we present the qualitative results of the evaluated methods, BC and IQL, with their best-performing trials on the low or medium randomness levels. The trained agents are generally capable of completing the first pick and place phases but struggle with more challenging tasks, such as insertion. More videos can be found in \href{https://clvrai.com/furniture-bench}{our website}

\begin{figure*}[ht]
    \begin{subfigure}[t]{\linewidth}
        \includegraphics[width=0.85\linewidth]{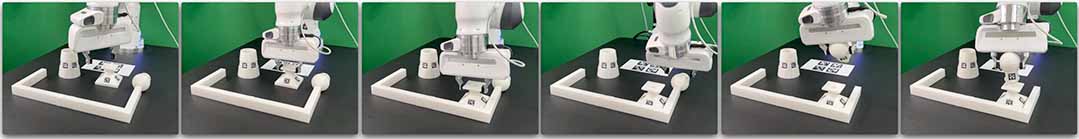}
        \vspace{1em}
        \\
        \includegraphics[width=0.57\linewidth]{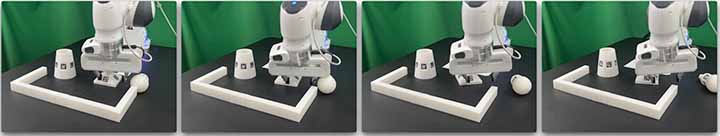}
        \caption{\textbf{Qualitative results on \texttt{lamp}.} (top) IQL successfully inserts the bulb into the lamp base, but it stops moving afterward. (bottom) BC struggles to grasp the lamp base due to small grasping points.}
    \end{subfigure}
    \vspace{1em}
    \\
    
    \begin{subfigure}[t]{\linewidth}
        \includegraphics[width=0.85\linewidth]{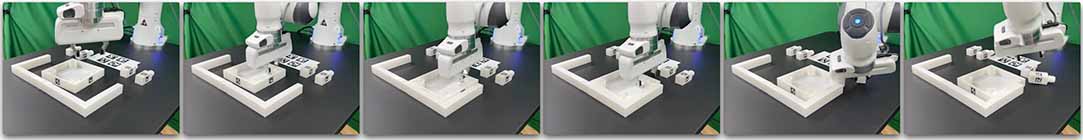}
        \vspace{1em}
        \\
        \includegraphics[width=0.57\linewidth]{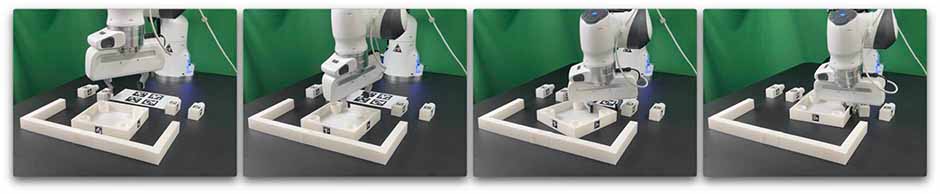}
        \caption{\textbf{Qualitative results on \texttt{square\_table}.} (top) IQL places the tabletop correctly, picks up the leg, but fails the insertion.
        (bottom) BC grasps the tabletop but fails to place it to the corner.}
    \end{subfigure}
    \vspace{1em}
    \\

    \begin{subfigure}[t]{\linewidth}
        \includegraphics[width=0.85\linewidth]{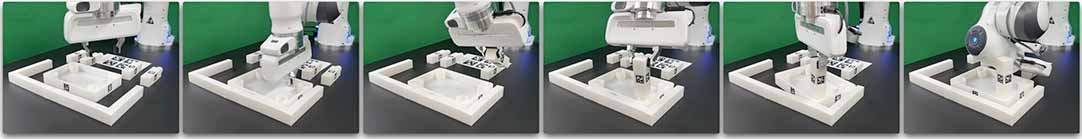}
        \vspace{1em}
        \\
        \includegraphics[width=0.71\linewidth]{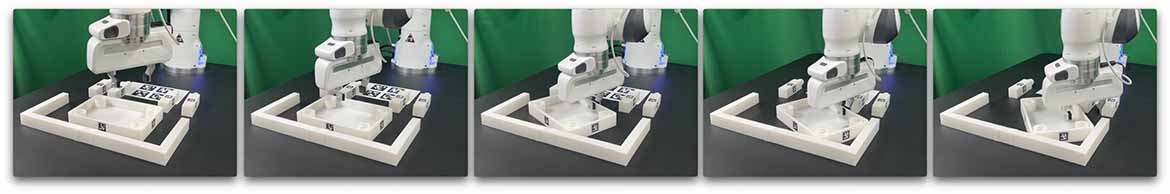}
        \caption{\textbf{Qualitative results on \texttt{desk}.} (top) IQL screws one leg and then grasps the other. (bottom) BC places the desk top in the corner but stops the motion after releasing the gripper.}
    \end{subfigure}

    \caption{\textbf{Sequences of policy evaluation}: (a-b)~\texttt{lamp} (c-d)~\texttt{square\_table} (e-f)~\texttt{desk}}
    
\end{figure*}%
\begin{figure*}[ht]\ContinuedFloat
        
    \begin{subfigure}[t]{\linewidth}
        \includegraphics[width=\linewidth]{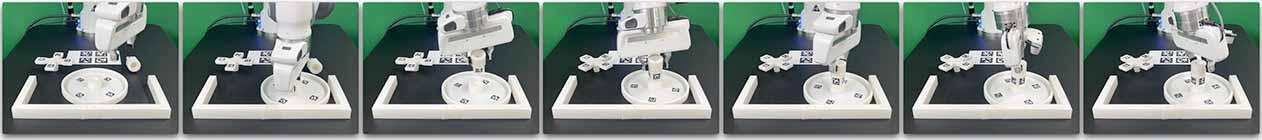}
        \vspace{1em}
        \\
        \includegraphics[width=0.57\linewidth]{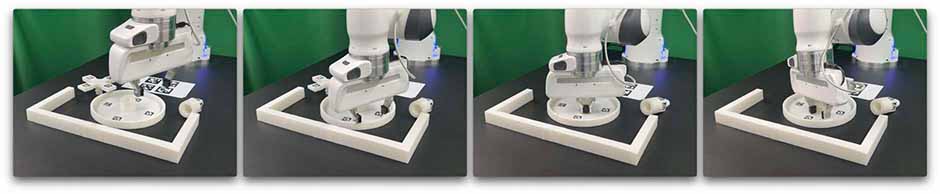}
        \caption{\textbf{Qualitative results on \texttt{round\_table}.} (top) IQL repeats the screwing even after the screw is tightened. (bottom) BC fails to grasp the round table top due to the small grasping region.}
    \end{subfigure}
    \vspace{1em}\\

    \begin{subfigure}[t]{\linewidth}
        \includegraphics[width=\linewidth]{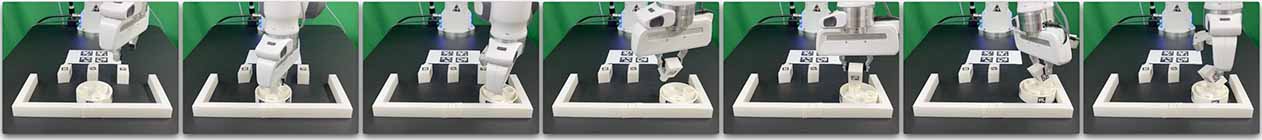}
        \vspace{1em}
        \\
        \includegraphics[width=0.86\linewidth]{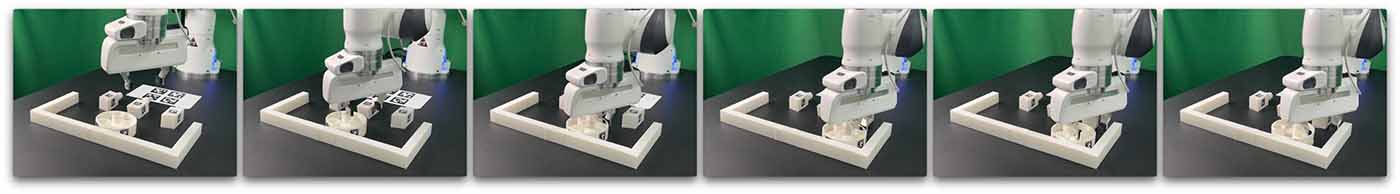}
        \caption{\textbf{Qualitative results on \texttt{stool}.} IQL tries to screw the leg but fails due to the tilted shape. (bottom) BC pushes the seat to the corner but stops moving.}
    \end{subfigure}
    \vspace{1em}\\

    \begin{subfigure}[t]{\linewidth}
        \includegraphics[width=\linewidth]{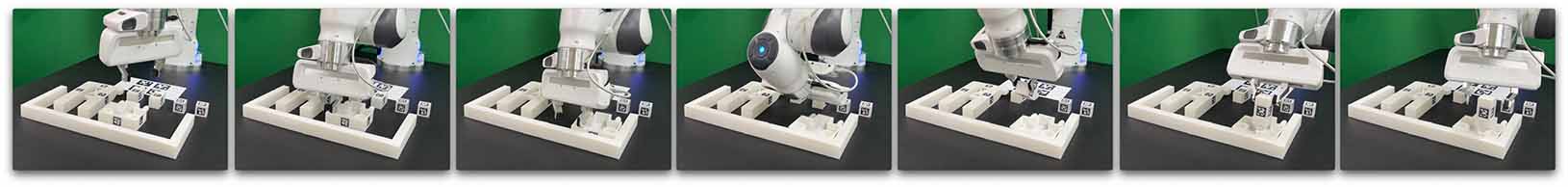}
        \vspace{1em}
        \\
        \includegraphics[width=0.57\linewidth]{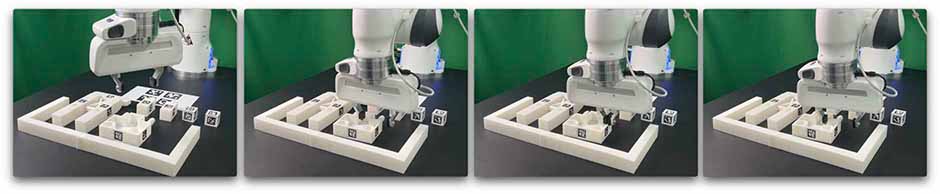}
        \caption{\textbf{Qualitative results on \texttt{chair}.} (top) IQL insertion fails due to misalignment of the hole.
        (bottom) BC fails to align the chair seat to the corner.}
    \end{subfigure}
    
    \caption{\textbf{Sequences of policy evaluation}: (g-h)~\texttt{round\_table} (i-j)~\texttt{stool} (k-l)~\texttt{chair}}
\end{figure*}%
\begin{figure*}[ht]\ContinuedFloat
    
    \begin{subfigure}[t]{\linewidth}
        \includegraphics[width=0.86\linewidth]{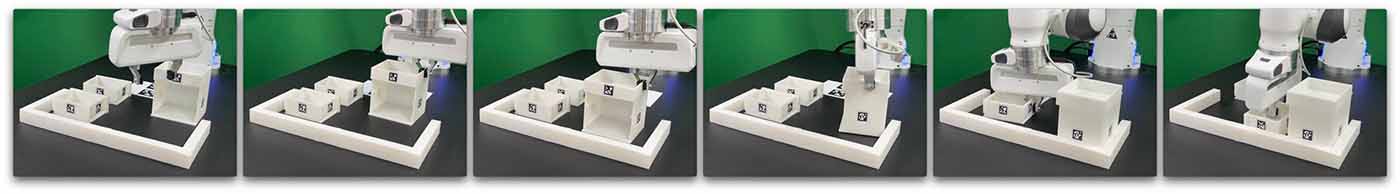}
        \vspace{1em}
        \\
        \includegraphics[width=\linewidth]{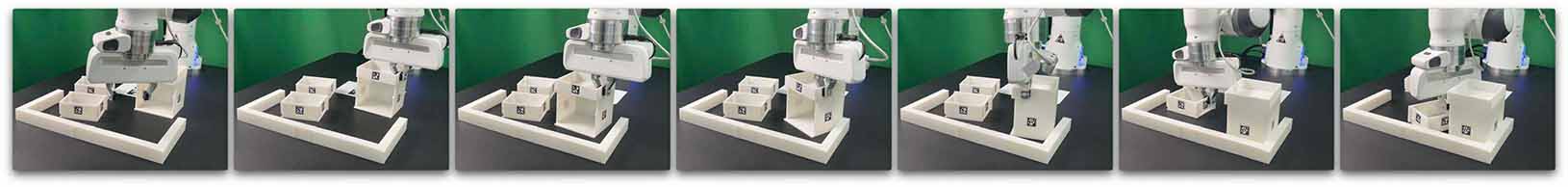}
        \caption{\textbf{Qualitative results on \texttt{drawer}.} (top) IQL fails on inserting the drawer box. (bottom) BC fails on inserting the drawer box.}
    \end{subfigure}
    \vspace{1em}\\
    
    \begin{subfigure}[t]{\linewidth}
        \includegraphics[width=0.71\linewidth]{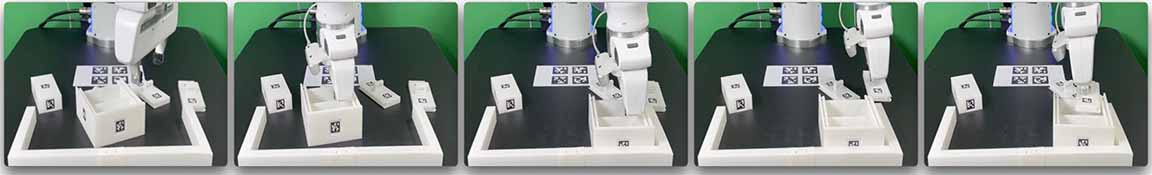}
        \vspace{1em}
        \\
        \includegraphics[width=0.86\linewidth]{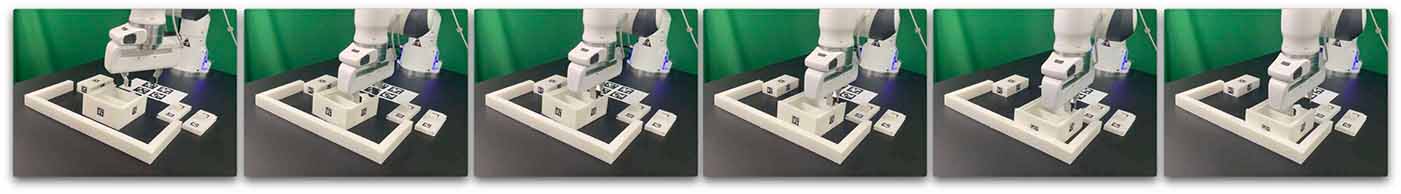}
        \caption{\textbf{Qualitative results on \texttt{cabinet}.} (top) IQL inaccurately inserts the door door. (bottom) BC halts after placing the body in the corner.}
    \end{subfigure}

    \caption{
        \textbf{Sequences of policy evaluation}: (a-b)~\texttt{lamp} (c-d)~\texttt{square\_table} (e-f)~\texttt{desk} (g-h)~\texttt{round\_table} (i-j)~\texttt{stool} (k-l)~\texttt{chair} (m-n)~\texttt{drawer} (o-p)~\texttt{cabinet}. The trajectories are selected with the best-performing rollout in the low or medium randomness level.
    }
    \label{fig:policy_sequences}
\end{figure*}
\newpage
\onecolumn

\section{Reproducible FurnitureBench Setup Instruction}
\label{sec:instruction}

This section elaborates on how to reproduce our FurnitureBench step by step (\href{https://clvrai.github.io/furniture-bench/docs}{online version}).

\textcolor{red}{\textbf{
    Please read and follow this instruction thoroughly. Examining the photos and text thoroughly is crucial for creating a reproducible environment. Attention to detail plays a vital role in ensuring precision in this process.
}}

\subsection{Environment Overview}

\begin{figure*}[ht]
    \centering
    \vspace{-1em}
    \includegraphics[width=0.9\linewidth]{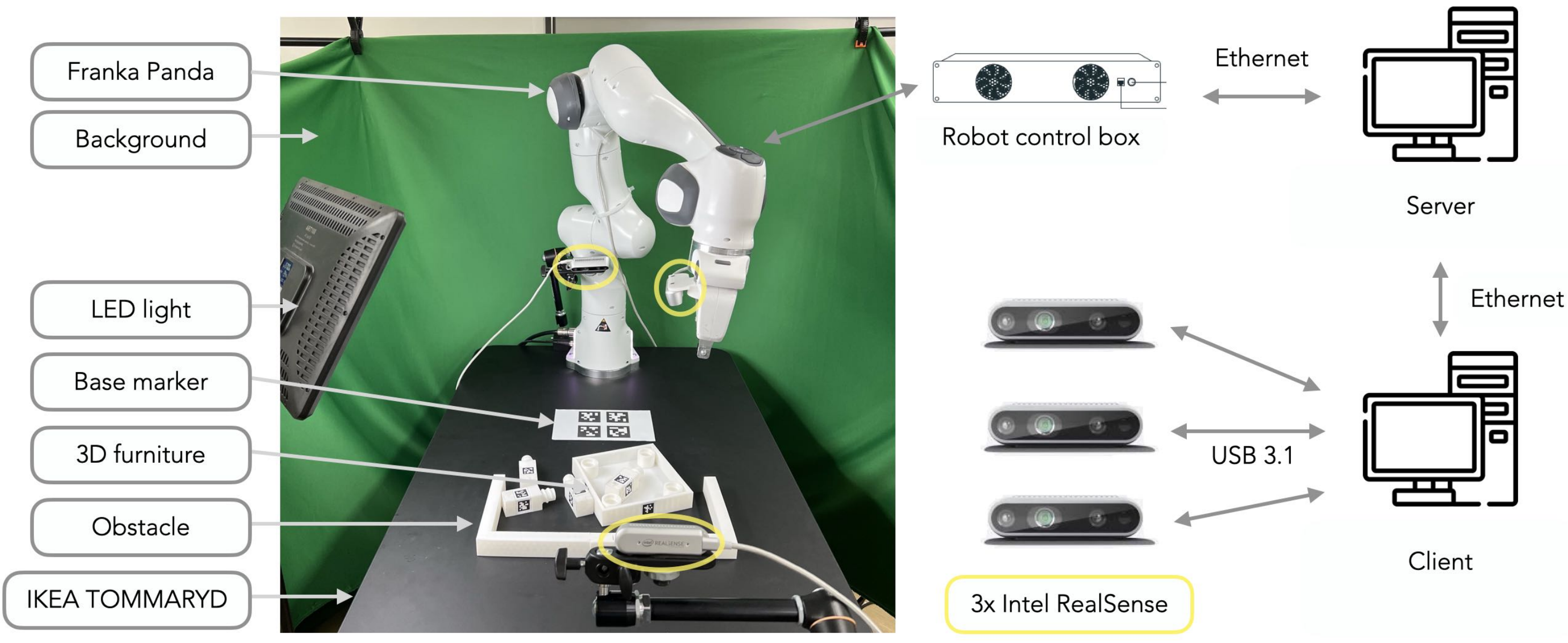}
    \caption{\textbf{FurnitureBench environment.} We illustrate our environment (left) and their connections to server and client computers (right). The client handles I/O, motion planning, and policy inference while the server manages real-time robot control.}
    \label{fig:instruction:setup_overview}
    \vspace{-1em}
\end{figure*}

\subsection{Mount Robot on Table}

\begin{figure}[ht]
    \centering
    \vspace{-1em}
    \begin{subfigure}[t]{0.18\textwidth}
        \includegraphics[width=\textwidth]{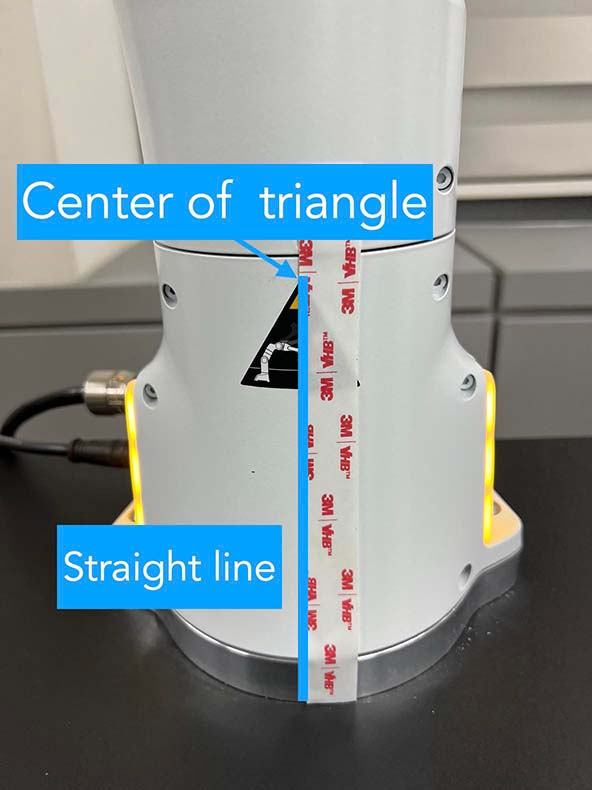}
        \caption{Center of the robot}
        \label{fig:instruction:center_of_robot_base}
    \end{subfigure}
    \begin{subfigure}[t]{0.18\textwidth}
        \centering
        \includegraphics[width=\textwidth]{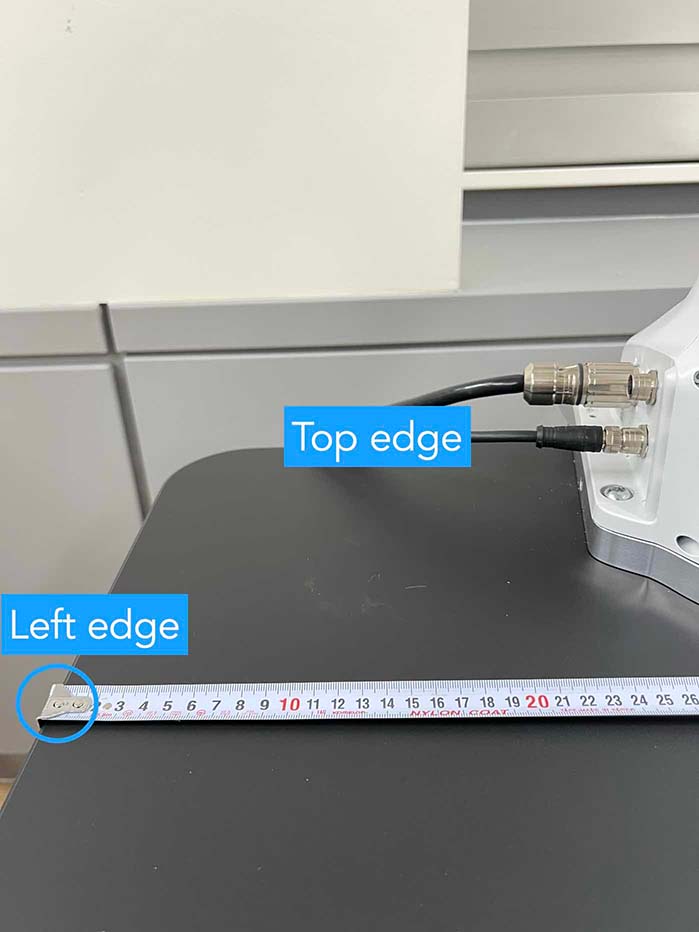}
        \caption{Ruler on table}
        \label{fig:instruction:robot_placement_ruler}
    \end{subfigure}
    \begin{subfigure}[t]{0.18\textwidth}
        \includegraphics[width=\textwidth]{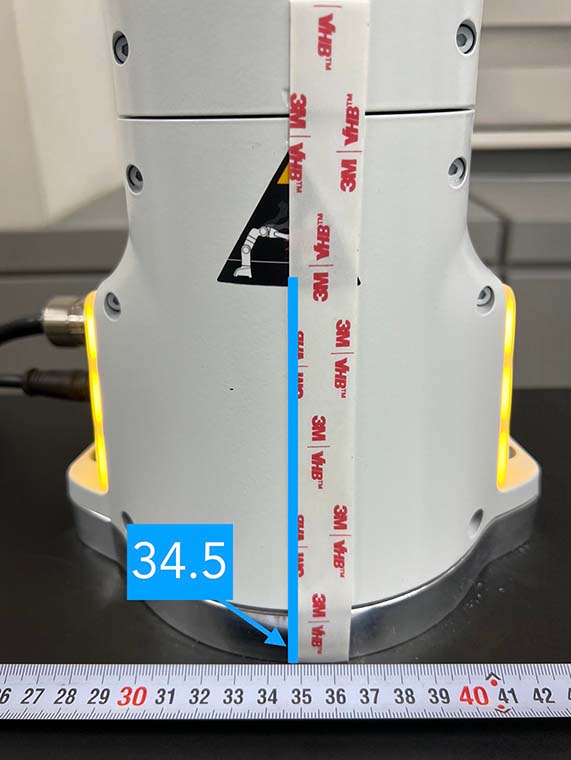}
        \caption{Robot position on table}
        \label{fig:instruction:robot_placement_horizontal}
    \end{subfigure}
    \begin{subfigure}[t]{0.18\textwidth}
        \includegraphics[width=\textwidth]{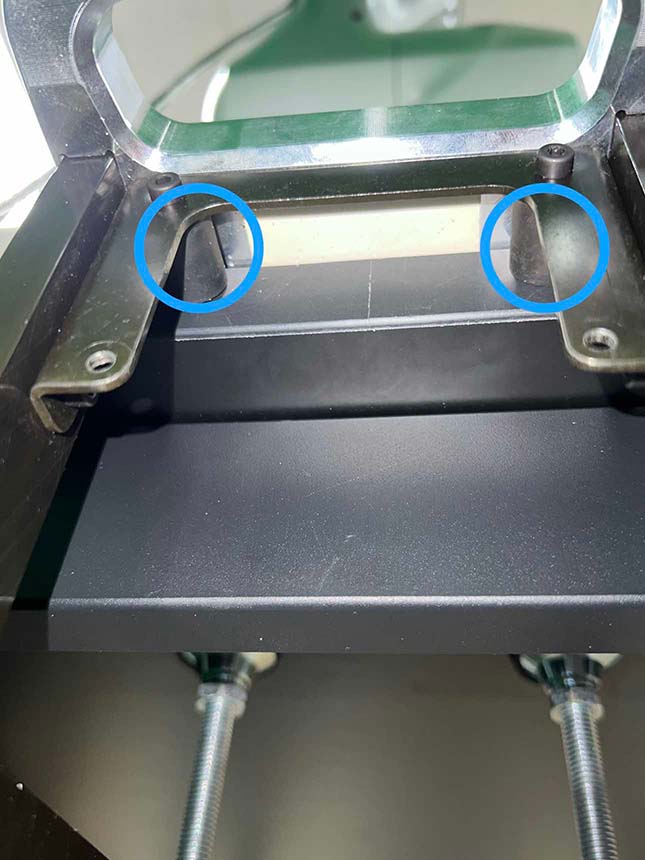}
        \caption{Two rear supports are indicated by blue circles}
        \label{fig:instruction:robot_placement_vertical}
    \end{subfigure}
    \label{fig:instruction:robot_placement}
    \begin{subfigure}[t]{0.18\textwidth}
        \includegraphics[width=\textwidth]
        {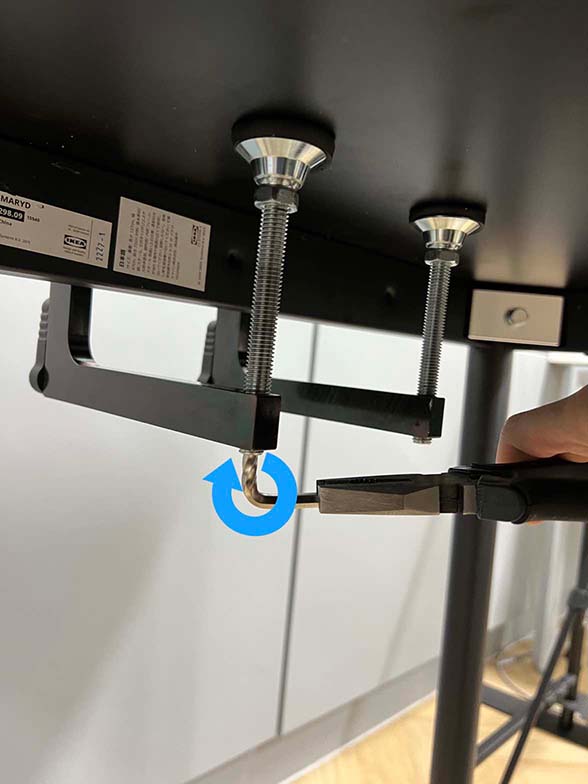}
        \caption{Screwing the mount}
        \label{fig:instruction:firm_screw}
    \end{subfigure}
    \caption{\textbf{Robot placement.}}
    \vspace{-1em}
\end{figure}

The first step is mounting the robot on the table. To place the robot precisely, follow the step-by-step instructions:
\begin{enumerate}
    \item As shown in  \Cref{fig:instruction:center_of_robot_base}, attach marking tapes to the robot body to specify its center. Make sure that the tape's left edge is in the center of the triangle sticker. The tape must be attached straight from the front view.
    \item Align one ruler so that \SI{0}{\centi\meter} is at the left edge of the table and extend it straight, as shown in \Cref{fig:instruction:robot_placement_ruler}.
    \item Place the robot center (indicated by the left edge of the tape) at point \SI{34.5}{\cm} of the ruler, as shown in \Cref{fig:instruction:robot_placement_horizontal}.
    \item Make sure the robot is tightly attached to the side of the table, with no room left between it and the table. To double-check, make sure that both rear support pads are closely pressed against the edge of the table, as shown in \Cref{fig:instruction:robot_placement_vertical}.
    \item Firmly attach the robot to the table by tightly screwing the robot mount, as shown in \Cref{fig:instruction:firm_screw}.
    \item Remove the affixed tape from the robot once this step is completed. 
\end{enumerate}

\medskip
\textbf{Important Checklist:}
\begin{Form}
\begin{itemize}
    \item Make sure the robot is installed at \textbf{\SI{34.5}{\cm}} off from the left edge of the table. \CheckBox[name=MountCheckbox1, width=0.7em, height=0.7em]{} 
    \item The robot should be tightly attached to the table \textbf{without margin}. \CheckBox[name=MountCheckbox2, width=0.7em, height=0.7em]{}
    \item The robot mount is tightly screwed.
    \CheckBox[name=MountCheckbox3, width=0.7em, height=0.7em]{}
\end{itemize}
\end{Form}

\clearpage

\subsection{Install Background}

\begin{wrapfigure}{r}{0.3\textwidth}
  \centering
  \vspace{-3em}
  \includegraphics[width=\linewidth]{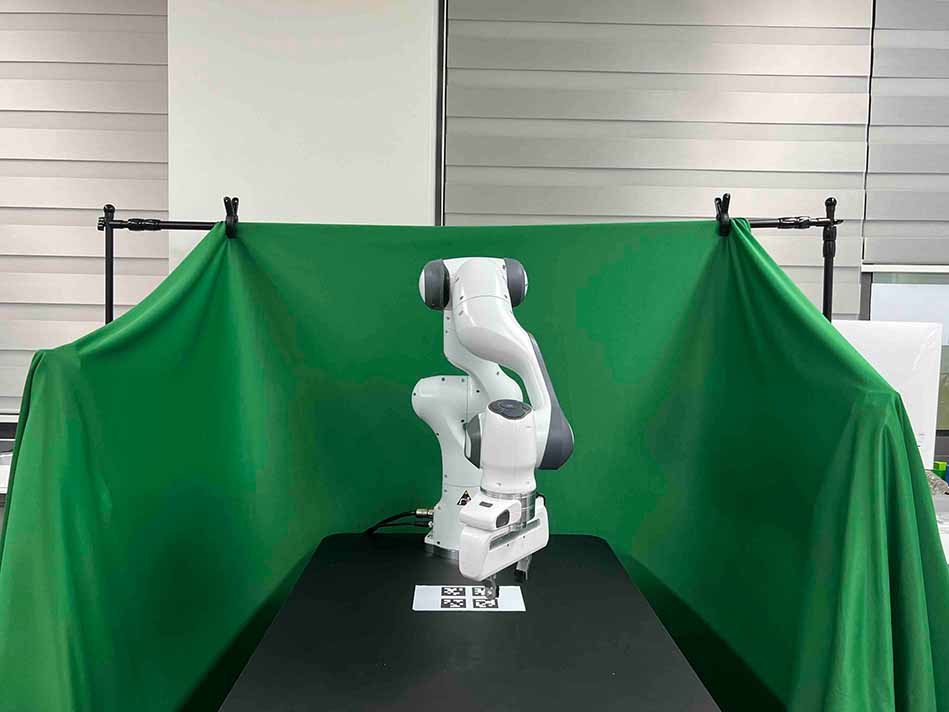}
  \caption{\textbf{Installed Background.}}
  \label{fig:instruction:background}
  \vspace{-3em}
\end{wrapfigure}

\begin{figure}[h]
    \centering
    \begin{subfigure}[t]{0.22\textwidth}
        \includegraphics[width=\textwidth]{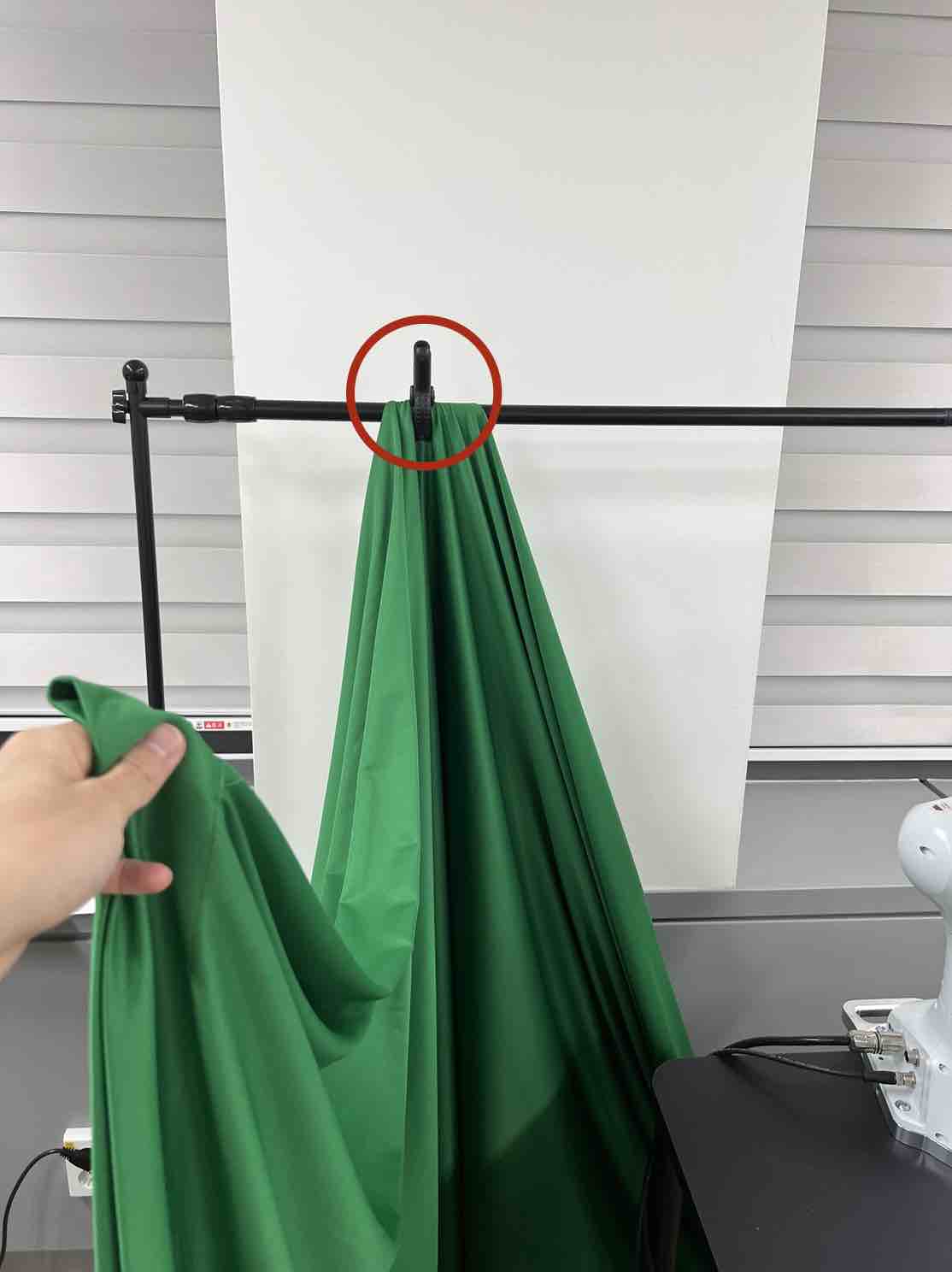}
        \caption{Background left clamp}
        \label{fig:instruction:backgrouond_left_clamp}
    \end{subfigure}
    \begin{subfigure}[t]{0.22\textwidth}
        \includegraphics[width=\textwidth]{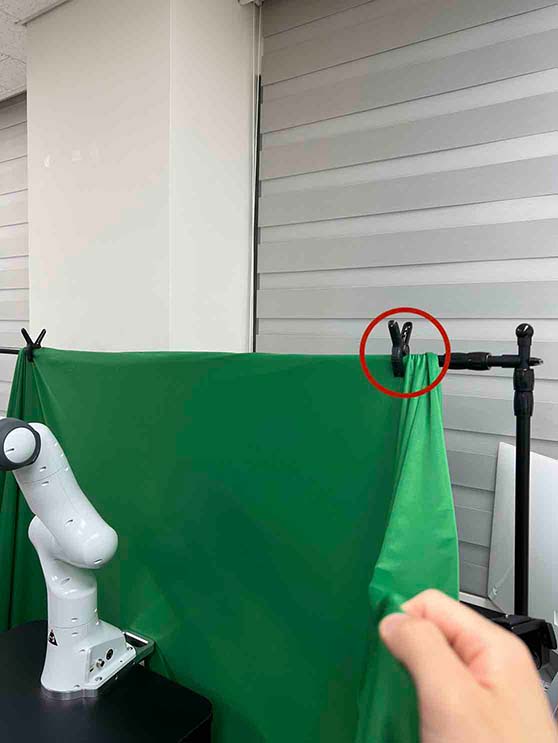}
        \caption{Background right clamp}
        \label{fig:instruction:background_right_clamp}
    \end{subfigure}
    \begin{subfigure}[t]{0.22\textwidth}
        \includegraphics[width=\textwidth]{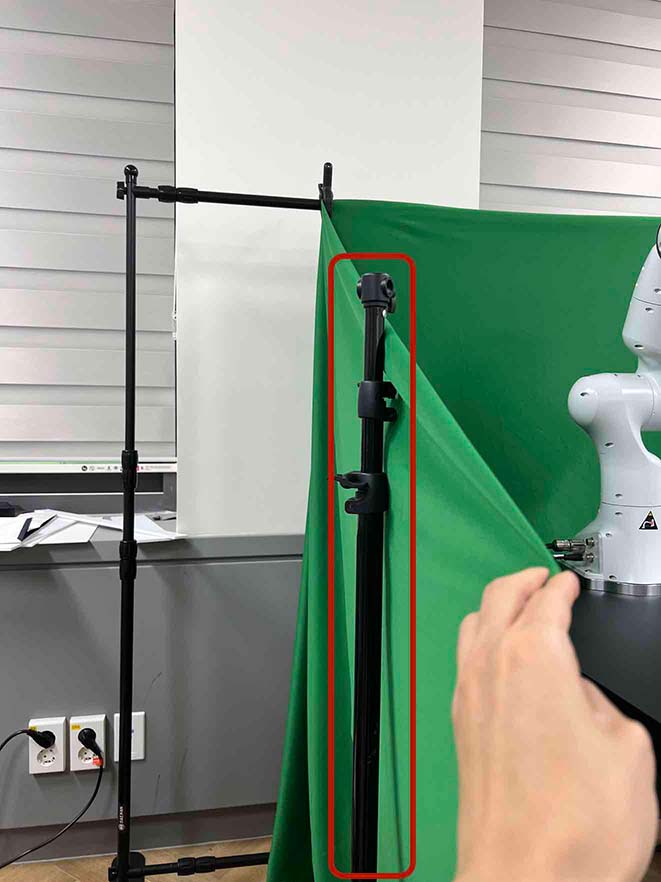}
        \caption{Left pole}
        \label{fig:instruction:background_left_pole}
    \end{subfigure}
    \begin{subfigure}[t]{0.22\textwidth}
        \includegraphics[width=\textwidth]{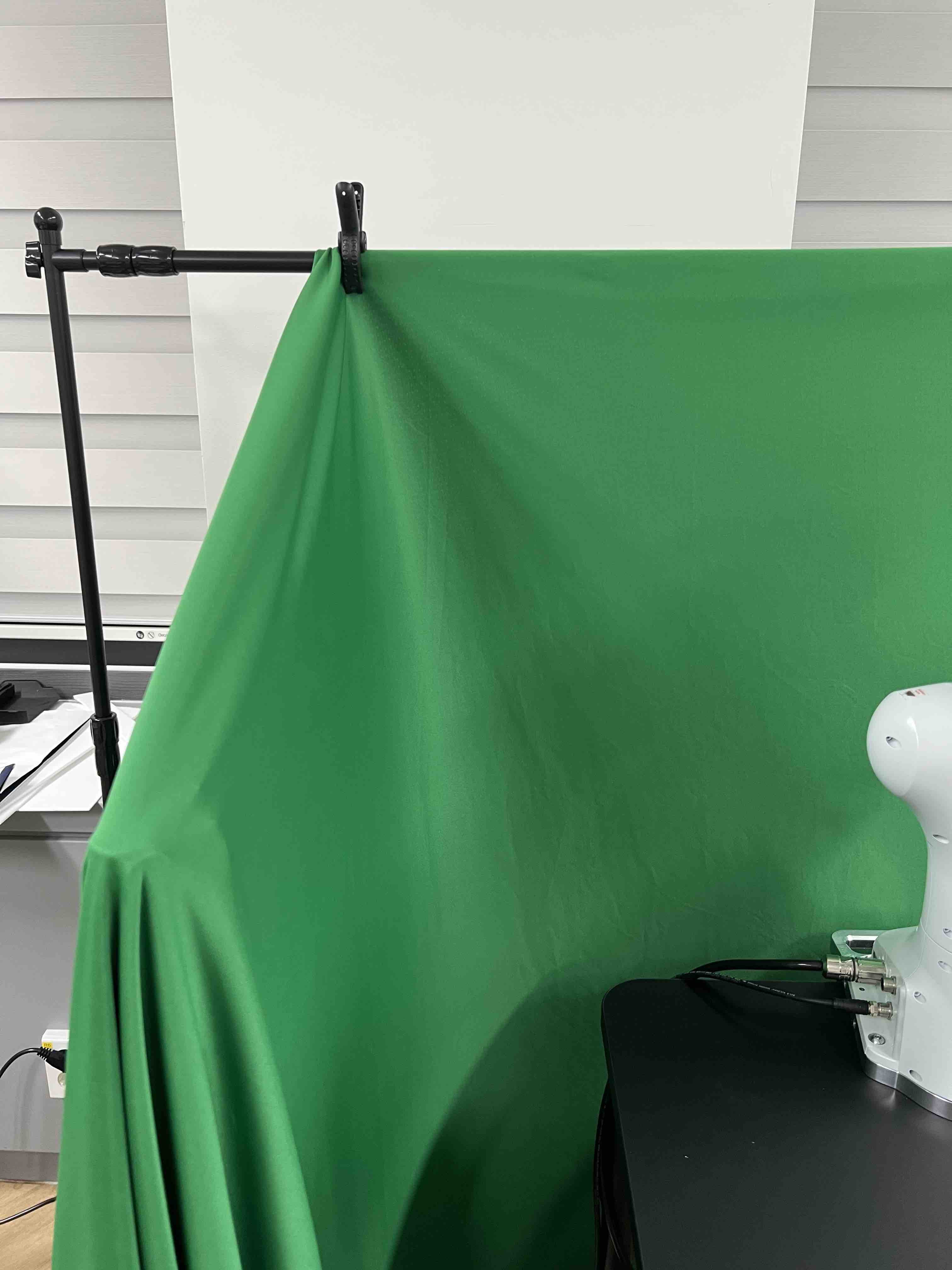}
        \caption{Left pole covered}
        \label{fig:instruction:background_left_pole_covered}
    \end{subfigure}
    \caption{\textbf{Background installation.}}
    \label{fig:instruction:background_instruction}
    \vspace{-1em}
\end{figure}

For consistent background across different lab environments, cover the background with a green backdrop.
\begin{enumerate}
    \item Clamp the left side of the backdrop, as shown in \Cref{fig:instruction:backgrouond_left_clamp}. Make sure to leave some extra cloth to cover the left side.
    \item Similarly, clamp the right side of the backdrop, as shown in \Cref{fig:instruction:background_right_clamp}.
    \item Place a tripod next to the table, and hang the left side of the backdrop to the tripod, as shown in \Cref{fig:instruction:background_left_pole} and \Cref{fig:instruction:background_left_pole_covered}.
    \item Repeat this process for the right side.
    \item The final background should look like \Cref{fig:instruction:background}.
\end{enumerate}

\medskip
\textbf{Important Checklist:}
\begin{itemize}
    \item Make sure there are \textbf{minimum wrinkles and shadows} on the cloth. \CheckBox[name=checkbox3, width=0.7em, height=0.7em]{} 
    \item Green cloth fully covers the narrow side of the table. \CheckBox[name=checkbox4, width=0.7em, height=0.7em]{} 
    \item The cloth covers the left and right edges of the table (at least $1/3$) so that it fully covers the front camera view. \CheckBox[name=checkbox5, width=0.7em, height=0.7em]{} 
\end{itemize}

\clearpage

\subsection{Place Base AprilTag}

\begin{wrapfigure}{r}{0.35\textwidth}
    \centering
    \vspace{-1em}
    \includegraphics[width=\linewidth]{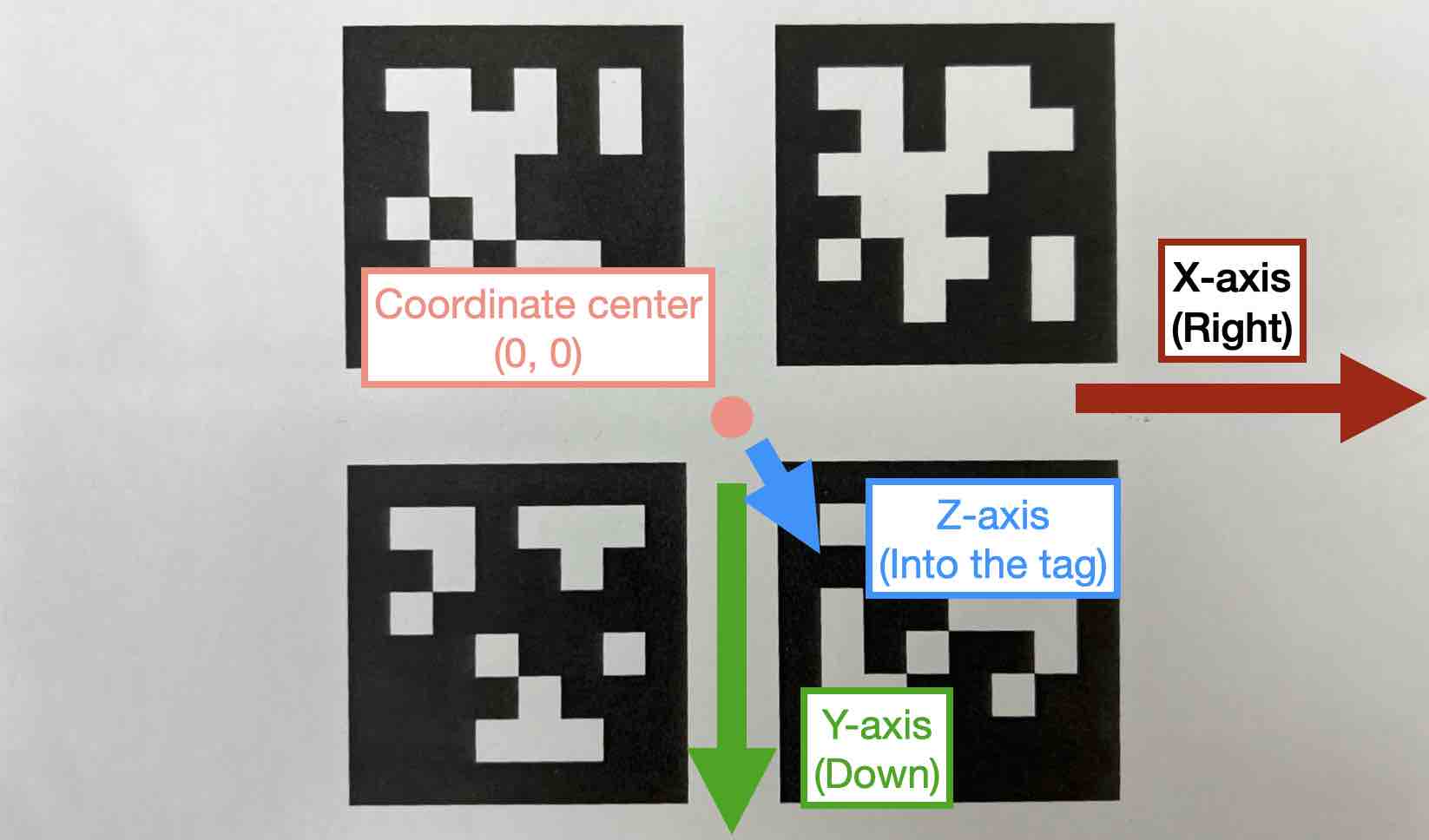}
    \caption{\textbf{Base AprilTag coordinate system.} The arrow indicates a positive direction. The coordinate system follows a right-handed convention for rotation.}
    \label{fig:instruction:base_apriltag_coordinate}
    \vspace{-1em}
\end{wrapfigure}
The base AprilTag defines the world coordinate system; therefore, the camera will be set relative to this base tag. The position and angle of the base tag are \textbf{critical for reproducibility}; thus, the placement of the base tag on the table should be very precise. Be cautious when attaching the AprilTag, as it can easily be attached with tilted angles. Ensure that both the rulers and AprilTag are properly aligned and straight.

\begin{figure}[ht]
    \centering
    \begin{subfigure}[t]{0.24\textwidth}
        \includegraphics[width=\textwidth]{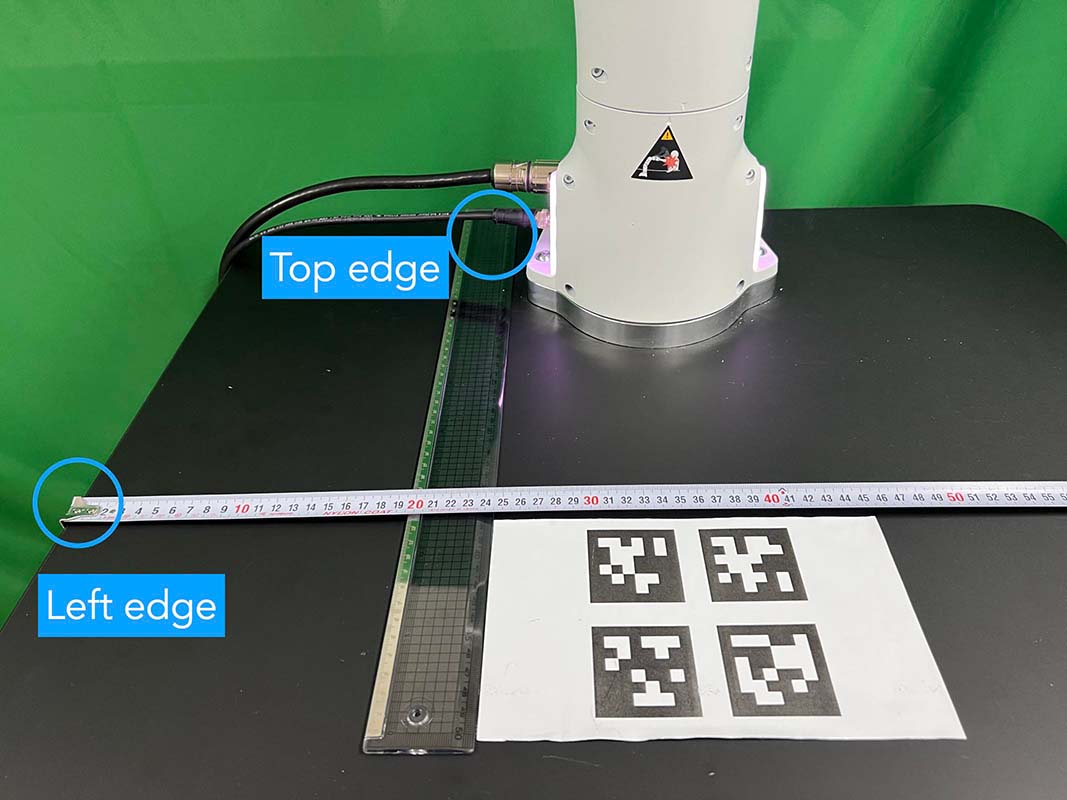}
        \caption{Rulers on table}
        \label{fig:instruction:base_apriltag_ruler}
    \end{subfigure}
    \begin{subfigure}[t]{0.24\textwidth}
        \includegraphics[width=\textwidth]{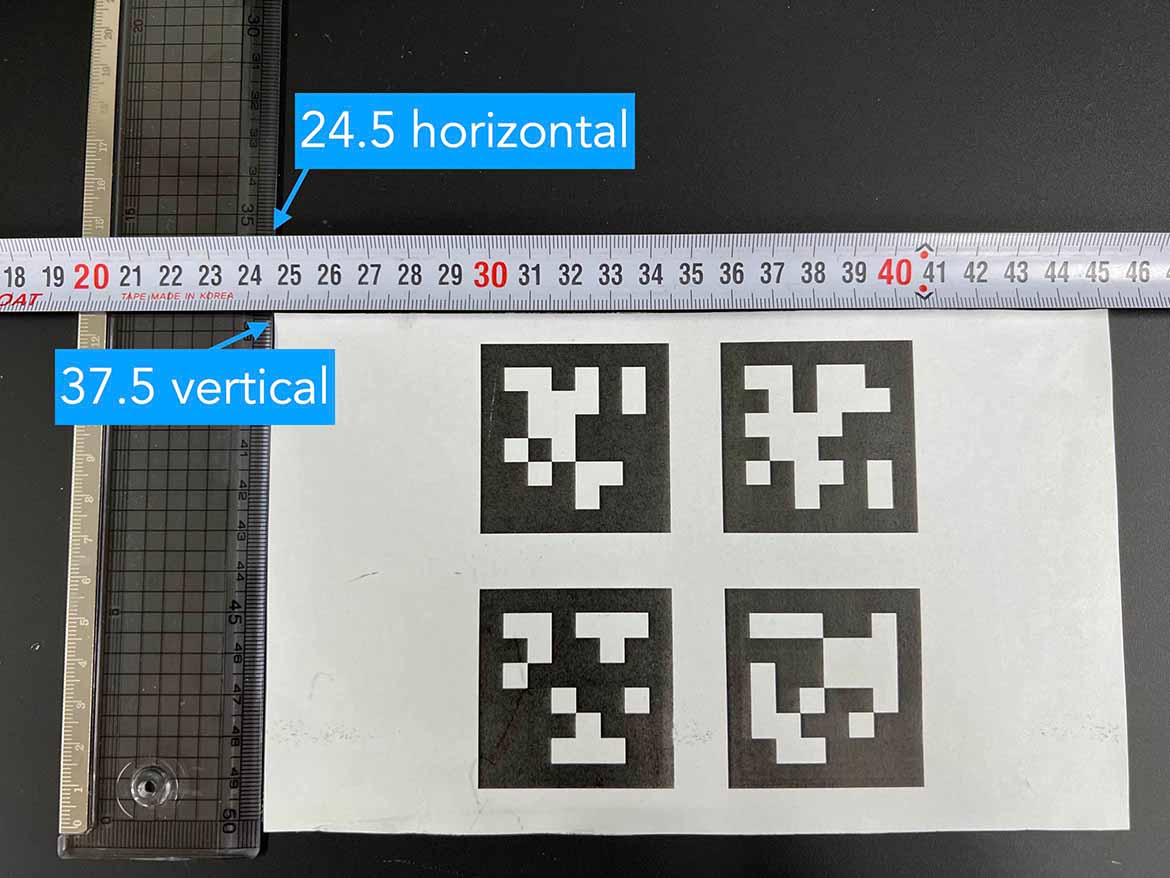}
        \caption{Base tag position on table}
        \label{fig:instruction:base_apriltag_placement}
    \end{subfigure}
    \begin{subfigure}[t]{0.24\textwidth}
        \includegraphics[width=\textwidth]{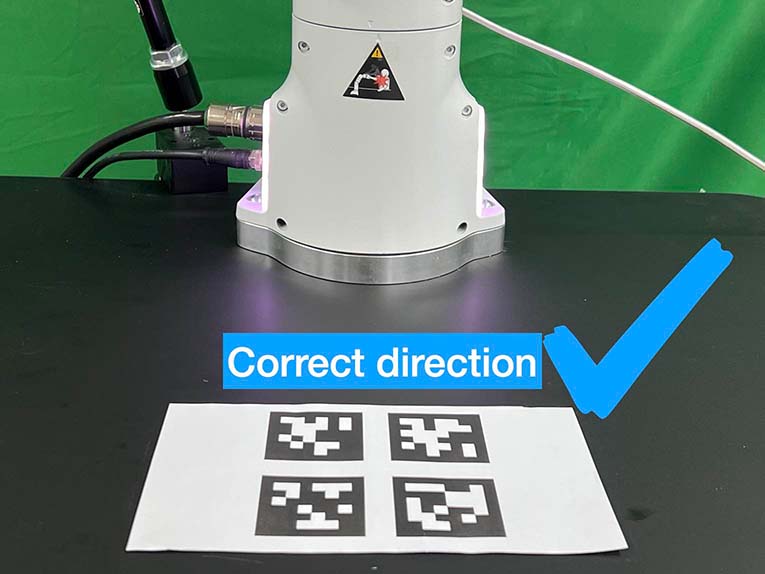}
        \caption{Correct base tag direction}
        \label{fig:instruction:base_apriltag_example}
    \end{subfigure}
    \begin{subfigure}[t]{0.24\textwidth}
        \includegraphics[width=\textwidth]{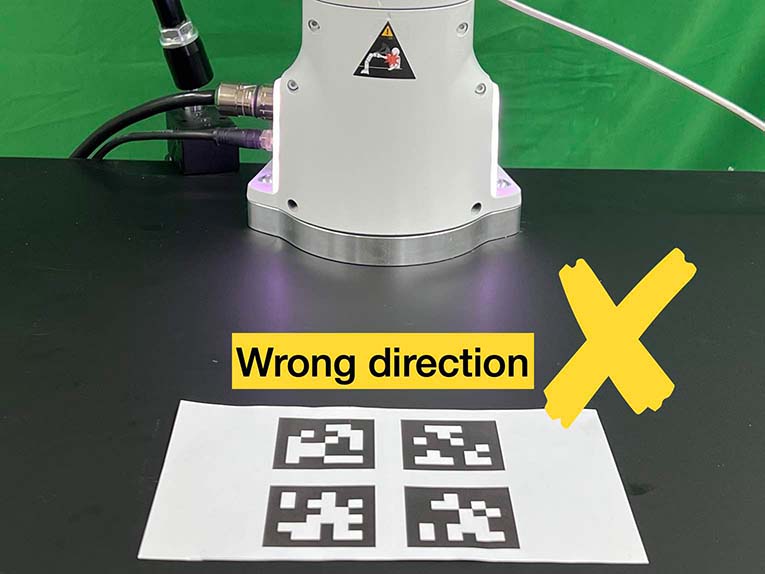}
        \caption{Wrong base tag direction}
    \end{subfigure}  
    \caption{\textbf{Base AprilTag.}}
    \label{fig:instruction:base_apriltag}
    \vspace{-1em}
\end{figure}

\begin{enumerate}
    \item Align tape ruler so that \SI{0}{\centi\meter} is at the left of the table and plastic ruler so that \SI{0}{\centi\meter} is at the top edge of the table, as illustrated in \Cref{fig:instruction:base_apriltag_ruler}. Make sure the two rulers are perpendicular.
    \item Place the center of the base tag at \SI{24.5}{\cm} horizontally and \SI{37.5}{\cm} vertically, as shown in \Cref{fig:instruction:base_apriltag_placement}, in a correct orientation of the patterns (\Cref{fig:instruction:base_apriltag_example}).
    \item Use double-sided tape to affix the base tag. Note that wrinkled paper can cause unreliable detection. Ensure the paper remains flat by attaching it with double-sided tape in all four corners.
\end{enumerate}

\medskip
\textbf{Important Checklist:}
\begin{itemize}
    \item Double-check the base AprilTag in the exact position (like, less than \SI{2}{\mm} error). \CheckBox[name=checkbox6, width=0.7em, height=0.7em]{}
    \item The base AprilTag is firmly attached flat without wrinkles. \CheckBox[name=AprilTagCheckbox2, width=0.7em, height=0.7em]{}
    \item Check the pattern of the base tag to ensure its correct direction. \CheckBox[name=checkbox8, width=0.7em, height=0.7em]{}
\end{itemize}

\clearpage

\subsection{Install Front and Rear Cameras}

\begin{wrapfigure}{r}{0.3\textwidth}
    \centering
    \vspace{-3em}
    \includegraphics[width=\linewidth]{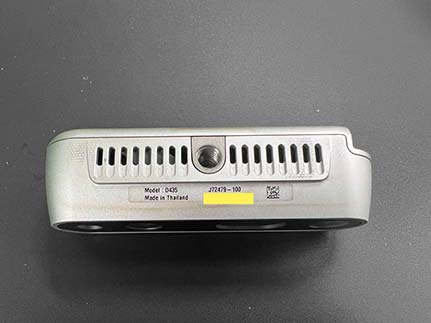}
    \caption{\textbf{Camera serial number} written on the bottom of the camera.}
    \label{fig:instruction:camera_serial}
    \vspace{-2em}
\end{wrapfigure}

Our system requires three cameras: front, rear, and wrist cameras. Prior to installation, determine the specific camera to be used for each view, and write down the serial numbers of the cameras for wrist, front, and rear cameras, as they will be required for subsequent steps. The serial number is written on the bottom of the camera, as shown in \Cref{fig:instruction:camera_serial}.

First, install the front and rear cameras. You can utilize any camera mount product for the front camera mount if they can match the same camera view with our original setup. However, we highly recommend users use a camera mount from either Ulanzi or Manfrotto, both of which we have confirmed to be reliable. We explain how to install the front camera using these two camera mounts, although you can further adjust it during fine-grained calibration in \Cref{sec:instruction:fine-tune}.

\begin{figure}[h]
    \begin{subfigure}{.7\textwidth}
        \centering
        \begin{subfigure}{0.30\textwidth}
            \includegraphics[width=\textwidth]{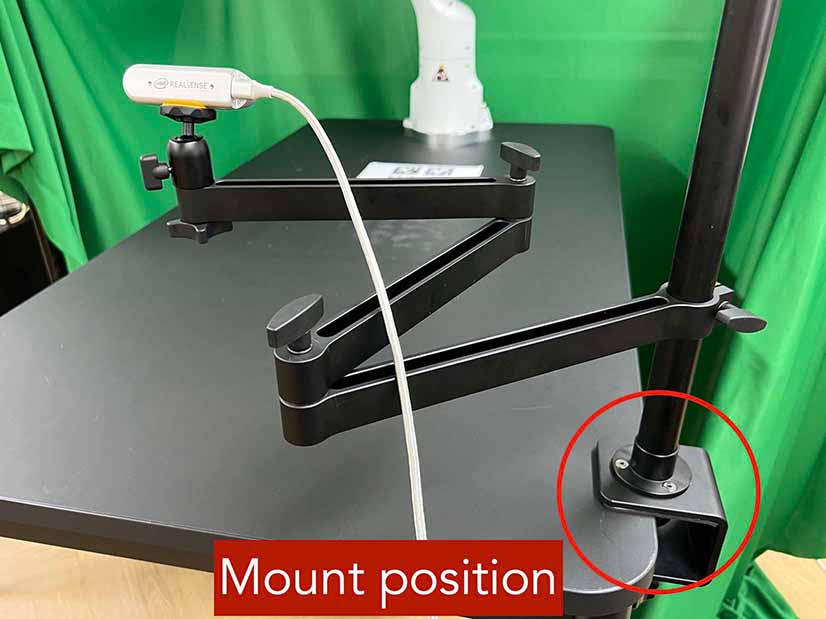}
            \caption{Ulanzi position}
            \label{fig:instruction:front_camera_position}
        \end{subfigure}
        \begin{subfigure}{0.30\textwidth}
            \includegraphics[width=\textwidth]{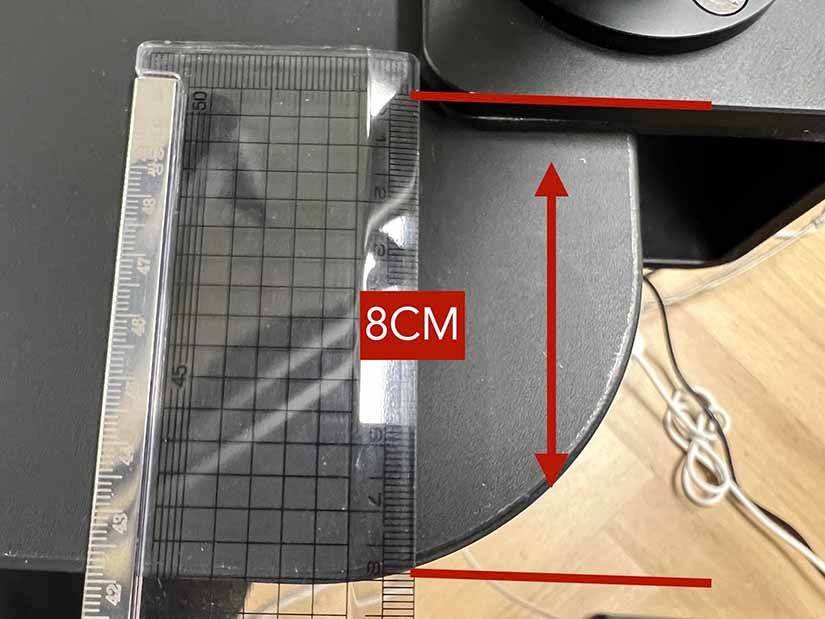}
            \caption{Ulanzi distance}
            \label{fig:instruction:front_camera_distance}
        \end{subfigure}
        \begin{subfigure}{0.30\textwidth}
            \includegraphics[width=\textwidth]{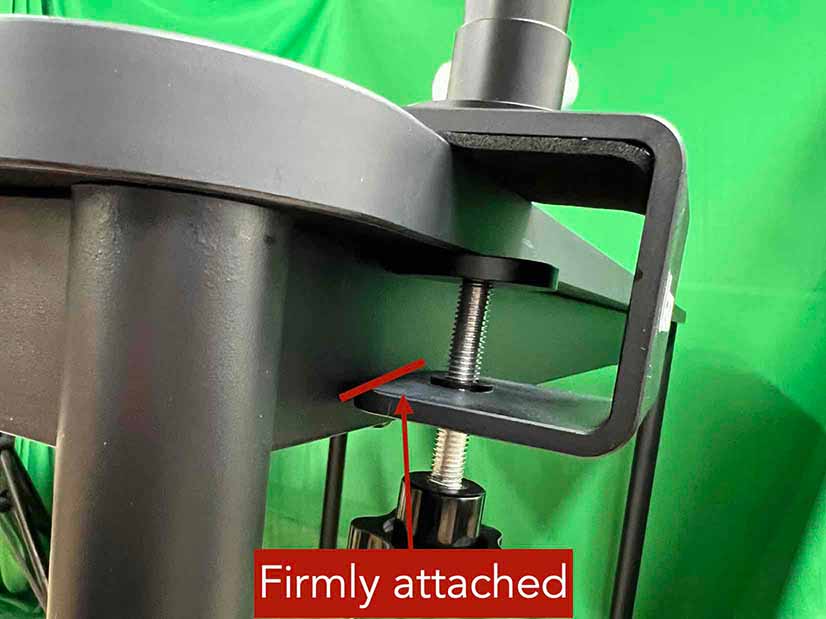}
            \caption{Ulanzi attacment}
            \label{fig:instruction:front_camera_firmly_attached}
        \end{subfigure}

        \vspace{1em}
        \begin{subfigure}{0.30\textwidth}
            \includegraphics[width=\textwidth]{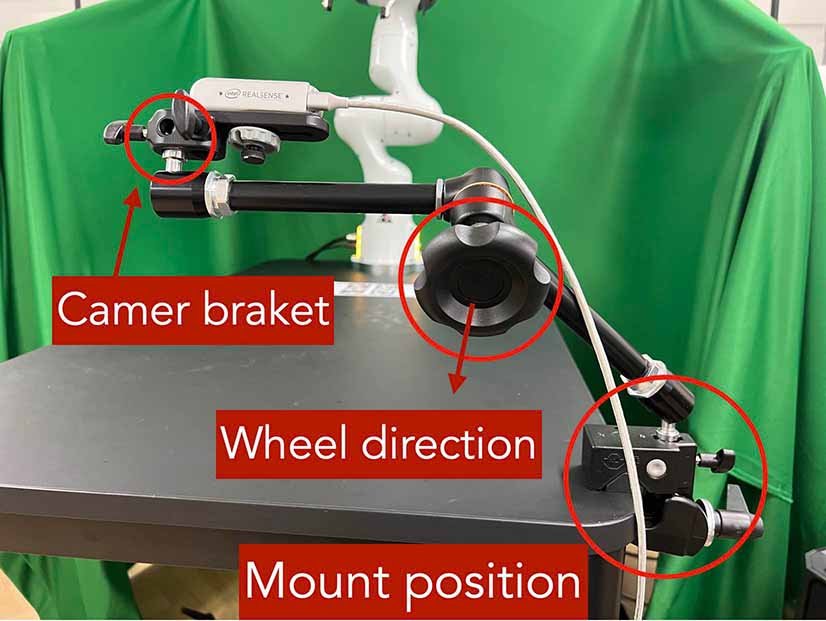}
            \caption{Manfrotto position}
            \label{fig:instruction:manfrotto_front_camera_position}
        \end{subfigure}
        \begin{subfigure}{0.30\textwidth}
            \includegraphics[width=\textwidth]{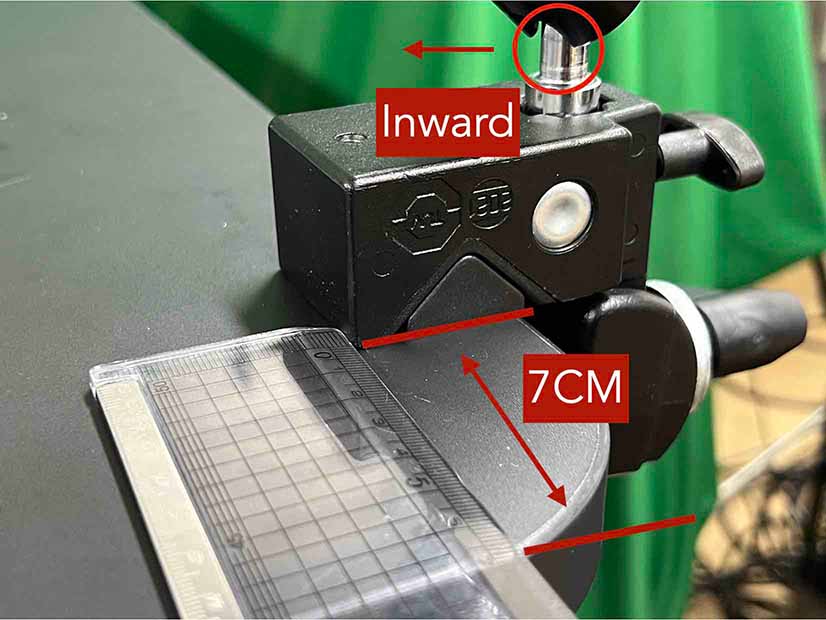}
            \caption{Manfrotto distance}
            \label{fig:instruction:manfrotto_front_camera_distance}
        \end{subfigure}
        \begin{subfigure}{0.30\textwidth}
            \includegraphics[width=\textwidth]{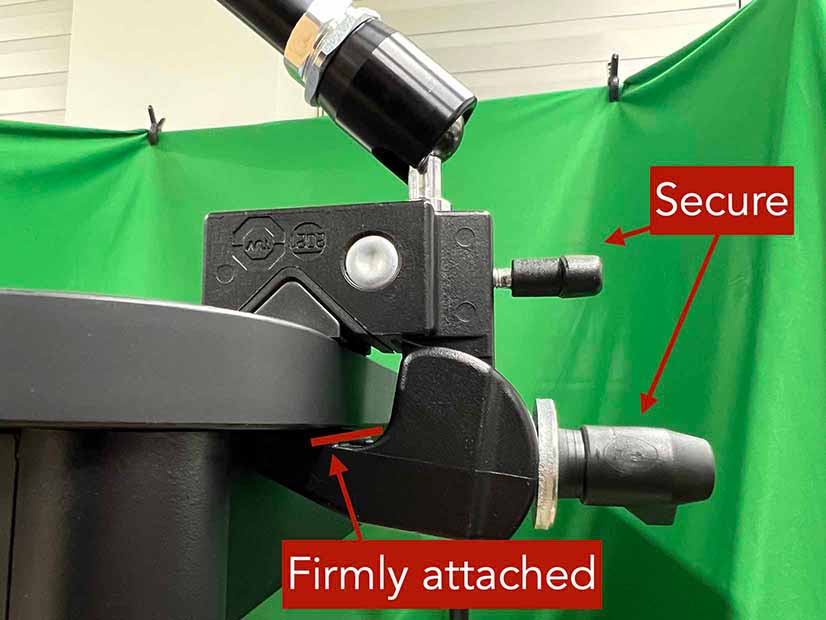}
            \caption{Manfrotto attachment}
            \label{fig:instruction:manfrotto_front_camera_firmly_attached}
        \end{subfigure}
    \end{subfigure}\hfill
    \begin{subfigure}{.3\textwidth}
        \centering
        \includegraphics[width=.95\textwidth]{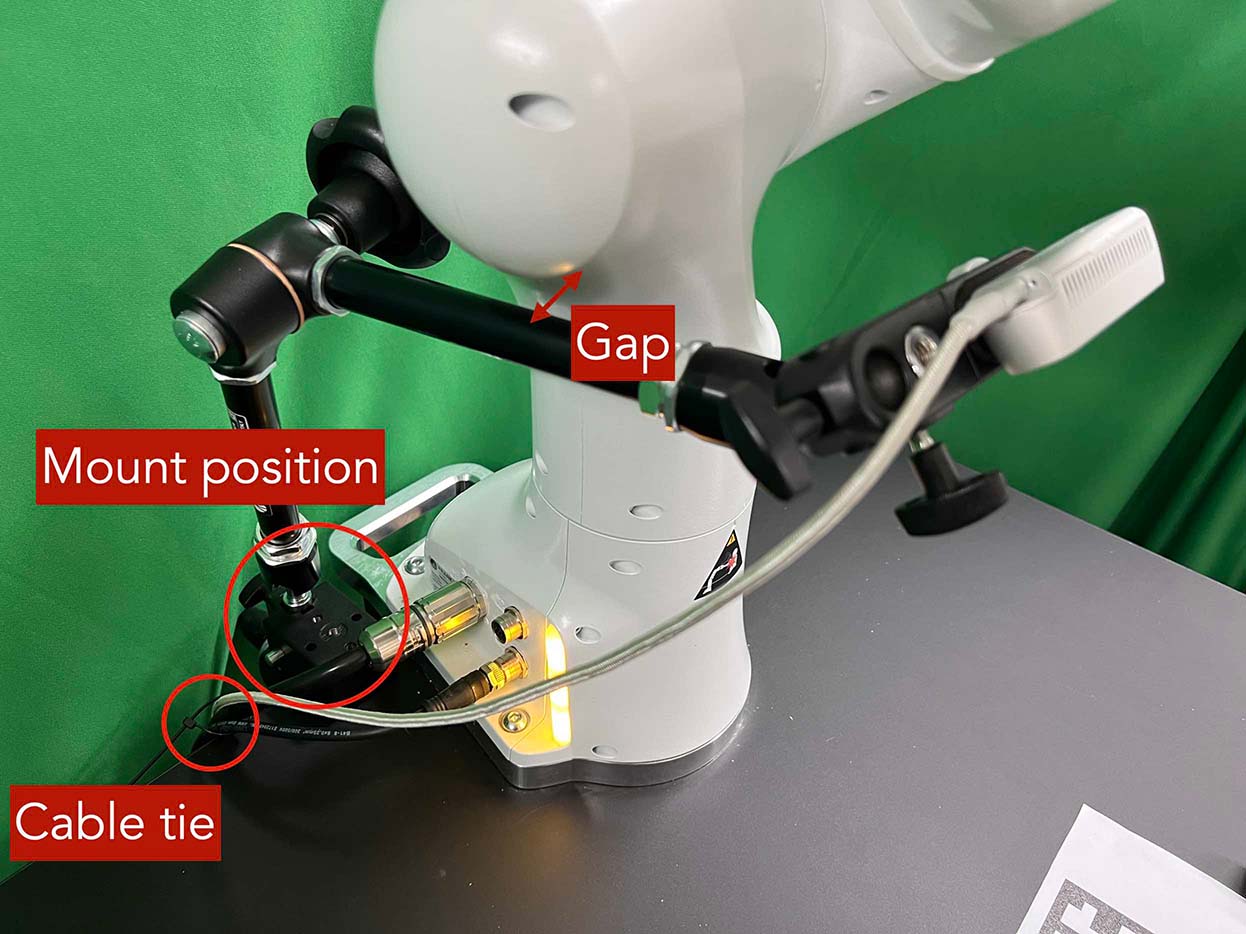}
        \caption{Rear camera installation}
        \label{fig:instruction:rear_camera_installation}
        \vspace{5em}
    \end{subfigure}

    \caption{\textbf{Front and rear camera installation} with Ulanzi and Manfrotto camera mounts.}
    \label{fig:instruction:camera_installations}
\end{figure}

\begin{itemize}
    \item \textbf{Option 1, Ulanzi:} Clamp the front camera mount to the right side of the table, as shown in \ref{fig:instruction:front_camera_position}. Position the camera mount \SI{8}{\cm} away from the table edge, as shown in \ref{fig:instruction:front_camera_distance}. While measuring the distance, ensure the camera mount's base is firmly attached, as illustrated in \ref{fig:instruction:front_camera_firmly_attached}.
    \item \textbf{Option 2, Manfrotto:} Clamp the front camera mount to the right side of the table. The camera bracket needs to be affixed using the left hole and the locking wheel should be oriented outward, as shown in \Cref{fig:instruction:manfrotto_front_camera_position}.  Position the camera mount \SI{7}{\cm} away from the table edge, as shown in \Cref{fig:instruction:manfrotto_front_camera_distance}. Arrange the deeper section to face the inside to provide better flexibility in camera movement. During the distance measurement, make sure that the camera mount's base is firmly attached and valves are securely fastened, as shown in \Cref{fig:instruction:manfrotto_front_camera_firmly_attached}.
    \item Place the camera approximately in the center (horizontally) of the table and orient it to face the base AprilTag. You will fine-tune its pose in a later section.
    \item Connect the front camera to \textbf{client} computer using a USB \textbf{3.1} cable.
    \item Clamp the rear camera mount next to the robot base, as shown in \Cref{fig:instruction:rear_camera_installation}. Plug USB \textbf{3.1} cable. Utilize a cable tie to fasten the pair of cables from the robot and the single cable from the camera. Ensure a sufficient gap between the camera mount and the robot to avoid any collision.
\end{itemize}

\clearpage

\subsection{Install Wrist Camera}

\begin{figure}[ht]
    \centering
    \begin{subfigure}[t]{0.25\textwidth}
        \includegraphics[width=\textwidth]{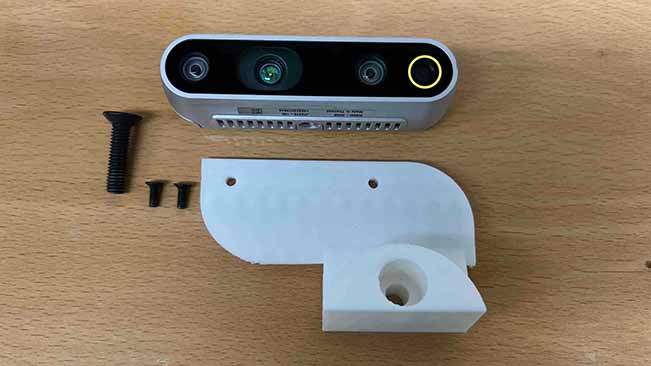}
        \caption{Camera, mount, and screws}
        \label{fig:instruction:wrist_camera_installation_ingredients}
    \end{subfigure}
    \begin{subfigure}[t]{0.25\textwidth}
        \includegraphics[width=\textwidth]{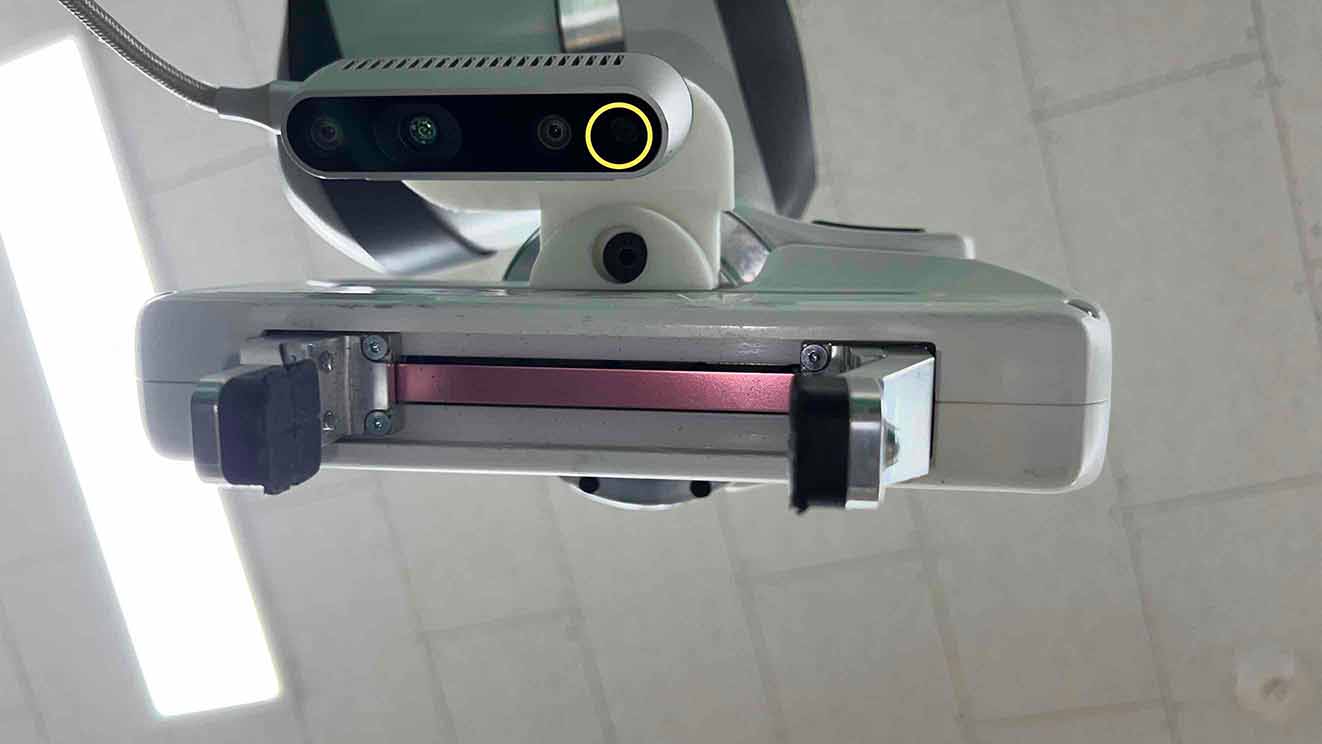}
        \caption{Wrist camera direction}
        \label{fig:instruction:wrist_camera_installation_view1}
    \end{subfigure}
   \begin{subfigure}[t]{0.25\textwidth}
        \includegraphics[width=\textwidth]{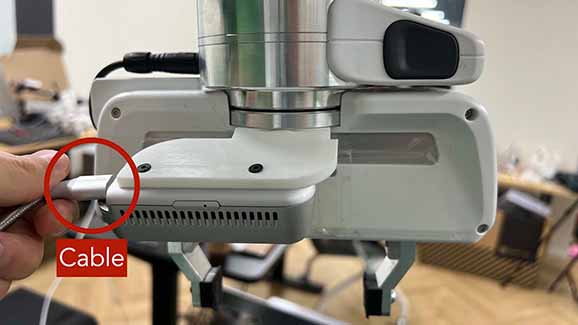}
        \caption{Wrist camera position}
        \label{fig:instruction:wrist_camera_installation_view2}
    \end{subfigure}
    \caption{\textbf{Wrist camera installation.} The yellow circles represent the RGB camera.}
    \label{fig:instruction:wrist_camera_installation}
    \vspace{-1em}
\end{figure}

\begin{figure}[h]
    \centering
    \includegraphics[width=.4\textwidth]{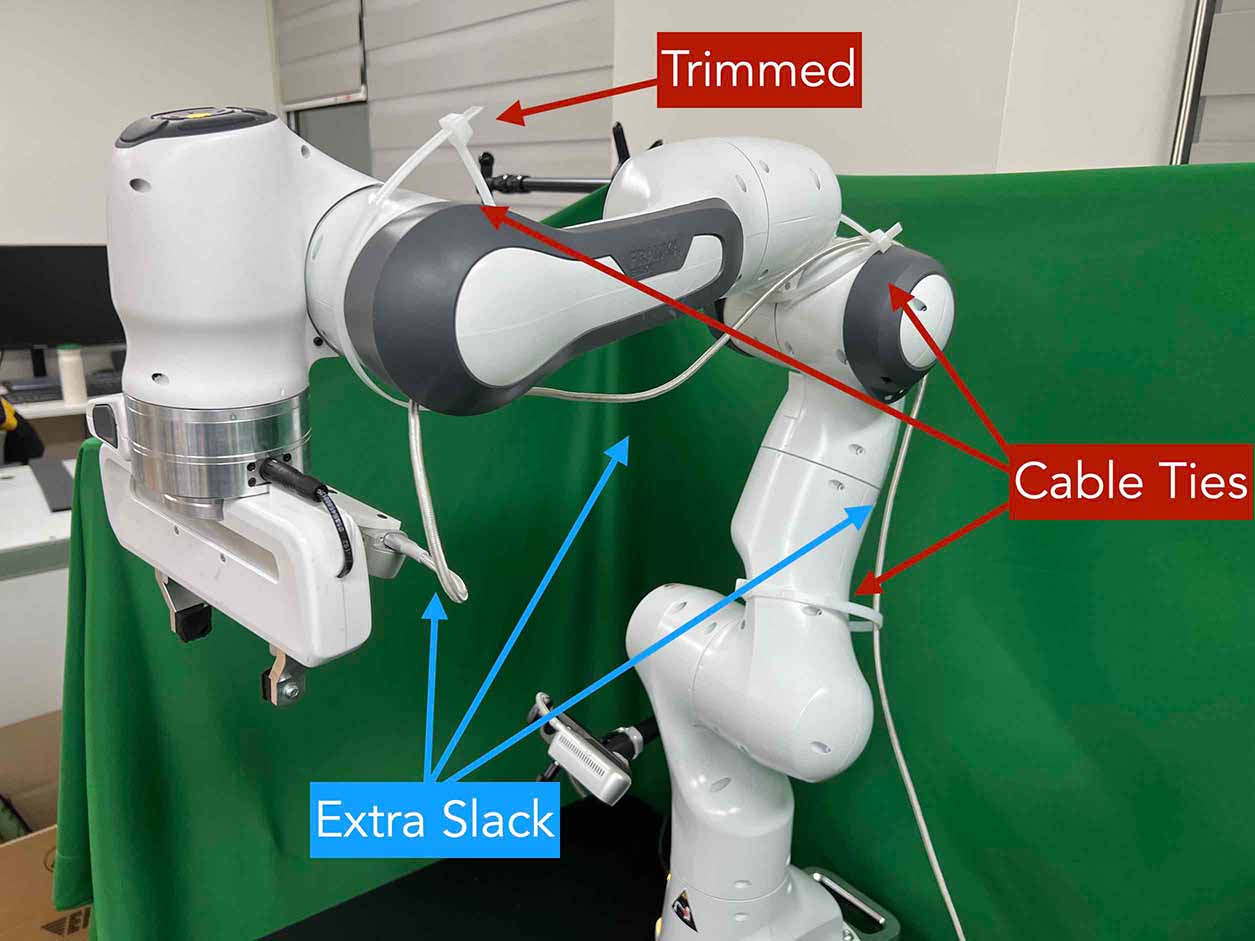}
    \caption{\textbf{Cable organization.}}
    \label{fig:instruction:wrist_camera_cable}
\end{figure}

Now, we install the wrist camera as follows:
\begin{enumerate}
    \item Install the wrist camera on the robot wrist using the 3D printed camera mount. Take note of the direction in which the RGB camera should face: it should be aimed toward the gripper's tip, as shown in \Cref{fig:instruction:wrist_camera_installation_view1}. The camera should be positioned on the rear side of the end-effector. Take a look at \Cref{fig:instruction:wrist_camera_installation_view2} and \Cref{fig:instruction:wrist_camera_cable} to gain  a clear understanding of its placement.
    \item Connect the wrist camera to \textbf{client} computer using a USB \textbf{3.1} cable.
    \item Fasten the cable to the robot arm with three cable ties, as shown in \Cref{fig:instruction:wrist_camera_cable}. Ensure to provide additional slack in the cable, allowing the robot to move without any tension from the cable. Trim the surplus length from the cable ties to ensure no extra material remains.
\end{enumerate}

\medskip
\textbf{Important Checklist:}
\begin{Form}
\begin{itemize}
        \item Ensure the direction of the wrist camera is correctly set; the camera is positioned on the end-effector's back side, and the cable is plugged to the left when viewed from the rear. \CheckBox[name=checkbox10, width=0.7em, height=0.7em]{} 
        \item Firmly attach the camera and camera mount to the robot by tightening the screws. \CheckBox[name=checkbox9, width=0.7em, height=0.7em]{} 
        \item Three cable ties are fastened as shown in \Cref{fig:instruction:wrist_camera_cable}. \CheckBox[name=checkbox11, width=0.7em, height=0.7em]{} 
        \item The cable has additional slack. \CheckBox[name=checkbox12, width=0.7em, height=0.7em]{} 
        \item The surplus length from the cable ties is trimmed. \CheckBox[name=checkbox13, width=0.7em, height=0.7em]{} 
\end{itemize}
\end{Form}

\clearpage

\subsection{Install Software}

\textbf{Skip this section if you already installed the software.}

Before moving on to the following installation steps, install our software stacks on server and client computers. More detailed instructions can be found in \href{https://clvrai.github.io/furniture-bench/docs}{our online documentation}.

\medskip
\textbf{Client computer:}
\begin{enumerate}
    \item Install Franka Control Interface (FCI) version 4.2.2. 
    \item Install \href{https://docs.nvidia.com/datacenter/cloud-native/container-toolkit/install-guide.html#installing-on-ubuntu-and-debian}{nvidia-docker2}.
    \item Clone our code and pull our docker image:
\begin{lstlisting}[language=bash]
 git clone https://github.com/clvrai/furniture-bench.git
 docker pull furniturebench/client-gpu:latest \end{lstlisting}
    \item Set environment variables:
\begin{lstlisting}[language=bash]
 # Set path to the furniture-bench repo, e.g., /home/<username>/furniture-bench
 export FURNITURE_BENCH=</path/to/furniture-bench> \end{lstlisting}
   \item Run client docker container:
\begin{lstlisting}[language=bash]
 ./launch_client.sh --gpu --pulled \end{lstlisting}
\end{enumerate}

\medskip
\textbf{Server computer:}
\begin{enumerate}
    \item Install \href{https://frankaemika.github.io/docs/installation_linux.html#setting-up-the-real-time-kernel}{real-time kernel}.
    \item Clone our code and pull our docker image:
\begin{lstlisting}[language=bash]
 git clone https://github.com/clvrai/furniture-bench.git
 docker pull furniturebench/server:latest \end{lstlisting}
    \item Set environment variables:
\begin{lstlisting}[language=bash]
 # Set path to the furniture-bench repo, e.g., /home/<username>/furniture-bench
 export FURNITURE_BENCH=</path/to/furniture-bench> \end{lstlisting}
\end{enumerate}


\subsection{Set Up Connection}

\begin{wrapfigure}{r}{0.4\textwidth}
    \centering
    \vspace{-3em}
    \includegraphics[width=\linewidth]{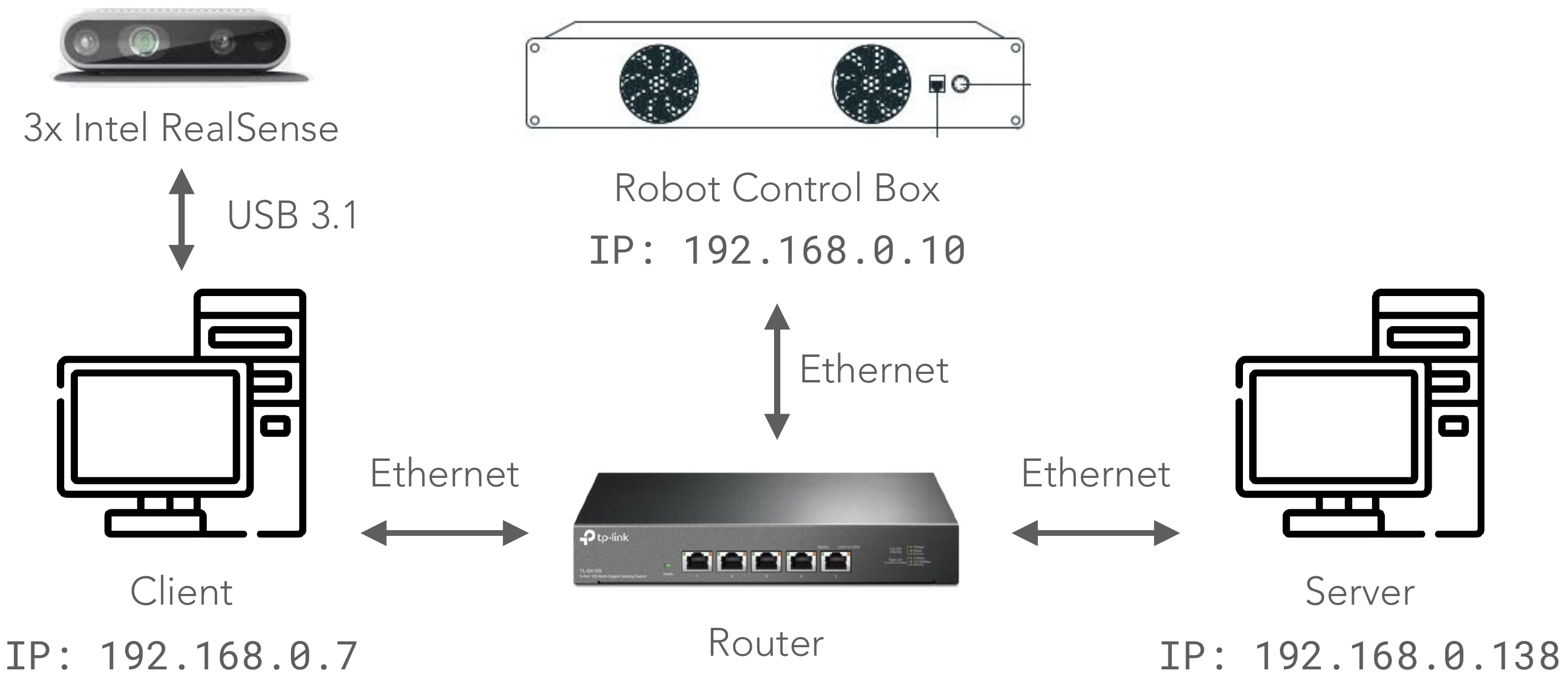}
    \caption{\textbf{Example network setup.} The IP addresses need to be adjusted according to your local network.}
    \label{fig:instruction:example_network_setup}
    \vspace{-5em}
\end{wrapfigure}

Server, client, and robot communicate through a local Ethernet network, as shown in \Cref{fig:instruction:example_network_setup}. To establish connections to server and cameras, the following environment variables need to be set in client container:

\begin{lstlisting}[language=bash]
 export SERVER_IP=<IP of the server> # e.g. 192.168.0.138
 export CAM_WRIST_SERIAL=<serial number of the wrist camera>
 export CAM_FRONT_SERIAL=<serial number of the front camera> 
 export CAM_REAR_SERIAL=<serial number of the rear camera>
\end{lstlisting}

Ensure that all the cameras are correctly installed and appropriately connected. Execute the following command and confirm the items in the checklist.

\begin{lstlisting}[language=bash]
 python furniture_bench/scripts/run_cam_april.py
\end{lstlisting}

\begin{figure}[ht]
    \centering
    \includegraphics[width=0.8\textwidth]{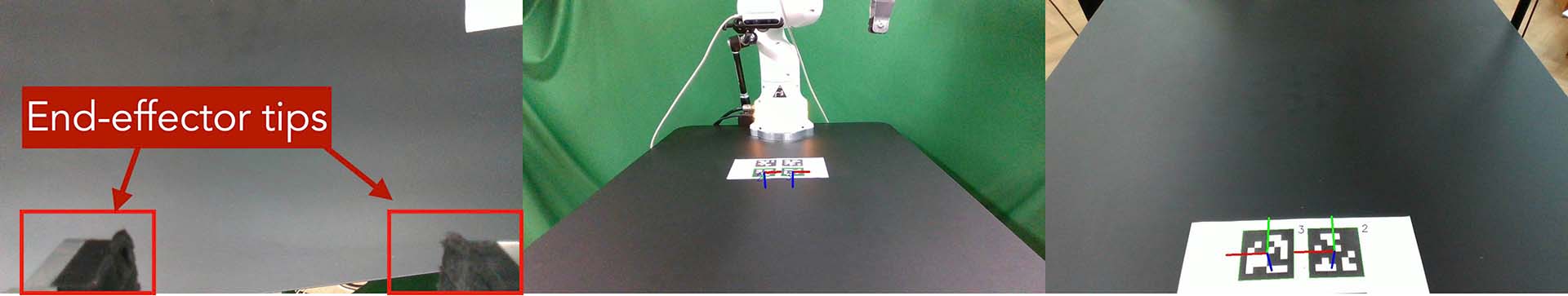}
    \caption{\textbf{Camera observations.}}
    \label{fig:instruction:camera_observations}
    \vspace{-1em}
\end{figure}

\medskip
\textbf{Important Checklist:}
\begin{itemize}
    \item Ensure that the camera displays the wrist, front, and rear views in left-to-right order, as shown in \Cref{fig:instruction:camera_observations}. \CheckBox[name=checkbox14, width=0.7em, height=0.7em]{} 
    \item The wrist camera view must observe both gripper tips as shown in the left figure of \Cref{fig:instruction:camera_observations}. \CheckBox[name=checkbox15, width=0.7em, height=0.7em]{} 
    \item The rear camera should be able to detect the two markers present on the base tag, as shown in the right image of \Cref{fig:instruction:camera_observations}. \CheckBox[name=checkbox16, width=0.7em, height=0.7em]{} 
\end{itemize}

\clearpage

\subsection{Launch and reset the robot}
\begin{figure}[ht]
    \centering
    \begin{subfigure}[t]{0.25\textwidth}
        \includegraphics[width=\textwidth]{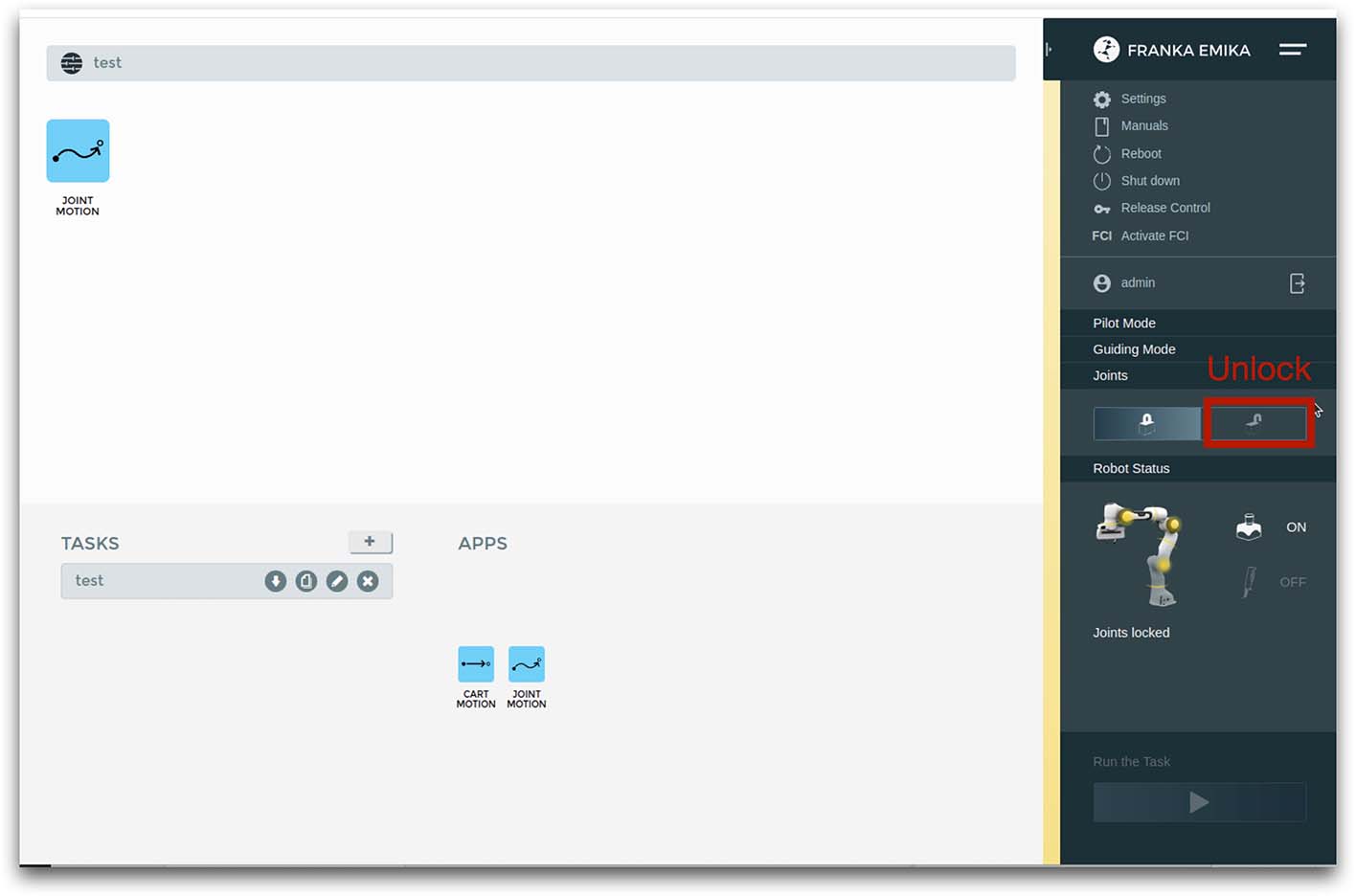}
        \caption{Unlock}
        \label{fig:instruction:unlock_robot}
    \end{subfigure}
    \begin{subfigure}[t]{0.25\textwidth}
        \includegraphics[width=\textwidth]{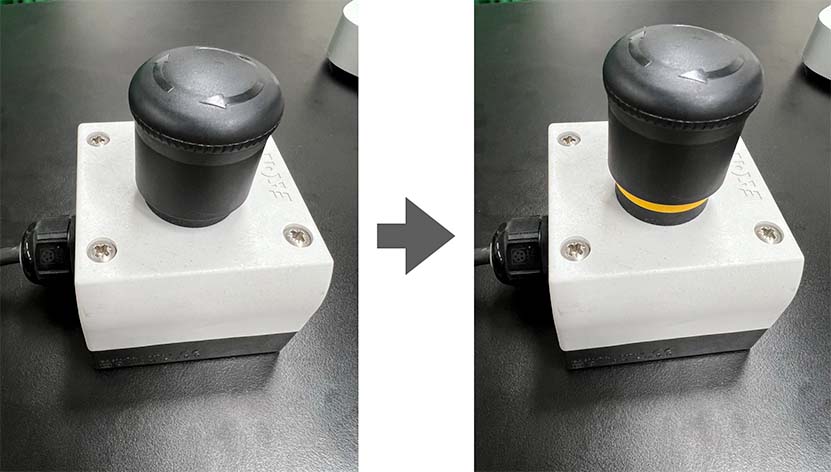}
        \caption{Release activation}
        \label{fig:instruction:release_activation}
    \end{subfigure}
    \begin{subfigure}[t]{0.25\textwidth}
        \includegraphics[width=\textwidth]{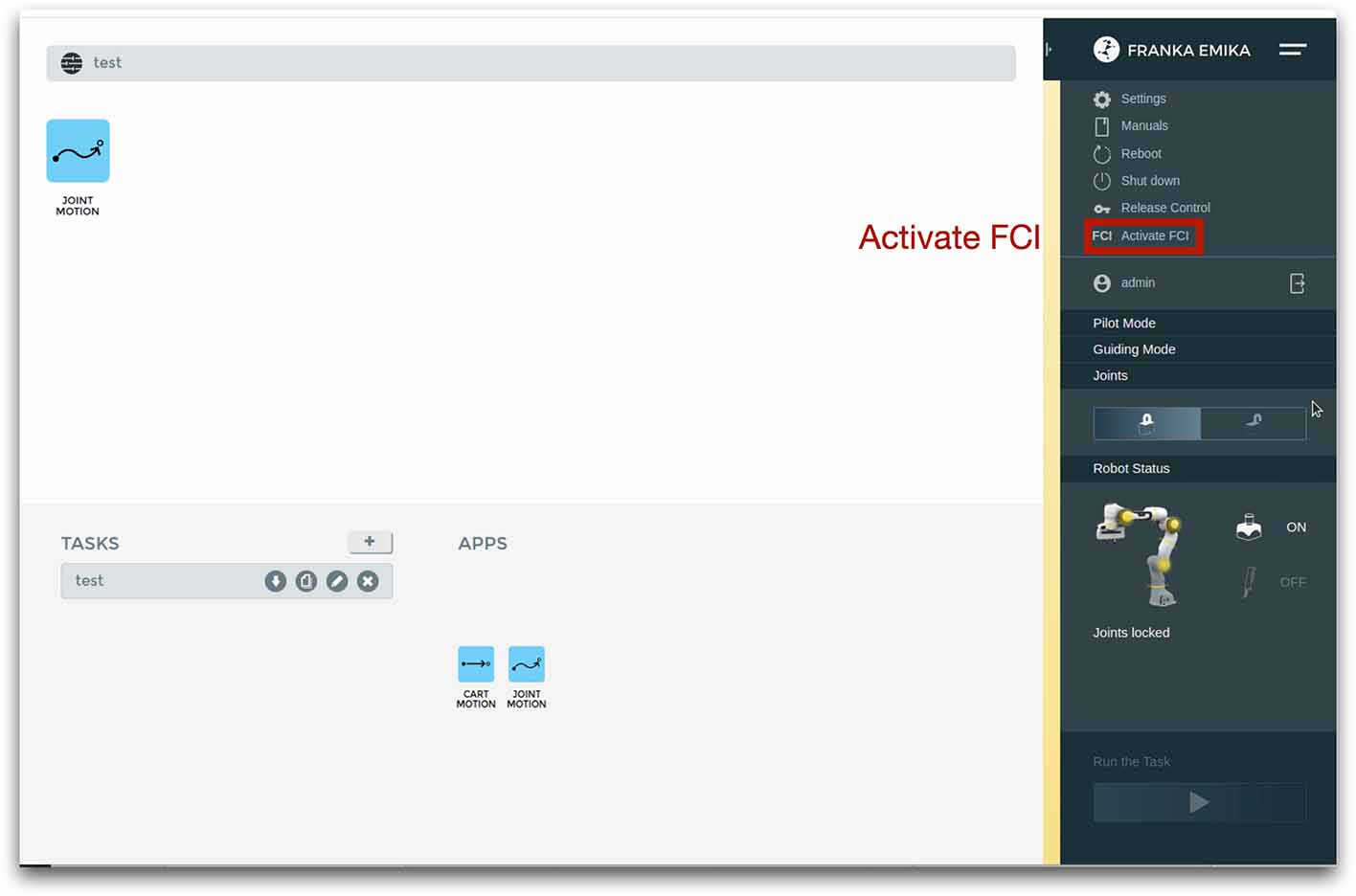}
        \caption{Activate FCI}
        \label{fig:instruction:activate_FCI}
    \end{subfigure}
    \caption{\textbf{Activate Franka Panda controller.}}
    \label{fig:instruction:running_robot}
    \vspace{-1em}
\end{figure}

In this section, we provide a guideline for operating and moving the robot to the predetermined position. To operate the robot, unlock the robot, activate FCI (Franka Control Interface), and launch a server-side daemon as explained below:

\begin{enumerate}
    \item Access the control interface website. You should be able to see a screen like in \Cref{fig:instruction:running_robot}.
    \item Unlock the robot in the Franka Emika web interface, as shown in \Cref{fig:instruction:unlock_robot}
    \item Release the activation button (\Cref{fig:instruction:release_activation}). The light on the robot should turn blue after releasing the button.
    \item Activate FCI in the web interface, as shown in \Cref{fig:instruction:activate_FCI}.
\end{enumerate}

Then, launch a server-side daemon.
\begin{enumerate}
    \item In the server computer, launch the server Docker image with: 
\begin{lstlisting}[language=bash]
 /launch_server.sh --pulled \end{lstlisting}
    \item On server docker container, set the environment variable for server IP by \texttt{export ROBOT\_IP=<robot\_ip>}. For example, \texttt{export ROBOT\_IP=192.168.0.10}
    \item In a \textit{``server''} terminal, launch server daemon with executing:
\begin{lstlisting}[language=bash]
 /furniture-bench/launch_daemon.sh\end{lstlisting}
    If everything is installed and configured correctly, it will alternate between opening and closing the gripper. This is the only program needed to be run on the server. Keep it running while executing commands in the \textit{``client-gpu''} terminal.
\end{enumerate}

Finally, execute the following in a \textit{``client-gpu''} terminal and see the robot moves to the reset pose.
\begin{lstlisting}[language=bash]
 python furniture_bench/scripts/reset.py \end{lstlisting}

\clearpage

\subsection{Fine-tune Front Camera Pose}
\label{sec:instruction:fine-tune}

\begin{figure}[ht]
    \centering
    \begin{subfigure}[t]{0.3\textwidth}
        \includegraphics[width=\textwidth]{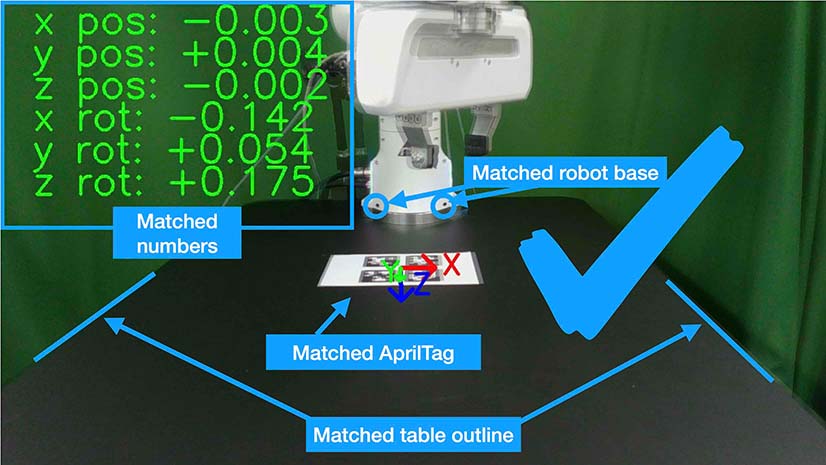}
        \caption{Correct numbers and images}
        \label{fig:instruction:setup_front_calibrated}
    \end{subfigure}
    \begin{subfigure}[t]{0.3\textwidth}
        \includegraphics[width=\textwidth]{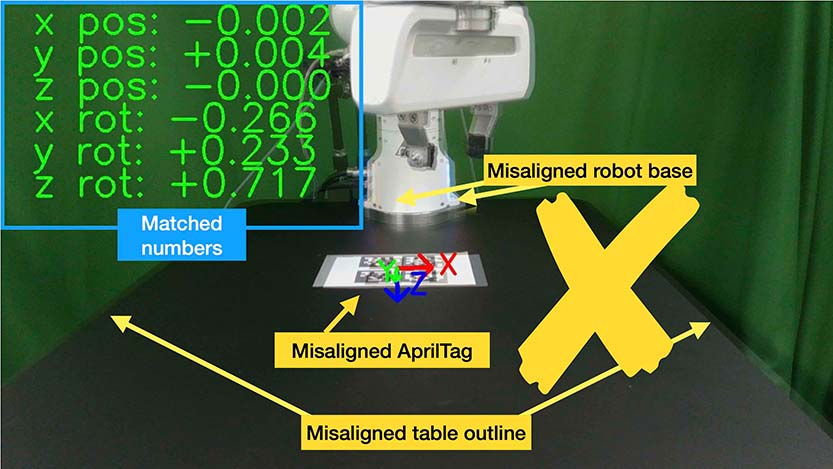}
        \caption{Correct numbers, incorrect images}
    \end{subfigure}
    \begin{subfigure}[t]{0.3\textwidth}
        \includegraphics[width=\textwidth]{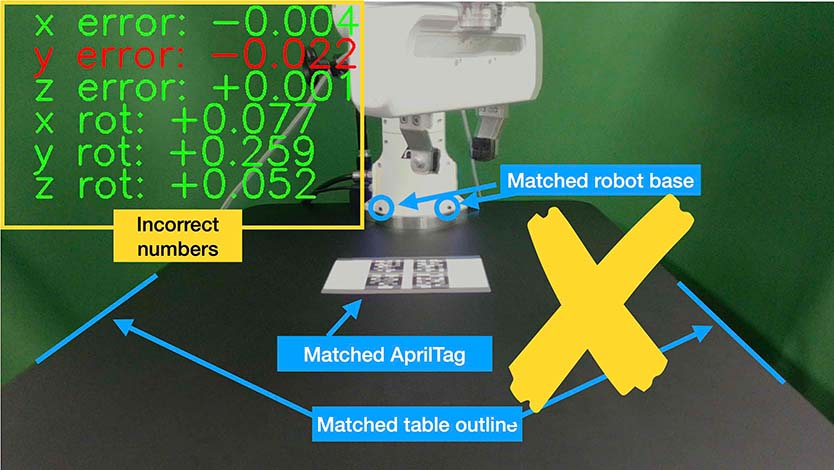}
        \caption{Correct images, incorrect numbers}
    \end{subfigure}
    \caption{\textbf{Camera calibration.} The numbers in the image indicate the deviation of the camera pose from desired camera pose. The green and red text indicates whether the camera pose is within the threshold or not.}
    \label{fig:instruction:camera_calibration}
\end{figure}

We provide a GUI tool to help calibrate the front camera pose with the pre-recorded view overlaid on top of the current camera view, as shown in \Cref{fig:instruction:camera_calibration}. The calibration can be achieved by matching the numbers and images shown in our calibration tool. 

In the GUI tool, the image from the current view is displayed as a solid layer, while the reference image you need to match appears transparent. The number indicates the deviation of the current camera poses from the desired pose. The red texts indicate that the deviation exceeds the threshold ($\pm0.004$ for the position (pos), $\pm0.8$ for the rotation (rot)), whereas green texts represent that it is within acceptable the boundary. Refer to \Cref{fig:instruction:base_apriltag_coordinate} for a better understanding of the coordinate system to adjust the camera pose.

\begin{enumerate}
    \item First, run the following command to move the robot up to prevent it from blocking the camera's view.
\begin{lstlisting}[language=bash]
 python furniture_bench/scripts/move_up.py \end{lstlisting}
    \item Run the camera calibration tool:
\begin{lstlisting}[language=bash]
 python furniture_bench/scripts/calibration.py --target setup_front \end{lstlisting}
    \item Adjust the camera to match \textcolor{red}{\textbf{both}} \textbf{images} and \textbf{numbers}.
\end{enumerate}

Here, we list \textbf{tips} to simplify the process of matching the camera pose.
\begin{itemize}
    \item For the Ulanzi camera mount, first adjust its height to match the z position and then fasten it in place.
    \item When dealing with the Manfrotto camera mount, prioritize matching all settings except for the x position, given that it can be independently modified using the camera bracket.
    \item In the beginning, ignore the numbers and focus on aligning the table outline and robot base (using the two holes in the robot base as reference points). Take a look at how the matched image looks like in \Cref{fig:instruction:setup_front_calibrated}
    \item Iterative adjust position and rotation to match the alignment and numbers. Based on our experience, it was simpler first to align the position and then adjust the rotation minutely for best alignment.
\end{itemize}

\textbf{Important Checklist:}
\begin{itemize}
    \item All numbers on the screen should turn green. \CheckBox[name=checkbox17, width=0.7em, height=0.7em]{} 
    \item The boundary of the table and the base AprilTag must be aligned with the pre-recorded image. \CheckBox[name=checkbox18, width=0.7em, height=0.7em]{} 
    \item The position of the robot base (i.e., two holes) should exactly match the pre-recorded image. \CheckBox[name=checkbox19, width=0.7em, height=0.7em]{} 
\end{itemize}

\clearpage

\subsection{Install Obstacle}

\begin{figure}[ht]
    \centering
    \includegraphics[width=0.6\linewidth]{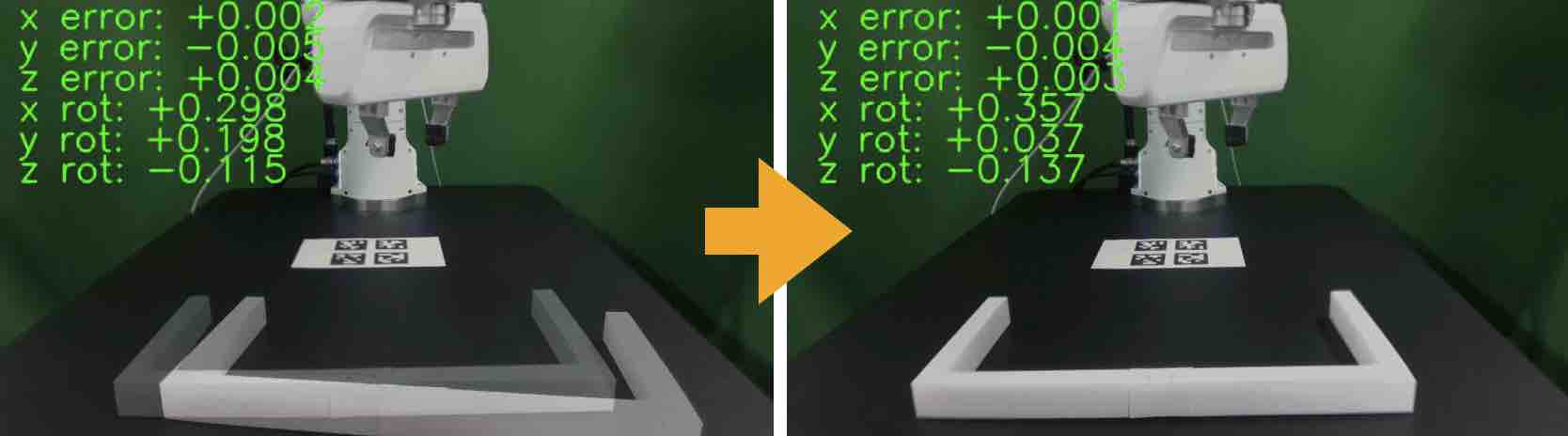}
    \caption{\textbf{Obstacle installation guidance tool.}}
    \label{fig:instruction:obstacle}
\end{figure}

The 3D-printed obstacle can be attached to the table using double-sided rubber tape. The exact pose of the obstacle can be viewed using our calibration tool, as shown in \Cref{fig:instruction:obstacle}.

\begin{figure}[ht]
    \centering
    \includegraphics[width=0.35\linewidth]{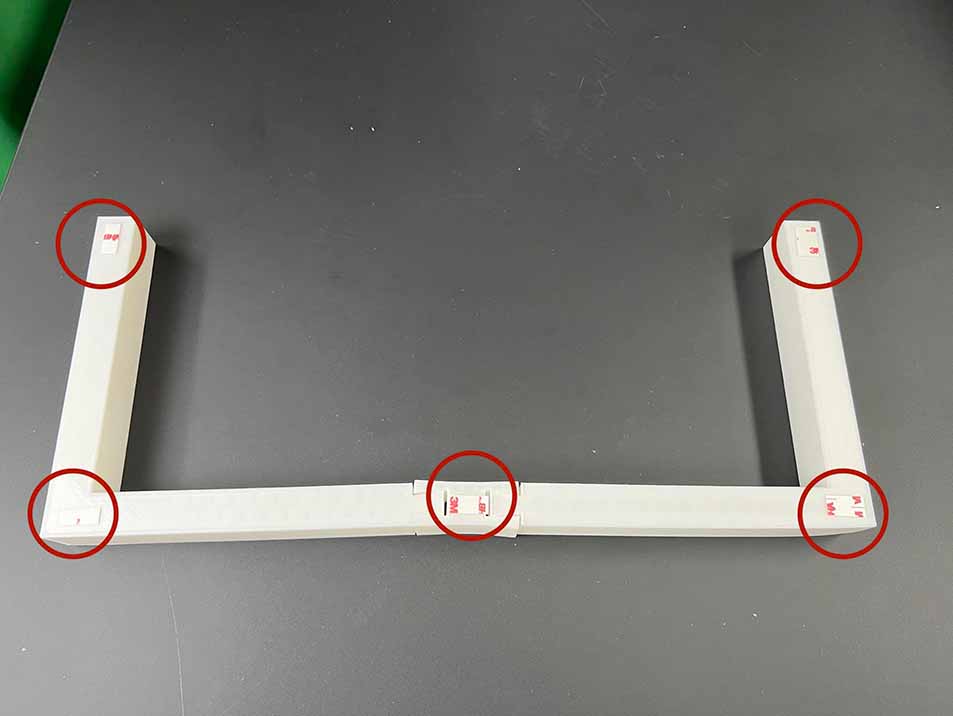}
    \caption{\textbf{Affix obstacle.} The red circles represent where to attach the double-sided rubber tape. Make sure the obstacle does not move when pushed.}
    \label{fig:instruction:obstacle_taping}
\end{figure}

\begin{enumerate}
    \item Install the obstacle with the guidance of the provided GUI tool: 
    \begin{lstlisting}[language=bash]
 python furniture_bench/scripts/calibration.py --target obstacle \end{lstlisting}
    \item Attach the obstacle to the table while aligning it with the pre-recorded obstacle pose.
    \item Affix the obstacle with double-sided rubber tape, as shown in \Cref{fig:instruction:obstacle_taping}.
\end{enumerate}

\textbf{Important Checklist:}
\begin{itemize}
    \item Adjust the obstacle to identically match the transparent one in the GUI tool, as shown in \Cref{fig:instruction:obstacle} on the right. There should be no discrepancy. \CheckBox[name=checkbox20, width=0.7em, height=0.7em]{} 
    \item Firmly attach the obstacle using double-sided rubber tape to prevent it from moving when pushed. \CheckBox[name=checkbox21, width=0.7em, height=0.7em]{} 
\end{itemize}

\subsection{Set Up Light}

During the data collection process, we randomize the light temperature between \SI{4600}{\kelvin}-\SI{6000}{\kelvin} as well as the intensity, position, and direction of the light. 
On the other hand, during the evaluation process, it is essential to maintain lighting conditions as similar as possible. In order to accomplish this, the light should be placed on the left side of the table as shown in \Cref{fig:instruction:setup_overview}. Furthermore, the temperature range of \SI{4600}{\kelvin} to \SI{6000}{\kelvin} and the brightness range of \SI{500}{\lumen} to \SI{1000}{\lumen} should be set for the lighting panel.


\clearpage

\subsection{Test the Environment}

To confirm that the environment setup has been done correctly, conduct a test run using a pre-trained policy for the \texttt{one\_leg} assembly task. The \texttt{one\_leg} assembly consists of phases: (1)~pick up the tabletop, (2)~push to the corner, (3)~pick up the leg, (4)~insert the leg, and (5)~screw the leg. The assessment outcomes can be contrasted with the initial environment depicted in \Cref{fig:reproducibility_performance}. The pre-trained policy is expected to complete over 3 phases on average (such as picking up the leg and occasionally inserting it), achieving a $15$-$30$\% success rate in the entire \texttt{one\_leg} assembly task.

\begin{figure}[ht]
    \centering
    \includegraphics[width=0.3\textwidth]{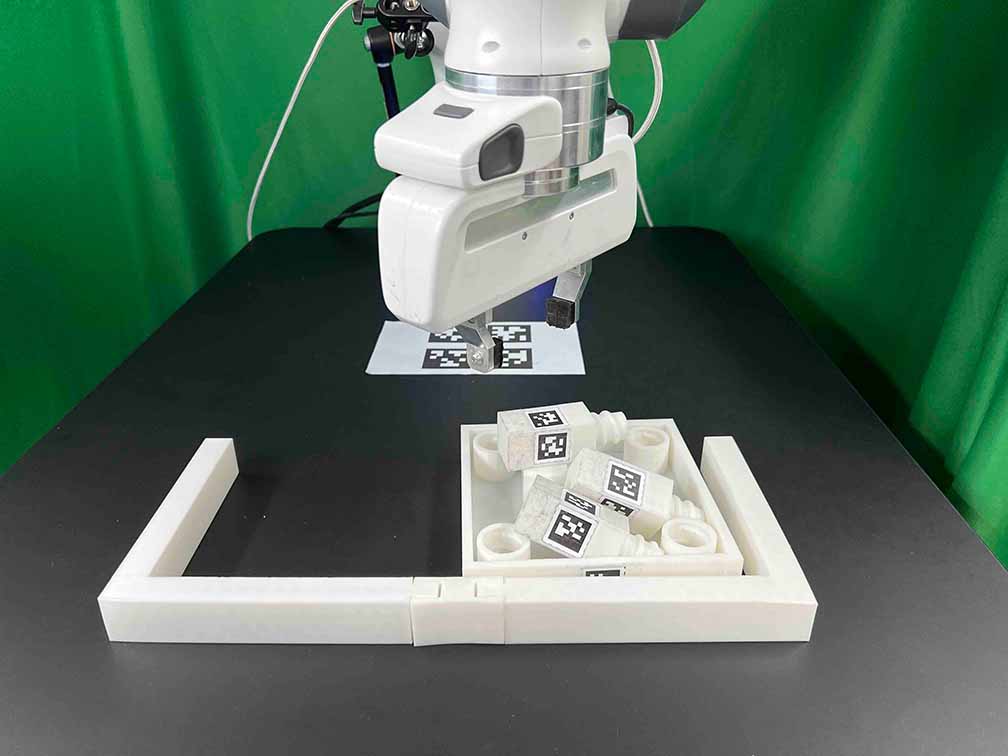}
    \caption{\textbf{Random furniture placement before accurate task initialization.}}
    \label{fig:instruction:furniture_placement}
\end{figure}

\begin{figure}[h]
    \centering
    \begin{subfigure}[t]{0.45\textwidth}
        \includegraphics[width=\textwidth]{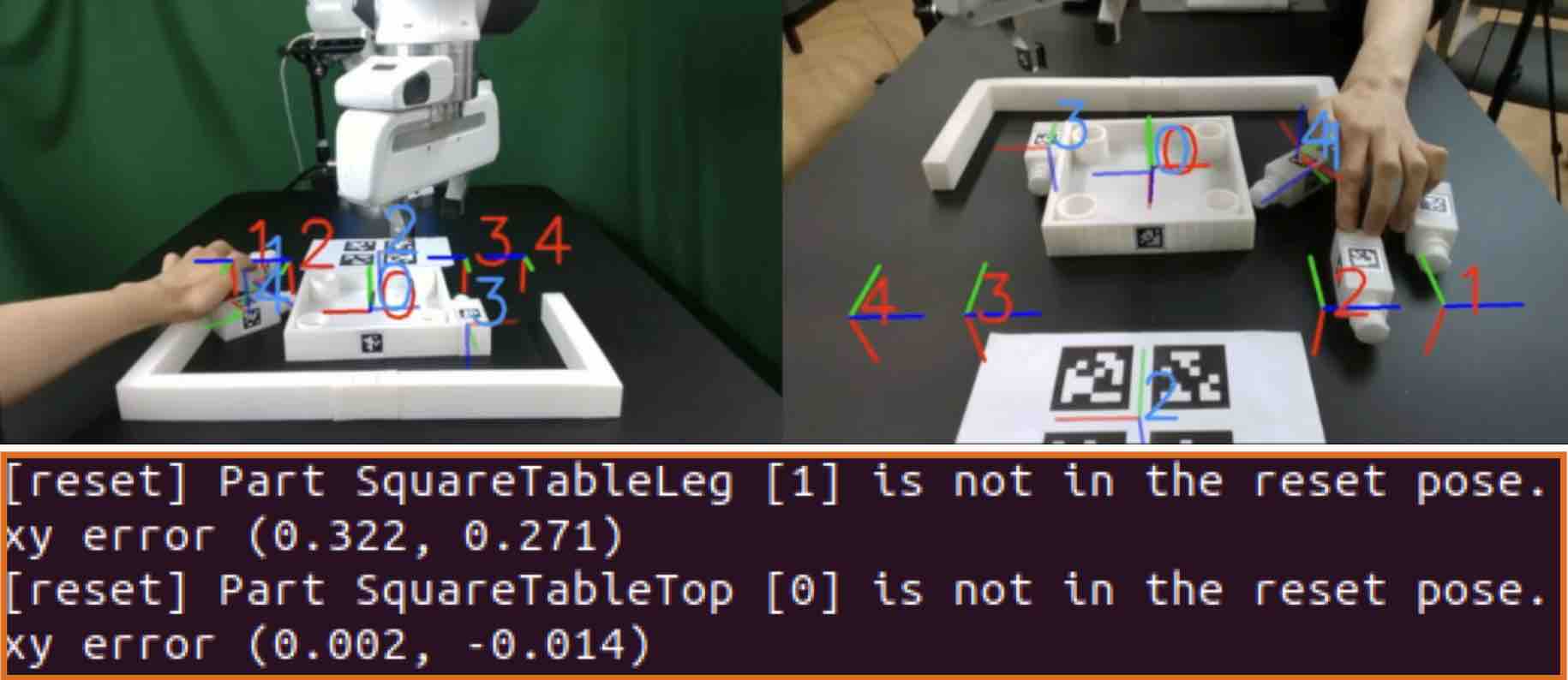}
        \caption{GUI tool and prompt indicate where to place each part.}
        \label{fig:instruction:initialization_GUI_prompt}
    \end{subfigure}
    \hspace{1em}
    \begin{subfigure}[t]{0.45\textwidth}
        \includegraphics[width=\textwidth]{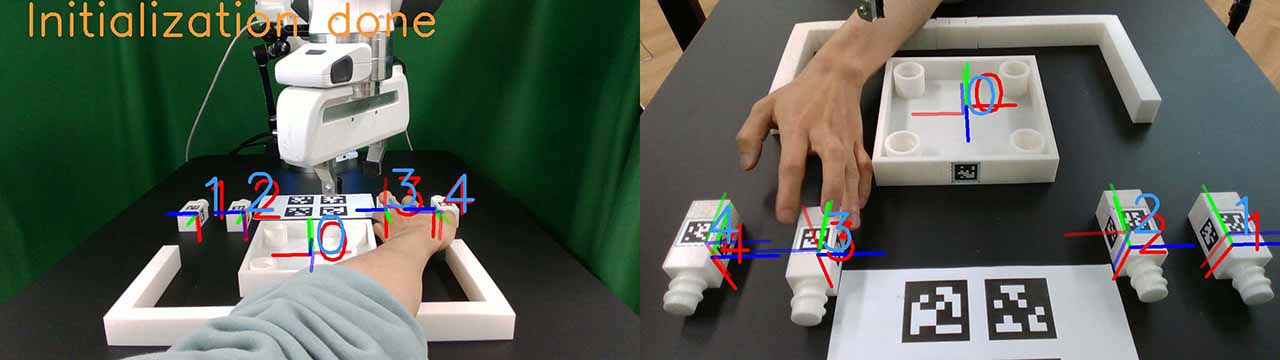}
        \caption{Once ``Initialization done'' is prompted, press ``Enter'' to proceed and execute the policy.}
        \label{fig:instruction:initialization_done}
    \end{subfigure}
    \caption{\textbf{Task initialization process using the GUI tool.}}
    \label{fig:instruction:initialization_process}
\end{figure}

\begin{enumerate}
    \item Before evaluation, double-check the camera calibration using the following script. All the numbers should be green and the robot base, obstacle, and base tag should be aligned accurately: 
\begin{lstlisting}[language=bash]
 python furniture_bench/scripts/calibration.py --target one_leg \end{lstlisting}
    \item Confirm the above \CheckBox[name=EvalCheck1, width=0.7em, height=0.7em]{}
    \item Green backdrop cloth has minimum wrinkles. \CheckBox[name=EvalCheck2, width=0.7em, height=0.7em]{}
    \item Wipe three camera lenses using a lens cloth, as they may be blurry from fingerprint smudges. \CheckBox[name=EvalCheck3, width=0.7em, height=0.7em]{}
 \item Install requirements for the evaluation:
\begin{lstlisting}[language=bash]
 pip install -r implicit_q_learning/requirements.txt
 pip install -e r3m
 pip install -e vip
\end{lstlisting}
    \item Place the furniture components randomly within the workspace, as shown in \Cref{fig:instruction:furniture_placement}.
    \item Evaluate the pre-trained policy using the following script:
\begin{lstlisting}[language=bash]
 ./evaluate.sh --low --one_leg \end{lstlisting}
 
    \item The above command will show a GUI tool and prompt to indicate where furniture parts should be positioned. Initialize the furniture parts, as shown in \cref{fig:instruction:initialization_GUI_prompt}. The screen will prompt ``initialization done'' when everything is correctly aligned, as shown in \Cref{fig:instruction:initialization_done}.
    \item Once the initialization is done, press ``Enter'' to execute the policy. Make sure there is nothing but furniture parts in the workspace.
\end{enumerate}

\end{document}